\documentclass[twocolumn]{svjour3}

\smartqed



\usepackage[OT1]{fontenc}
\usepackage[utf8]{inputenc}

\usepackage{amsmath,amsfonts,amssymb,amsthm}

\usepackage{graphicx}
\usepackage[table]{xcolor}
\usepackage{booktabs}
\usepackage{multirow}
\usepackage{array}
\usepackage{tabularx}
\usepackage{makecell}
\usepackage{ragged2e}
\usepackage{subcaption}
\usepackage{algorithm}
\usepackage[noend]{algpseudocode}
\usepackage[authoryear]{natbib}
\usepackage{pifont}

\usepackage{soul}
\usepackage{ifthen}
\usepackage[misc]{ifsym}
\usepackage{xspace}
\usepackage{siunitx}
\usepackage{enumitem}
\usepackage{pifont}
\usepackage{xcolor}

\usepackage[pagebackref=true,breaklinks=true,colorlinks,bookmarks=false]{hyperref}
\definecolor{cvprblue}{rgb}{0.21,0.49,0.74}
\definecolor{leafgreen}{rgb}{0.047,0.533,0.255}
\hypersetup{
    linkcolor = leafgreen,
    citecolor = cvprblue,
}

\newcommand{\figStaticComparison}{%
    \begin{figure*}[t!]
        \centering
        \renewcommand{\arraystretch}{0}
        \setlength{\tabcolsep}{1pt}
        \def \mywidth{0.33\linewidth}
        
        \resizebox{\textwidth}{!}{%
            \begin{tabular}{@{} P{0.5cm} P{\mywidth}P{\mywidth}P{\mywidth} @{}}

            \rotatebox{90}{\textbf{MVSplat}} & 
            \includegraphics[width=\linewidth]{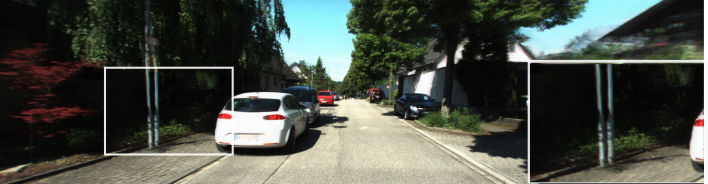}&
            \includegraphics[width=\linewidth]{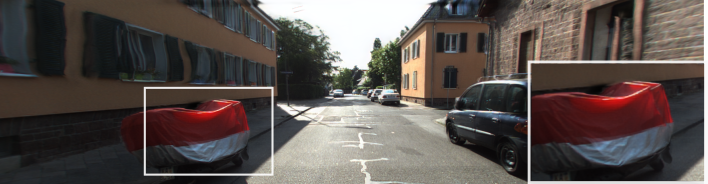}&
            \includegraphics[width=\linewidth]{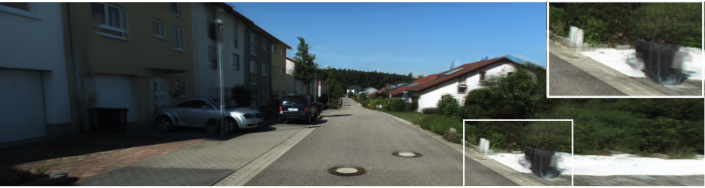} 
            \\

            \noalign{\vspace{1pt}}
            \rotatebox{90}{\textbf{EDUS}} & 
            \includegraphics[width=\linewidth]{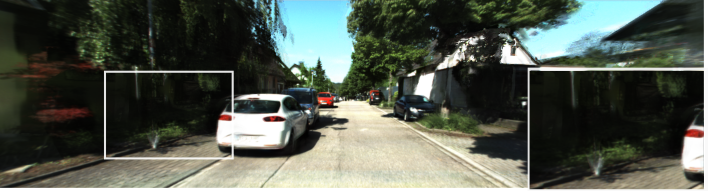}&
            \includegraphics[width=\linewidth]{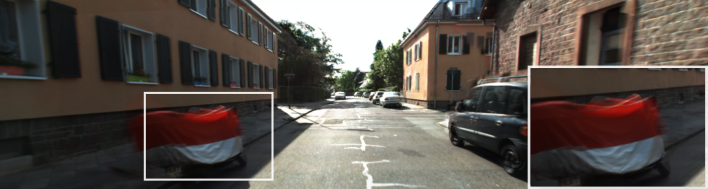}&
            \includegraphics[width=\linewidth]{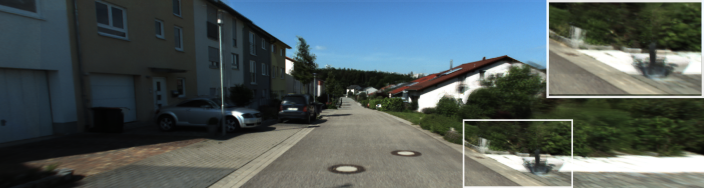} 
            \\

            \noalign{\vspace{1pt}}
            \rotatebox{90}{\textbf{AnySplat}} & 
            \includegraphics[width=\linewidth]{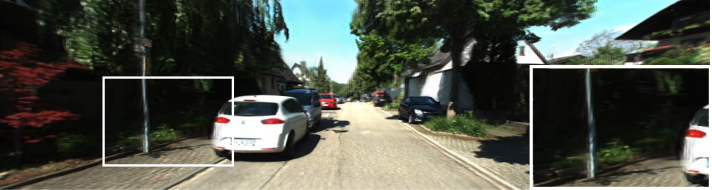}&
            \includegraphics[width=\linewidth]{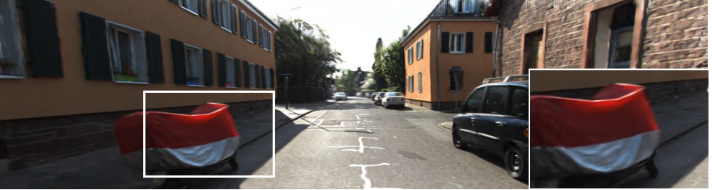}&
            \includegraphics[width=\linewidth]{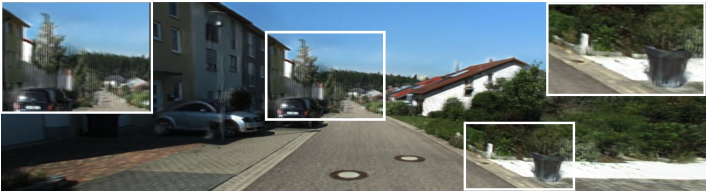} 
            \\

            \noalign{\vspace{1pt}}
            \rotatebox{90}{\textbf{Ours}} & 
            \includegraphics[width=\linewidth]{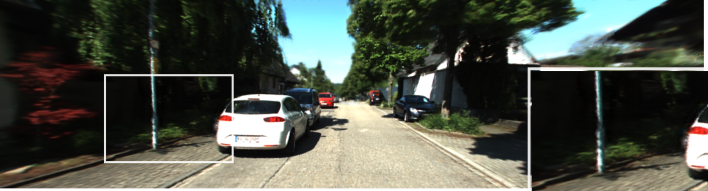}&
            \includegraphics[width=\linewidth]{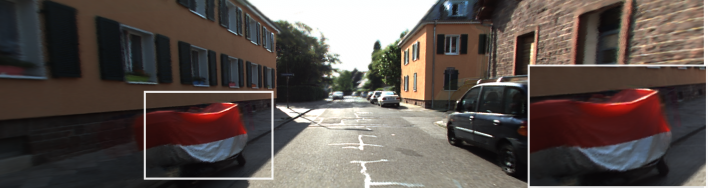}&
            \includegraphics[width=\linewidth]{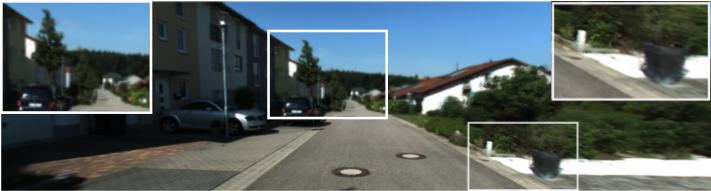} 
            \\

            \noalign{\vspace{1pt}}
            \rotatebox{90}{\textbf{GT}} & 
            \includegraphics[width=\linewidth]{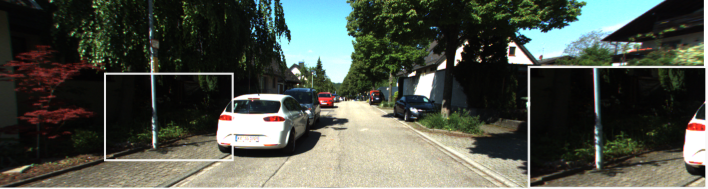}&
            \includegraphics[width=\linewidth]{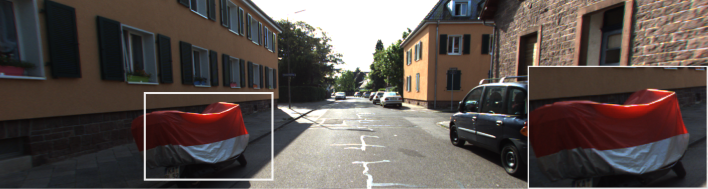}&
            \includegraphics[width=\linewidth]{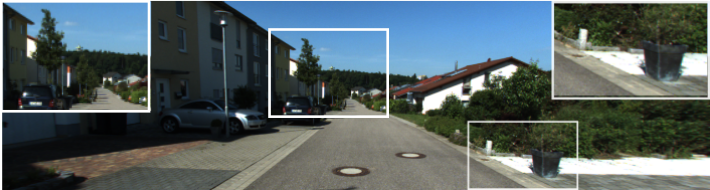}
            \\
            
            \end{tabular}%
        }
        
        \vspace{2mm}
        \caption{\textbf{Qualitative Comparison} with feed-forward baselines on the KITTI-360 dataset.}  
        \label{fig:static fig}
    \end{figure*}%
}

\newcommand{\figDynamicComparison}{%
    \begin{figure*}[t!]
    \centering
    \renewcommand{\arraystretch}{0} %
    \setlength{\tabcolsep}{1pt}     %
    
    \def\imgwidth{0.194\textwidth}
    
    \resizebox{\textwidth}{!}{%
        \begin{tabular}{@{} m{0.5cm} 
                            >{\centering\arraybackslash}m{\imgwidth}
                            >{\centering\arraybackslash}m{\imgwidth}
                            >{\centering\arraybackslash}m{\imgwidth}
                            >{\centering\arraybackslash}m{\imgwidth}
                            >{\centering\arraybackslash}m{\imgwidth} @{}}

        \rotatebox{90}{\textbf{Waymo}} & 
        
        {\includegraphics[width=\linewidth]{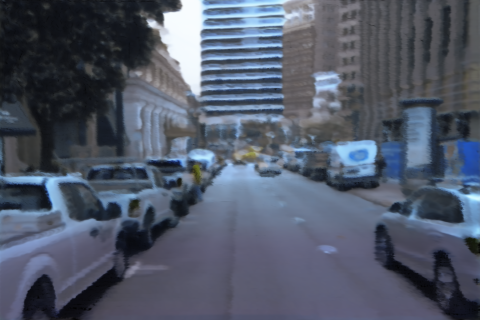} \par \vspace{1pt} \includegraphics[width=\linewidth]{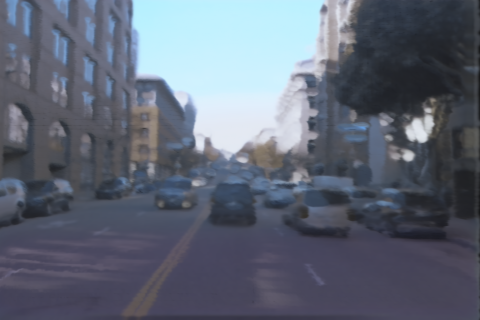}} &
        
        {\includegraphics[width=\linewidth]{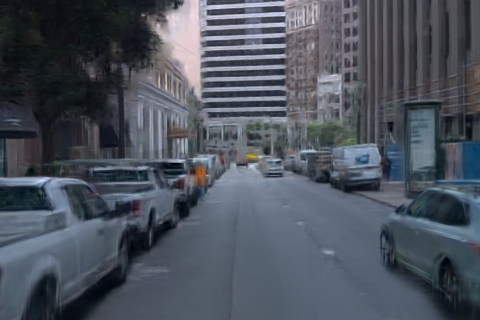} \par \vspace{1pt} \includegraphics[width=\linewidth]{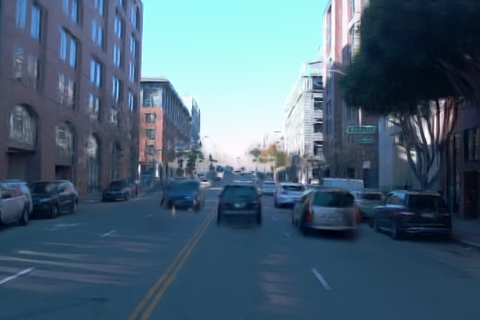}} &
        
        {\includegraphics[width=\linewidth]{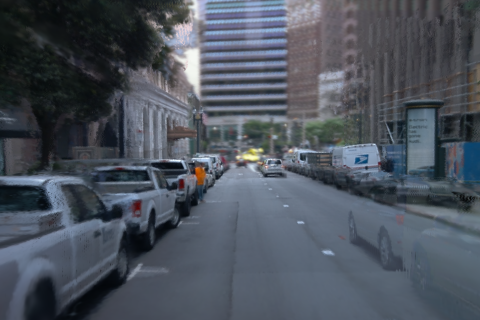} \par \vspace{1pt} \includegraphics[width=\linewidth]{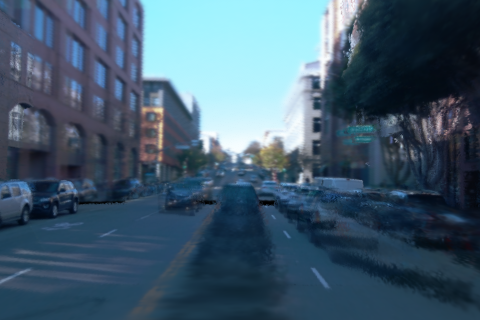}} &
        
        {\includegraphics[width=\linewidth]{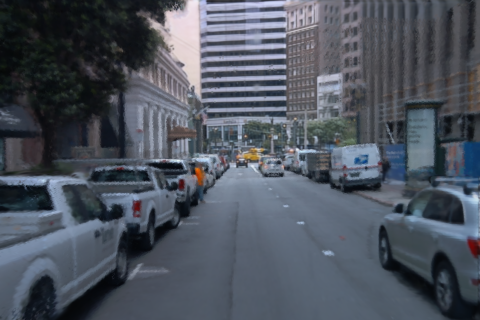} \par \vspace{1pt} \includegraphics[width=\linewidth]{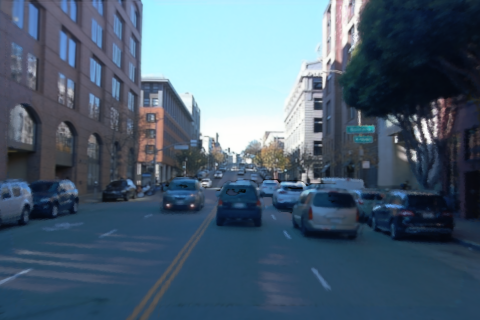}} &

        {\includegraphics[width=\linewidth]{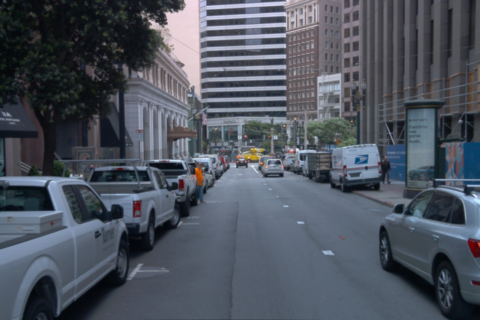} \par \vspace{1pt} \includegraphics[width=\linewidth]{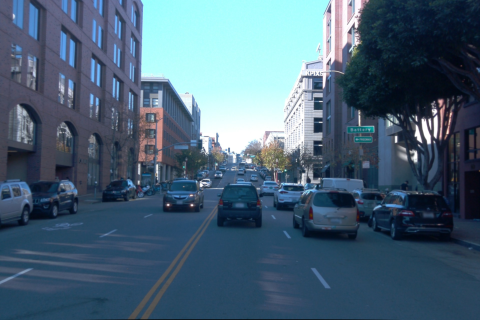}} 
        \\ 

        \noalign{\vspace{2pt}} 
        
        \rotatebox{90}{\textbf{PandaSet}} & 
        
        {\includegraphics[width=\linewidth]{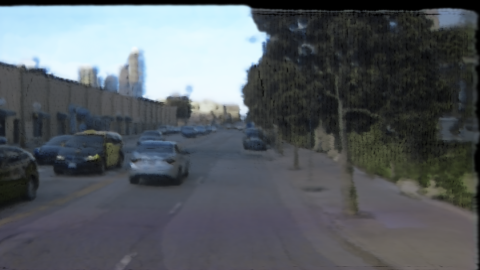} \par \vspace{1pt} \includegraphics[width=\linewidth]{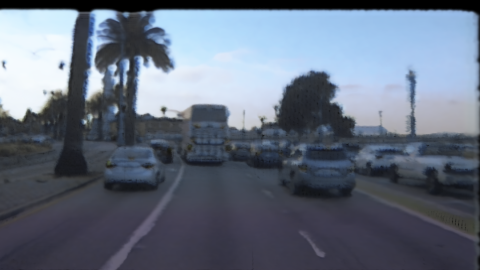}} &
        
        {\includegraphics[width=\linewidth]{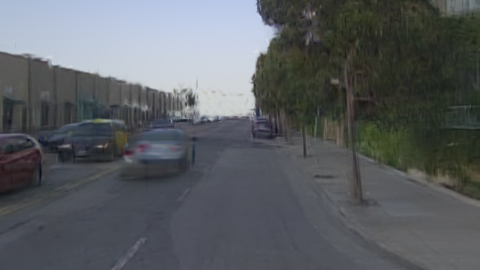} \par \vspace{1pt} \includegraphics[width=\linewidth]{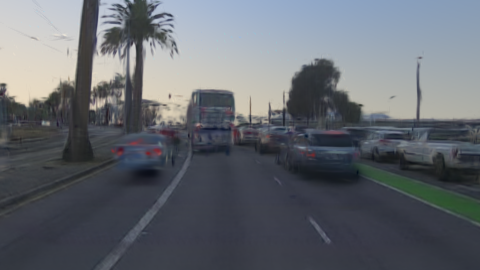}} &
        
        {\includegraphics[width=\linewidth]{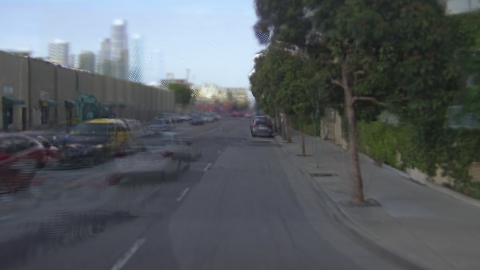} \par \vspace{1pt} \includegraphics[width=\linewidth]{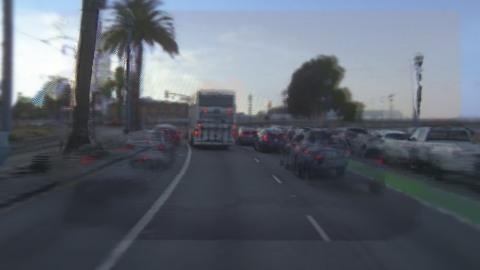}} &
        
        {\includegraphics[width=\linewidth]{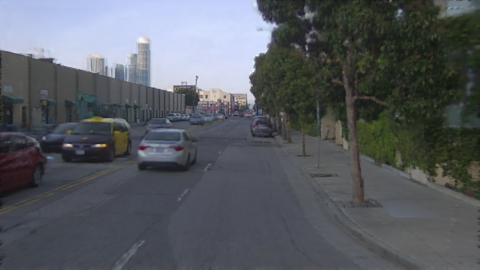} \par \vspace{1pt} \includegraphics[width=\linewidth]{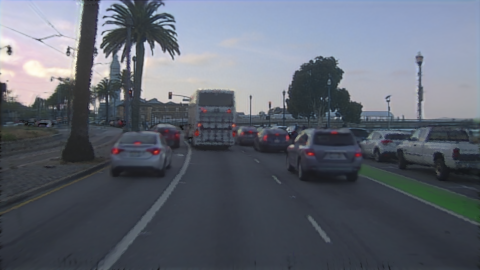}} &

        {\includegraphics[width=\linewidth]{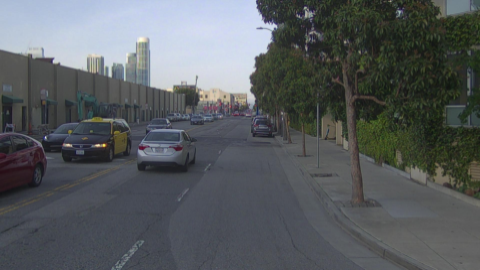} \par \vspace{1pt} \includegraphics[width=\linewidth]{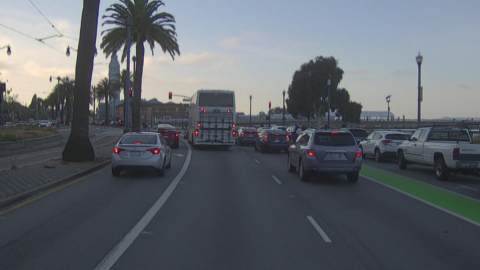}} 
        \\

        \noalign{\vspace{2pt}}
        & \textbf{(a) DrivingRecon} 
        & \textbf{(b) STORM} 
        & \textbf{(c) EVolSplat} 
        & \textbf{(d) Ours} 
        & \textbf{(e) GT} \\ %
        
        \end{tabular}%
    }
    
    \vspace{2mm}
    \caption{\textbf{Qualitative Comparisons} with baselines for feed-forward inference on the Waymo Open dataset and PandaSet. Each column represents a different method, while each pair of rows corresponds to a different dataset. The rendering images resolution are 480 $\times$ 320 for Waymo and 480 $\times$ 270 for PandaSet.}
    \label{fig:dynamic fig}
    \end{figure*}%
}

\newcommand{\figAblation}{%
\begin{figure}[t]
   \centering
   
     \begin{subfigure}[h]{0.48\linewidth}
         \centering
         \includegraphics[width=\textwidth]{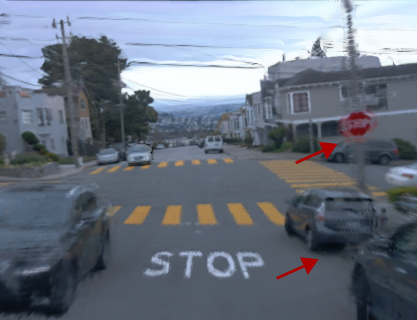}
        \caption{w/o volume branch}
        \label{fig:wo volume branch}
     \end{subfigure}
     \begin{subfigure}[h]{0.48\linewidth}
         \centering
         \includegraphics[width=\textwidth]{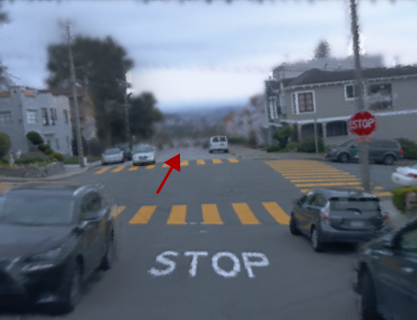}
         \caption{w/o pixel branch}
        \label{fig: wo pixel branch}
     \end{subfigure} \\
     
     \begin{subfigure}[h]{0.48\linewidth}
         \centering
         \includegraphics[width=\textwidth]{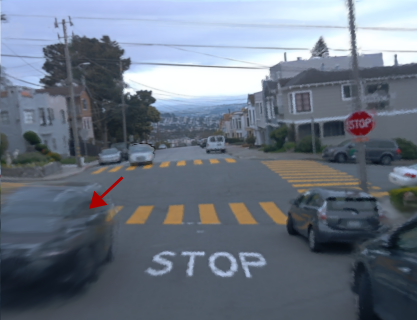}
        \caption{w/o motion-adjusted IBR}
        \label{fig: wo motion ibr}
     \end{subfigure}
      \begin{subfigure}[h]{0.48\linewidth}
         \centering
         \includegraphics[width=\textwidth]{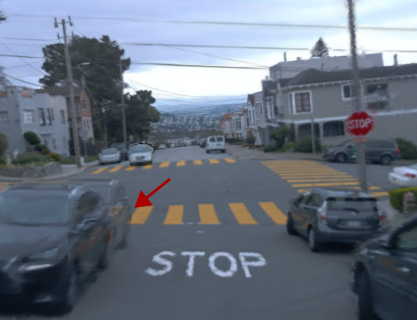}
        \caption{w/o Occ. Check}
        \label{fig:occ check}
     \end{subfigure} \\
     
    \begin{subfigure}[h]{0.48\linewidth}
         \centering
         \includegraphics[width=\textwidth]{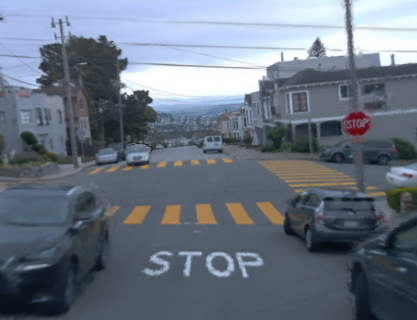}
        \caption{Full model}
        \label{fig:full model}
     \end{subfigure}
      \begin{subfigure}[h]{0.48\linewidth}
         \centering
         \includegraphics[width=\textwidth]{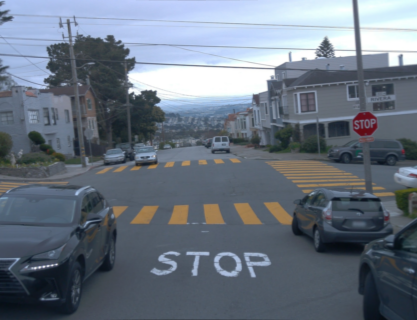}
        \caption{Ground Truth}
        \label{fig:ground truth}
     \end{subfigure}

     \vspace{-0.2cm}
             \caption{\textbf{Qualitative Results of Ablation Study.} We represent rendering images via feed-forward inference on one novel scene.}
        \label{fig:Ablation study.}
\end{figure}
}

\newcommand{\figEdit}{%
    \begin{figure}[t]
         \centering
         \captionsetup[subfigure]{justification=centering} 

    \begin{subfigure}{\columnwidth} %
        \centering
        
        \includegraphics[width=0.49\linewidth]{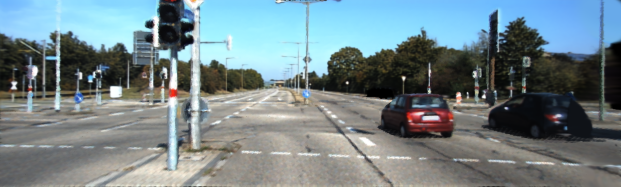}
        \hfill %
        \includegraphics[width=0.49\linewidth]{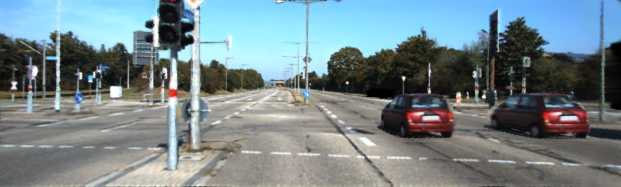}
        
        \caption{Object Replacement} 
        \label{fig:replaced}
    \end{subfigure}
    \\

    \begin{subfigure}{\columnwidth} %
        \centering
        \includegraphics[width=0.49\linewidth]{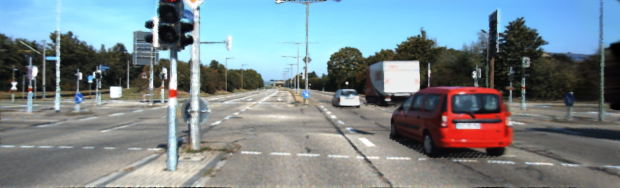}
        \hfill
        \includegraphics[width=0.49\linewidth]{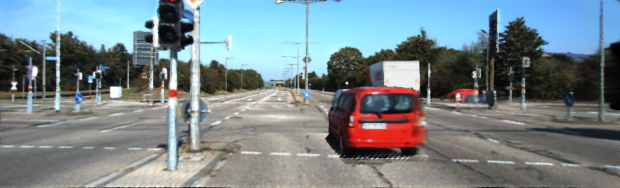}
        \caption{Object Shift} 
        \label{fig:shift}
    \end{subfigure}
    \\

    \begin{subfigure}{\columnwidth} %
        \centering
        \includegraphics[width=0.49\linewidth]{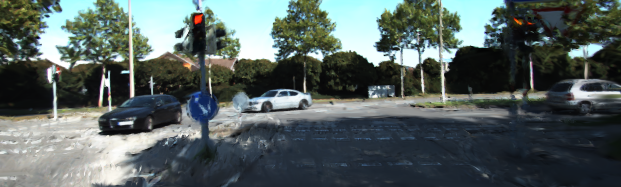}
        \hfill
        \includegraphics[width=0.49\linewidth]{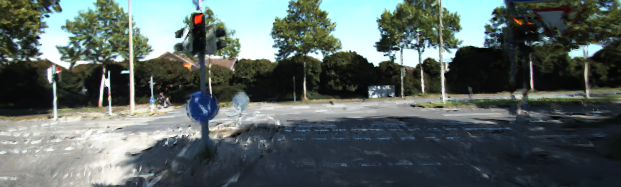}
        
        \caption{Object Deletion} 
        \label{fig:delete}
    \end{subfigure}
    
         \caption{\textbf{Scene Editing on KITTI.} Images in the first and second columns represent the results before and after editing.} 
         \label{fig:scene edit}
         \vspace{-0.4cm}
    \end{figure}
}

\newcommand{\figfinetuning}{%
    \begin{figure}[tpb]
      \centering
       \includegraphics[width=\linewidth]{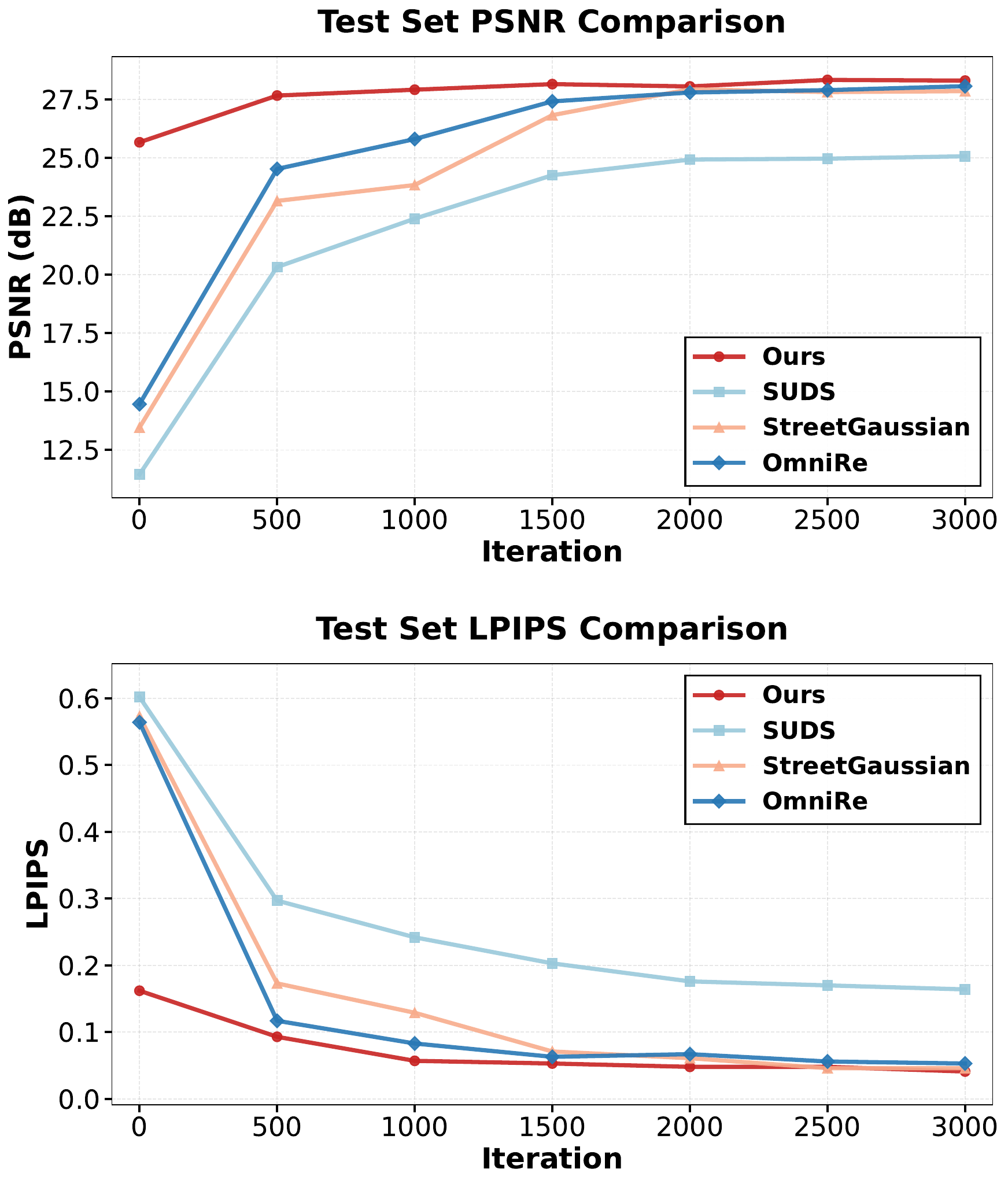}
       \vspace{-0.7cm}
       \caption{\textbf{Convergence Analysis} on the Waymo Dataset. We plot PSNR and LPIPS on one test set at different training steps. Compared with optimization-based baselines, our method with generalizable priors converges faster ($\approx$ 1000 steps) and achieves better PSNR.}
       \label{fig:fineuning}
    \end{figure}
}

\newcommand{\tablefinetuning}{%
    \begin{table}[t]
    \centering

    \renewcommand{\arraystretch}{1.2} %
    
    \resizebox{\linewidth}{!}{%
    \begin{tabular}{@{}lcccc@{}}
    \toprule
                   Method & PSNR$\uparrow$  & SSIM$\uparrow$ &  LPIPS$\downarrow$ & Rec. Time$\downarrow$ \\
    \midrule 
    SUDS~\citep{turki2023suds}       &25.03   &0.813  & 0.167 &45min \\
    EmerNeRF~\citep{emernerf}       & 27.03   &0.886  & 0.069  &21min   \\
    StreetGaussian~\citep{streetgaussian} &27.27  &0.891  & 0.057 & 7min  \\
    DeSiRe-GS~\citep{peng2025desire} &27.35  &0.889  & 0.094 & 14min \\
    OmniRe~\citep{chen2024omnire}    &28.16  &\textbf{0.912}  & 0.051 & 9min  \\
    \midrule 
    EVolSplat4D (Ours$_{\text{0-iter}}$)   &25.39  &0.841  &0.119  & \textbf{1.3s} \\
    EVolSplat4D (Ours$_\text{1000-iter}$)   &\textbf{28.29}  &0.903  &\textbf{0.048}  & \textbf{3min 40s} \\
    \bottomrule
    \end{tabular}
    }
    \vspace{-0.2cm}
        \caption{\textbf{Comparison with Optimization-Based} baselines on 3 sequences of the Waymo NOTR dataset under the Drop 80\% sparsity level. The \textbf{Rec. Time} refers to the reconstruction time needed to reach convergence and achieve the reported metrics.}
    \label{tab:table_finetuning} 
    \end{table}
}

\providecommand{\tick}{\textcolor{green!60!black}{\ding{52}}}
\providecommand{\cross}{\textcolor{red}{\ding{56}}}

\newcommand{\baselineCategoryTableBody}{%
\centering
\setlength{\tabcolsep}{4.2pt}
\renewcommand{\arraystretch}{1.12}
\resizebox{\linewidth}{!}{%
\begin{tabular}{llccccccc}
\toprule
Cat.
& Method
& Sensors
& 3D Box Tracks
& Rep.
& Feed-Forward
& Dynamic
& Real-Time
& Scene Edit
\\
\midrule

\multirow{5}{*}{\shortstack[c]{Per-scene\\Opt.}}
& SUDS~\citep{turki2023suds}
& LiDAR+RGB & N & Non-PA & \cross & \tick & \cross & \cross \\

& EmerNeRF~\citep{emernerf}
& LiDAR+RGB & N & Non-PA & \cross & \tick & \cross & \cross \\

& StreetGaussian~\citep{streetgaussian}
& LiDAR+RGB & Pred & Non-PA & \cross & \tick & \tick & \tick \\

& OmniRe~\citep{chen2024omnire}
& LiDAR+RGB & GT & Non-PA & \cross & \tick & \tick & \tick \\

& DeSiRe-GS~\citep{peng2025desire}
& LiDAR+RGB & N & Non-PA & \cross & \tick & \tick & \cross \\

\midrule

\multirow{5}{*}{\shortstack[c]{FeedForward\\Recon.}}
& AnySplat~\citep{Jiang2025AnySplatF3}
& RGB & N & PA & \tick & \cross & \tick & \cross \\

& DrivingForward~\citep{tian2024drivingforward}
& RGB & N & PA & \tick & \cross & \tick & \cross \\

& STORM~\citep{yang2024storm}
& RGB & N & PA & \tick & \tick & \tick & \tick \\

& DrivingRecon~\citep{lu2024drivingrecon}
& LiDAR+RGB & GT & PA & \tick & \tick & \tick & \tick \\

& \textbf{EVolSplat4D (Ours)}
& LiDAR+RGB & Pred & Non-PA & \tick & \tick & \tick & \tick \\

\bottomrule
\end{tabular}
}
\caption{
\textbf{Categorization of representative baselines.}
We categorize representative methods into per-scene optimization and feed-forward reconstruction methods.
PA and Non-PA denote pixel-aligned and non-pixel-aligned representations, respectively.
}
}

\newcommand{\baselineCategorize}{%
\begin{table*}[htbp]
\baselineCategoryTableBody
\label{tab:baseline_group}
\end{table*}
}

\newcommand{\cref}[1]{\autoref{#1}}
\newcommand{\Cref}[1]{\autoref{#1}}

\graphicspath{{gfx/}}
\renewcommand{\arraystretch}{1.05}

\newcommand{\bx}{\mathbf{x}}

\newcommand{\bS}{\mathbf{S}}

\newcommand{\bR}{\mathbf{R}}

\newcommand{\bp}{\mathbf{p}}

\newcommand{\bc}{\mathbf{c}}

\newcommand{\bSigma}{\boldsymbol{\Sigma}}

\newcommand{\nR}{\mathbb{R}}

\newcommand{\figref}[1]{\Fig~\ref{#1}}
\newcommand{\secref}[1]{Section~\ref{#1}}

\renewcommand{\eqref}[1]{Eq.~\ref{#1}}
\newcommand{\tabref}[1]{Table~\ref{#1}}

\renewcommand{\b}{\ensuremath{\mathbf}}

\makeatletter
\DeclareRobustCommand\onedot{\futurelet\@let@token\@onedot}
\def\@onedot{\ifx\@let@token.\else.\null\fi\xspace}
\def\eg{e.g\onedot}

\def\Fig{Fig\onedot}   
\makeatother

\definecolor{applegreen}{rgb}{0.55, 0.71, 0.0}

\newcommand{\boldparagraph}[1]{\vspace{0.2cm}\noindent{\bf #1:} }

\newcolumntype{P}[1]{>{\centering\arraybackslash}m{#1}}

\newcommand{\del}[1]{}

\newif\ifcomment
\commenttrue
\ifcomment
	\newcommand{\ag}[1]{ \noindent {\color{red} {\bf Andreas:} {#1}} }
	\newcommand{\yl}[1]{ \noindent {\color{cyan} {\bf Yiyi:} {#1}} }

\else
	\newcommand{\ag}[1]{}
	\newcommand{\jx}[1]{}
	\newcommand{\kc}[1]{}
	\newcommand{\yl}[1]{}
\fi

\ifcomment

\else
	\newcommand{\ag}[1]{}
	\newcommand{\yl}[1]{}
    \newcommand{\ms}[1]{}
\fi

\newcommand{\method}{EVolSplat4D}

\begin{document}

\title{EVolSplat4D: Efficient Volume-based Gaussian Splatting for 4D Urban Scene Synthesis}

\author{Sheng Miao \and Sijin Li \and Pan Wang \and Dongfeng Bai \and Bingbing Liu \and Yue Wang \and Andreas Geiger \and Yiyi Liao$\textsuperscript{\Letter}$}

\institute{
    Yiyi Liao (\Letter) \at \email{yiyi.liao@zju.edu.cn} \\
    Sheng Miao \at \email{shengmiao@zju.edu.cn} \and
    S. Miao, S. Li, Y. Wang, Y. Liao \at Zhejiang University, China. \and
    P. Wang, D. Bai, B. Liu \at Huawei, China. \and
    A. Geiger \at KE:SAI, University of T{\"u}bingen, Germany. \\[1em]
    \textbf{Project Page:}~\href{https://xdimlab.github.io/EVolSplat4D/}{https://xdimlab.github.io/EVolSplat4D/}
}

\maketitle

\begin{abstract}

Novel view synthesis (NVS) of static and dynamic urban scenes is essential for autonomous driving simulation, yet existing methods often struggle to balance reconstruction time with quality. While state-of-the-art neural radiance fields and 3D Gaussian Splatting approaches achieve photorealism, they often rely on time-consuming per-scene optimization. Conversely, emerging feed-forward methods frequently adopt per-pixel Gaussian representations, which lead to 3D inconsistencies when aggregating multi-view predictions in complex, dynamic environments. We propose EvolSplat4D, a feed-forward framework that moves beyond existing per-pixel paradigms by unifying volume-based and pixel-based Gaussian prediction across three specialized branches. For close-range static regions, we predict consistent geometry of 3D Gaussians over multiple frames directly from a 3D feature volume, complemented by a semantically-enhanced image-based rendering module for predicting their appearance. For dynamic actors, we utilize object-centric canonical spaces and a motion-adjusted rendering module to aggregate temporal features, ensuring stable 4D reconstruction despite noisy motion priors. Far-Field scenery is handled by an efficient per-pixel Gaussian branch to ensure full-scene coverage. Experimental results on the KITTI-360, KITTI, Waymo, and PandaSet datasets show that EvolSplat4D reconstructs both static and dynamic environments with superior accuracy and consistency, outperforming both per-scene optimization and state-of-the-art feed-forward baselines.
\end{abstract}

\section{Introduction}\label{sec:introduction}

\begin{figure*}
  \centering
   \includegraphics[width=\linewidth]{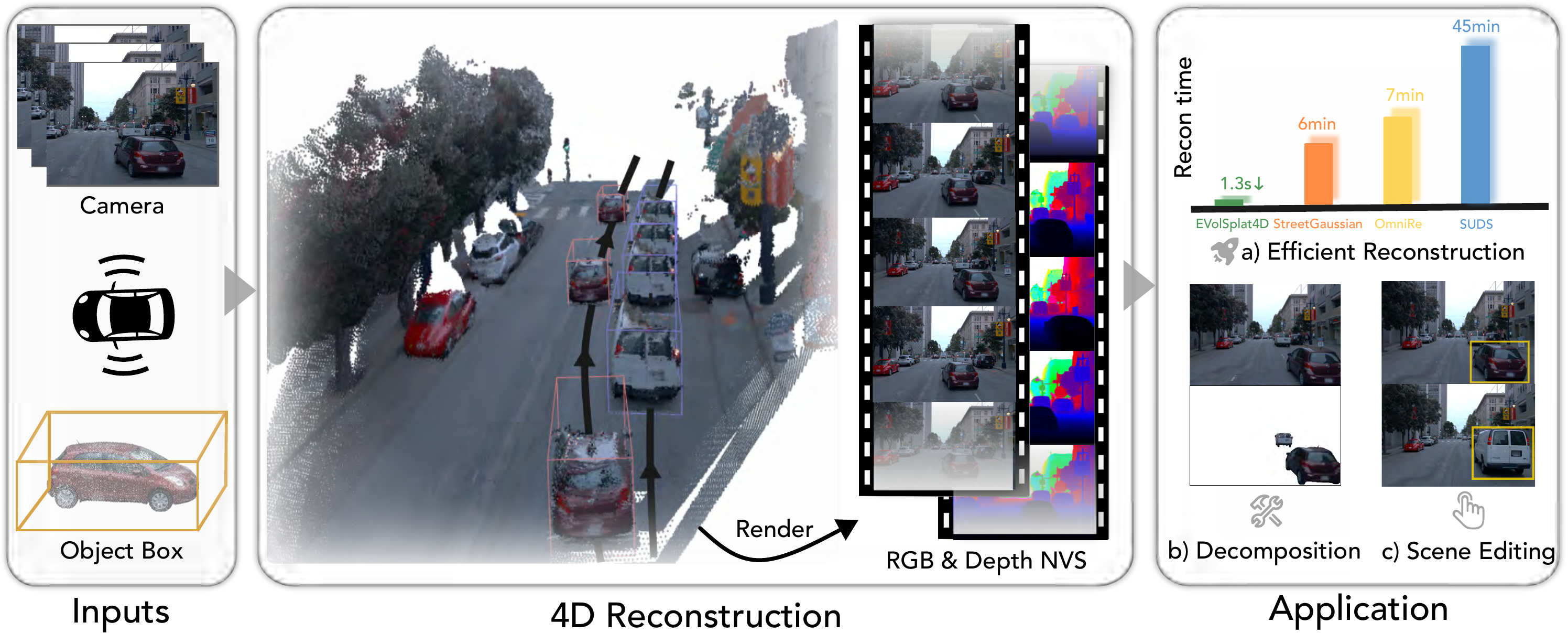}
   \caption{\textbf{Overview of~\method}. We propose~\method, a unified feed-forward 3D Gaussian Splatting framework tailored for static \& dynamic urban scenes that achieves real-time rendering speeds. Leveraging both camera and tracked 3D bounding box as inputs, EVolSplat4D completes scene reconstruction in approximately 1.3 seconds, achieving photo-realistic quality comparable to time-consuming per-scene optimization methods. EVolSplat4D also supports various downstream applications, including high-fidelity scene editing and scene decomposition.}
   \label{fig: teaser}
   \vspace{-0.5cm}
\end{figure*}

Novel view synthesis (NVS) and scene reconstruction are central to autonomous driving simulation, where photo-realistic and geometrically accurate digital twins of real-world environments at scale are required for training and evaluating autonomous driving algorithms across diverse scenarios. 
Building on the recent progress in neural radiance fields (NeRF)~\citep{nerf} and 3D Gaussian Splatting (3DGS)~\citep{3dgs}, several works~\citep{urbanradiancefield,unisim,f2nerf,ye2024gaustudio,hierarchicalgs,wu2024recent} have achieved promising NVS performance for urban street scenes.
However, all these approaches rely on per-scene optimization, which becomes prohibitively time-consuming when scaling to a large number of urban environments for large-scale simulation.

To remedy this issue, feed-forward methods aim to regress scene representations in a single forward pass. Early approaches learn generalizable NeRFs~\citep{xu2023murf,miao2024efficient,mvsnerf} from input images, but suffer from slow inference due to expensive volumetric sampling. To enable real-time rendering, recent works instead regress 3D Gaussians directly. These methods share a similar design, using 2D CNN and transformer architectures to predict \textit{pixel-aligned} 3D Gaussians from 2D reference images~\citep{szymanowicz2024splatter,charatan2024pixelsplat,chen2024mvsplat,xu2025depthsplat}.
Although effective for efficient reconstruction and real-time, high-quality NVS in static scenes, these approaches do not directly generalize to dynamic driving scenarios. Recently, pioneering works such as DrivingRecon~\citep{lu2024drivingrecon} and STORM~\citep{yang2024storm} extend pixel-aligned designs to dynamic street reconstruction. However, they still rely on \textit{per-pixel} 3D Gaussian representations, which can easily lead to 3D inconsistencies (e.g., multi-layer artifacts) when a target view aggregates 3D Gaussian predictions from multi-view neighboring viewpoints. The issue is further intensified for dynamic objects, where accurate aggregation additionally depends on precise 3D motion estimation. Moreover, the number of 3D Gaussians scales linearly with the number of 2D pixels, leading to high memory cost and redundant primitives in overlapping regions.

In this paper, we investigate a key question in feed-forward scene reconstruction: is per-pixel prediction the optimal representation, particularly for dynamic urban scenes? 
We propose \method, a feed-forward 3D Gaussian prediction framework for both static and dynamic urban scenes that goes beyond per-pixel representations. Our key idea is to directly learn scene representations in 3D Euclidean space for close-range regions and dynamic objects, while retaining pixel-aligned Gaussian prediction for far-range regions. This hybrid design enables multi-view consistent reconstruction for close-range static volumes and dynamic objects while efficiently modeling unbounded regions.

Specifically, we decompose the scene into three components: the static close-range volume, dynamic vehicles, and the far-field scenery, and design a dedicated feed-forward module for each. For the static close-range volume, we depart from per-pixel–aligned prediction and instead first construct a global 3D semantic field by lifting 2D semantic features (DINO~\citep{oquab2023dinov2}) across multiple frames into Euclidean space using noisy depth estimates. Geometry is then predicted directly in 3D by decoding this field into Gaussian parameters with a 3D convolutional network, yielding view-consistent structure. Given the predicted geometry, appearance is queried via an occlusion-aware image-based rendering (IBR) module that leverages semantic features to blend multi-view colors robustly.
For dynamic vehicles, we adopt the same principle of avoiding per-pixel–aligned prediction but operate in an object-centric canonical space. Each moving instance is disentangled from the static scene and mapped to a canonical coordinate system, utilizing estimated 3D bounding boxes to parameterize time-dependent rigid motions. While these bounding boxes may be noisy, our motion-adjusted IBR module robustly aggregates appearance features across multiple timestamps and viewpoints, mitigating misalignment and enabling consistent, feed-forward dynamic reconstruction.
Finally, far-range regions and the sky are handled by a per-pixel Gaussian decoder based on a cross-view attention 2D U-Net. To obtain the final image, we integrate the predicted Gaussians from all three modules and perform $\alpha$-blending, allowing efficient generalization to the full scene.

We validate EVolSplat4D with extensive experiments on KITTI-360, KITTI, Waymo, and PandaSet across multiple autonomous driving settings. We summarize our contributions as follows:

\begin{itemize}[noitemsep]
\item We introduce EvolSplat4D, a novel Gaussian-based framework for unbounded urban scenes that factorizes the environment into three generalizable components: close-range volume, dynamic vehicles, and far-field scenery. Our method supports both static and dynamic feed-forward reconstruction, enabling real-time, photo-realistic rendering from sparse input images.

\item We propose a volume-based 3D Gaussian reconstruction approach for close-range static regions, in which geometry and appearance predictions are decoupled. Geometry is predicted in a view-consistent 3D space across multiple frames, while a semantic-enhanced IBR module robustly infers appearance and mitigates noise in the geometry.

\item We leverage predicted 3D bounding boxes to model dynamic actors in object-centric canonical spaces and propose a motion-adjusted IBR module for generalizable 4D reconstruction. By robustly aggregating temporal features, this module mitigates the effect of noisy 3D bounding boxes to enable high-fidelity rendering and controllable scenario editing.

\end{itemize}

This journal paper is an extension of a conference paper published at CVPR 2025, EVolSplat \cite{miao2025evolsplat}. In comparison to EVolSplat, which is limited to static reconstruction, we 1) incorporate additional geometric cues (e.g., 3D bounding boxes) and propose a novel motion-adjusted IBR technique to achieve unified feed-forward reconstruction for both static and dynamic scenes;  2) design a specialized far-field branch equipped with cross-view attention to improve the background fidelity; 3) integrate self-supervised semantic priors to augment the occlusion module for more robust detection of occluded regions, and 4) extensively enrich the experimental section by including new datasets (KITTI and PandaSet) and conduct additional scene editing experiments.

\section{Related Work}
\label{related work}
\subsection{Novel View Synthesis for Urban Scenes}

\boldparagraph{Static Scenes}
The rapid development of radiance field techniques~\citep{3dgs,nerf,ma20243d,ye2025gaussian} have significantly advanced novel view synthesis for driving scenes. These approaches represent street scenes as implicit neural fields~\citep{streetsurf,snerf,urbanradiancefield,blocknerf,f2nerf}  or anisotropic 3D Gaussians ~\citep{lin2024vastgaussian,liu2024citygaussian,hierarchicalgs}, achieving expressive synthesis.
A few works~\citep{snerf,urbanradiancefield,unisim} leverage sparse vehicle-mounted LiDAR sensors to boost the training and learn a robust geometry. Other works also incorporate additional semantic and geometry cues as priors or supervision to enhance scene comprehension, including semantic understanding~\citep{fu2023panopticnerf,pnf} and geometric constraints~\citep{cui2025streetsurfgs, streetsurf}. However, these approaches primarily target static scenes and struggle to handle dynamic agents in urban environments.

\boldparagraph{Dynamic Scenes}
Subsequent methods have addressed dynamic elements by constructing scene graphs, where each moving instance is separated from the static background using annotated 3D bounding boxes, as adopted in ~\citep{hugs,zhou2024drivinggaussian,streetgaussian,chen2024omnire,khan2024autosplat,wu2023mars,Neuralscenegraph}.  While this enables accurate reconstruction and object-level editing, these methods use 3D box tracks as motion priors and still require expensive per-scene optimization. In contrast, our feed-forward method uses predicted 3D box tracks with the motion-adjusted IBR module to directly produce renderable object-centric Gaussians, and aims to perform efficient reconstruction on novel dynamic scenes. Other approaches eliminate the need for 3D bounding box by learning to decompose the static and dynamic components in a self-supervised manner. SUDS~\citep{turki2023suds} and EmerNeRF~\citep{emernerf} model motion implicitly in 4D spacetime between frames, using foundation models like DINO~\citep{oquab2023dinov2} to establish correspondences in challenging conditions. In parallel, Dynamic Gaussian methods learning attributes like displacement~\citep{huang2024textit} or lifespan~\citep{chen2023periodic,peng2025desire} for each primitive, distinguishing static from dynamic ones through available 2D image observations. Despite these advances, separation remains incomplete~\citep{fan2025advances}. In our research, we utilize predicted 3D bounding box to decouple scenes into a static reference space and time-dependent deformations, facilitating pose editing and novel pose synthesis applications that are challenging with 4D spacetime representations.

\subsection{2D Supervised Feed-Forward Reconstruction}
To perform feed-forward reconstruction, researchers train neural networks across large-scale datasets to gain domain-specific prior knowledge. These methods~\citep{pixelnerf,charatan2024pixelsplat,chen2024mvsplat,zhang2024transplat,gslrm,szymanowicz2024splatter} have evolved to work with sparse image sets and can directly reconstruct scenes via fast feed-forward inference. Generalizable NeRF performs ray-based rendering and estimates the 3D radiance field of scenes, which enables superior cross-dataset generalization capabilities in both small~\citep{mvsnerf,ibrnet,pixelnerf} and large baseline~\citep{xu2023murf,du2023learning} settings. However, NeRF based methods are notoriously slow to render. Another line of work directly predicts pixel-aligned 3D Gaussians from sparse reference images~\citep{szymanowicz2024flash3d,szymanowicz2024splatter,gslrm,chen2024mvsplat,liang2024bullettimer,chen2025g3r}. These methods either utilize an epipolar transformer~\citep{charatan2024pixelsplat,wewer2024latentsplat} or construct local cost volumes~\citep{chen2024mvsplat,zhang2024transplat,liu2025mvsgaussian} for learning geometry. However, most of these methods focus on small-scale scenes and require overlaps in the input images, making them difficult to apply in driving scenes with large camera movements and small parallax. Our method predicts the global scene representation instead of per-pixel Gaussian by accumulating and refining depth information in 3D space. This improves global consistency and reduces ghosting artifacts when rendering new images. VolSplat~\citep{wang2025volsplat} and Omni-Scene~\citep{wei2025omni} also adopt a volume-based architecture to predict the Gaussians but are designed for static scenes. In comparison, we leverage 3D bounding box tracks to enable feed-forward inference on 4D environments.
 
In related work, STORM~\citep{yang2024storm} and DrivingRecon~\citep{lu2024drivingrecon} perform feed-forward 4D reconstruction in driving environments but still adhere to the per-pixel Gaussian paradigm. 
The geometric inconsistencies of these methods are more pronounced in close-range regions due to stronger viewpoint-induced parallax, while far-range regions are inherently less sensitive to such misalignment. Flux4D~\citep{wang2025flux4d} initializes Gaussians from LiDAR points and employs a 3D U-Net to recover the scene, but it requires additional iterative refinement to improve background appearance and reconstruction quality. Motivated by these observations, \method
~adopts a hybrid architecture that employs a volume-based representation to ensure close-range geometric consistency, while retaining pixel-aligned regression for synthesizing high-fidelity far-field scenery.
 
\subsection{3D Supervised Feed-Forward Reconstruction}
Recent approaches in 3D vision, such as DUST3R~\citep{wang2024dust3r} and VGGT~\citep{wang2025vggt}, replace traditional multi-stage pipelines with a single large-scale model to directly regress geometric structures including camera poses and depth from images.  Follow-up works ~\citep{zhang2024monst3r,wang2025cut3r,sucar2025dynamic,jiang2025geo4d,feng2025st4rtrack,fei2024driv3r} extend this line of methods to tackle dynamic content. Despite the progress, these methods produce only point clouds, lacking the ability for high-quality view synthesis. 
To bridge this gap, some approaches ~\citep{Sun2025Uni3RU3, Zhang2025FLAREFG, Lin2025DepthA3} further integrate an additional prediction head to infer pixel-aligned 3D Gaussians, but they are still restricted to static scenes. 
Moreover, all these methods heavily rely on significant amounts of training data with 3D annotations for supervision, and such data is hard to collect for dynamic scenes, especially in the real world. In contrast, \method~achieves end-to-end reconstruction from sparse views while relying solely on readily available 2D images for supervision.

\begin{figure*}
  \centering
   \includegraphics[width=\linewidth]{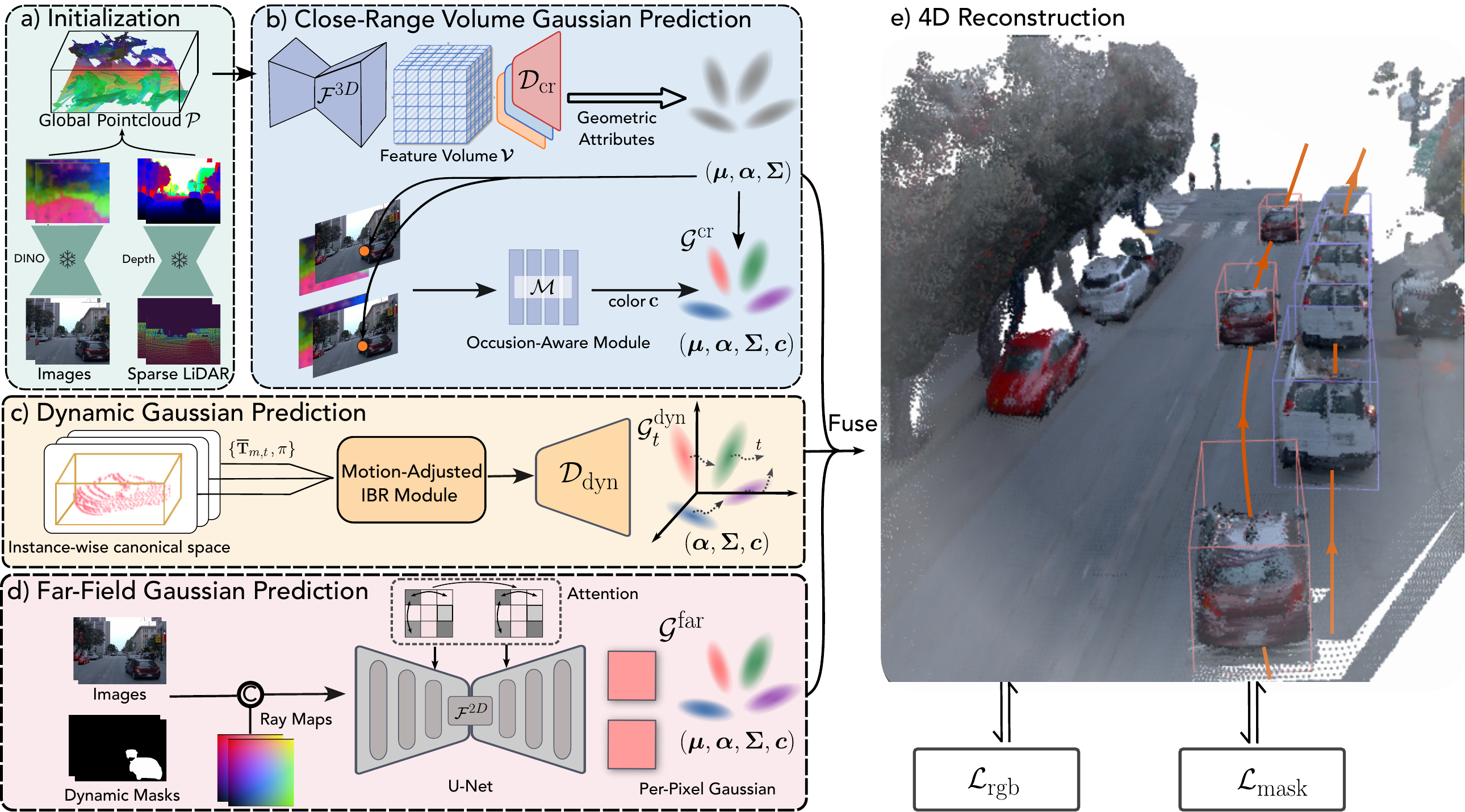}
   \caption{\textbf{Method Overview.} We reconstruct urban scenes by disentangling them as close-range volume, dynamic actors, and far-field scenery, predicting 3D Gaussians of each in a feed-forward manner. a) Given a set of images, we initialize our model with the pretrained depth model and DINO feature extractor.  b) In close-range volume, we leverage the 3D context of $\mathcal{F}^\text{3D}$ to predict the geometry attributes of 3D Gaussians and project the 3D Gaussians to the reference views to retrieve 2D context, including color window and visibility maps to decode their color. c) For dynamic actors, we model each instance using an instance-wise canonical space and perform feed-forward reconstitution through our proposed motion-adjusted IBR module. d) To model far-range regions, we employ a 2D U-Net backbone $\mathcal{F}^\text{2D}$ with cross-view self-attention to aggregate information from nearby reference images and predict per-pixel Gaussians. e) The composition of the three parts leads to our full model for unbounded scenes.  }
   \label{fig: pipeline}
   \vspace{-0.5cm}
\end{figure*}

\section{Method}
\label{METHOD}
Taking sparse images and tracked 3D bounding boxes predicted from LiDAR observations as inputs, we aim to learn a feed-forward model that directly outputs a set of 3D Gaussians for both static and dynamic regions. Our framework supports both static and dynamic street scene reconstruction, with optional fine-tuning for further improvement. An overview of ~\method~is provided in \figref{fig: pipeline}. We decompose the scene into three components: a close-range volume, dynamic vehicles, and a far-field scenery, each represented by a set of 3D Gaussians, denoted as $\mathcal{G}^{\mathrm{\text{cr}}}$, $\mathcal{G}^{\text{dyn}}$, and $\mathcal{G}^{\text{far}}$, respectively. The \textbf{close-range volume} refers to a predefined spatial region that covers the vehicle’s trajectory while excluding \textbf{dynamic actors}, whereas the \textbf{far-field scenery} corresponds to the far-range regions and sky outside this volume. 

We develop three feed-forward branches, one for each component. This decomposition is motivated by the distinct properties of 4D urban driving scenes, where close-range regions, dynamic actors, and far-field scenery require specialized representations. For the close-range volume branch, we initialize global primitives from depth priors and semantic image features, which are then transformed into an encoding volume to predict geometry of 3D Gaussian primitives, followed by an occlusion-aware IBR module for appearance parameters (\secref{sec:close-range gs}). For dynamic actors, we model each actor in a canonical space and propose a motion-adjusted IBR module to regress Gaussian attributes (\secref{sec:dynamic gaussian}). For the far-field scenery in street scenes, we train a generalizable U-Net with cross-view attention mechanisms to predict pixel-aligned Gaussians (\secref{sec: far-field scenery gaussian}). Finally, these three parts are fused to represent unbounded street scenes (\secref{sec: compositional rendering}).

\subsection{Preliminary}
\label{sec:Preliminary}
\boldparagraph{3D Gaussian Splatting} 3D Gaussian Splatting~\citep{3dgs} explicitly parameterizes the 3D radiance field of the underlying scene as a collection of 3D Gaussians primitives $\mathcal{G}=\left\{\left(\mathbf{\mu}_i,\alpha_i,\boldsymbol{\Sigma}_i, \bc_i\right)\right\}_{i=1}^G$, with attributes: a mean position $\b\mu_i$, an opacity $\alpha_i$, a covariance matrix $\boldsymbol{\Sigma_i}$ and view-dependent color $\bc_i$ (computed by spherical harmonics). The 3D covariance matrix $\bSigma\in \nR^{3 \times 3}$ of each Gaussian is defined as:

\begin{equation}
\bSigma =\b{R} \b{S} \b{S}^{\top} \b{R}^{\top}
\end{equation}
where $\bR \in \mathbb{R}^{3\times 3}$ denotes the rotation (also represented as quaternion $\mathbf{q}$ during training) and $\bS=\mathrm{diag}(s_x,s_y,s_z)$ denotes the scale. A 3D Gaussian is defined as follows:
\begin{equation}
G(\bx ) =  \alpha \exp \left(-\frac{1}{2} (\bx-\b\mu)^T \bSigma^{-1} (\bx-\b\mu) \right)
\end{equation}

This efficient representation avoids expensive volumetric sampling and enables high-speed rendering. Given a set of sorted 3D Gaussians $\mathcal{N}$ along the ray, we employ the tile-based rasterizer for Gaussian splats to pre-sort and blend the ordered primitives using differentiable volumetric rendering:

\begin{equation}
\mathbf{C}=\sum_{i \in \mathcal{N}} \bc_i \alpha_i^{\prime} \prod_{j=1}^{i-1}\left(1-\alpha_j^{\prime}\right)
\end{equation}
Here $\alpha^{\prime}$ is determined by the projected 2D Gaussian and the 3D opacity $\alpha$.

\subsection{Close-Range Volumetric Gaussian branch}
\label{sec:close-range gs}

\boldparagraph{Global Point Cloud Initialization}
\label{sec: initialization}
To overcome the limitations of per-pixel representations, our method leverages depth priors to construct a global point cloud, which serves as the foundation for decoding a set of multi-view consistent 3D Gaussians via a 3D CNN. We first unproject multi-view depth maps into a shared world coordinate system using their respective camera poses. Specifically, the depth maps $\{D_{n}\}_{n=1}^{N}$ for $N$ input images are generated from sparse LiDAR observations using an off-the-shelf depth completion model~\citep{wang2025depth}. Notably, our framework is depth-modality agnostic: our ablation studies demonstrate that both metric monocular depth predictions and sparse LiDAR completion yield comparable performance.

Regarding the attributes of this global point cloud, we opt for semantic features over raw RGB information to facilitate the prediction of refined geometry in the subsequent volumetric branch. We extract feature maps $\{F_{n}\}_{n=1}^{N}$ from the $N$ input images using DINOv2~\citep{oquab2023dinov2}. To ensure memory efficiency, Principal Component Analysis (PCA) is applied to compress these features into 16 channels. These feature maps are then lifted into 3D space and accumulated to construct a global semantic point cloud $\mathcal{P} \in \mathbb{R}^{N_p \times 16}$ in world coordinates, utilizing calibrated intrinsic matrices $\{\mathbf{K}_n\}_{n=1}^N$ and extrinsic matrices $\mathbf{T}_n = \{\mathbf{R}_n \mid \mathbf{t}_n\}_{n=1}^N$. The construction process is defined as:
\begin{equation}
\mathcal{P}=\bigcup^N \pi^{-1}(F_n, D_n, \mathbf{T}_n, \mathbf{K}_n)
\end{equation}
where $\pi^{-1}$ denotes pixel unprojection. 

Unlike previous pixel-aligned 3DGS methods that inevitably generate redundant primitives in overlapping regions, our approach avoids this issue by operating directly in Euclidean space. We apply a spatial uniform filter to the raw points $\mathcal{P}$ to explicitly prune primitives in high-density regions, thereby eliminating redundancy while preserving the underlying scene structure. Additionally, we employ a statistical filter to remove floaters. Most importantly, we introduce a learned refinement module to optimize primitive locations, as detailed in the subsequent section.

\boldparagraph{Neural Volume Prediction}
We draw inspiration from~\citep{huang2023ponder,chen2024lara} and map $\mathcal{P}$ into a 3D latent neural volume using a generalizable 3D CNN that captures prior knowledge across scenes. Specifically, we voxelize the global point cloud $\mathcal{P}$ into a sparse tensor and apply an efficient sparse U-Net-style 3D CNN $\mathcal{F}^\text{3D}$ to produce a neural feature volume $\boldsymbol{\mathcal{V}} \in \mathbb{R}^{H \times W \times D \times C}$, where $H \times W \times D$ denotes the spatial resolution and $C$ the feature dimension. The sparse 3D ConvNet adopts an encoder-decoder architecture with a low-resolution bottleneck, and skip connections are used to preserve both local and global information:
\begin{equation}
\boldsymbol{\mathcal{V}}=\mathcal{F}^\text{3D}(\mathcal{P}).
\end{equation}
Different from other feed-forward GS works using a 2D image-to-image neural network to encode the scene, this 3D latent representation helps integrate volumetric context in different network layers for faithful reconstructions. Note that the output feature volume $\boldsymbol{\mathcal{V}}$ is aligned with the world coordinate system. 

\boldparagraph{Gaussian Geometry Decoding}
We decode the geometry attributes of $N_p$ 3D Gaussians based on the feature volume $\boldsymbol{\mathcal{V}}$, taking the scene point cloud $\mathcal{P}$ as the initial Gaussian centers, i.e., $\mu_i^{init}=\bp_i \in \mathcal{P}$. 
For each $\mu_i^{init}$, we query a latent feature from $\boldsymbol{\mathcal{V}}$ using trilinear interpolation, and map it to a position offset $\Delta \mu_i$, opacity $\alpha_i$, and covariance matrix $\bSigma_i$ based on three independent MLP heads, respectively.

Considering that the network is supposed to gradually move the 3D Gaussians to the correct locations over the training process, we aim to use the updated locations $\mu_i^{init} + \Delta \mu_i$ to query the feature volume to decode $\alpha_i$ and $\bSigma_i$. This encourages the network to predict $\alpha_i$ and $\bSigma_i$ at a more rendering-consistent position, typically near the object surface. However, we do not have the direct 3D supervision of $\Delta \mu_i$. Therefore, we build an offset updating rule that learns $\Delta \mu_i$ from 2D rendering losses.

Specifically, we keep track of the network's prediction of $\Delta \mu_i$ from the last iteration as $\Delta \mu_i^{prev}$. During training, $\Delta \mu_i^{prev}$ is a cached prediction from the previous training step and is detached from the computation graph. Thus, the refinement is not unrolled, and gradients are only back-propagated through the current offset prediction. For each iteration, we obtain the feature vector $\mathbf{h}_i$ by applying the interpolation operator $\operatorname{Interp}(\cdot)$ to the feature volume $\boldsymbol{\mathcal{V}}$ at $\mu_i^{init} + \Delta \mu_i^{prev}$:

\begin{align}
   \mathbf{h}_i = &~ \operatorname{Interp}\!\left(
   \boldsymbol{\mathcal{V}}, \mu^{init}_i + \Delta \mu_i^{prev}
   \right) \label{eq:offset1} \\
   \alpha_i = & D_{\text{cr}}^{opa}(\mathbf{h_i}),~~~\bSigma_i=D_{\text{cr}}^{cov}(\mathbf{h_i}) \\
   \Delta \mu_i = & \textit{Tanh}(D_{\text{cr}}^{pos}(\mathbf{h_i}))\cdot v_\text{size}, ~~~\mu_i= \mu_i^{init}  + \Delta \mu_{i}  \label{eq:offset2}
\end{align}

where $D_{\text{cr}}^{opa}$, $D_{\text{cr}}^{cov}$ and $D_{\text{cr}}^{pos}$ denote MLP heads, and $\mu_i, \alpha_i$ and $\bSigma_i$ are subsequently used for rendering. Here, the output of $D_{\text{cr}}^{pos}$ is passed through a $\textit{Tanh}()$ activation function, and then scaled by the voxel size $v_\text{size}$ to yield the final adjustment. 
Since $\Delta \mu_i$ is predicted from features queried at the previously updated location $\mu_i^{init}+\Delta \mu_i^{prev}$, this forms a recursive update of $\Delta \mu_i$. During training, we empirically observe that this recursive update remains stable. At inference time, we initialize $\Delta \mu_i^{prev}=0$, recursively execute \eqref{eq:offset1} and \eqref{eq:offset2}, and use the final updated location to query $\alpha_i$ and $\bSigma_i$. As verified in our ablation study, two recursive updates are sufficient in practice.

As shown in our ablation experiment, the inclusion of offset allows our method to compensate for noisy positions and enhances the representational capacity.

\begin{figure}[tb]
  \centering
    \captionsetup[subfigure]{}
    \begin{subfigure}{\linewidth}
        \centering
        \includegraphics[width=1.0\linewidth]{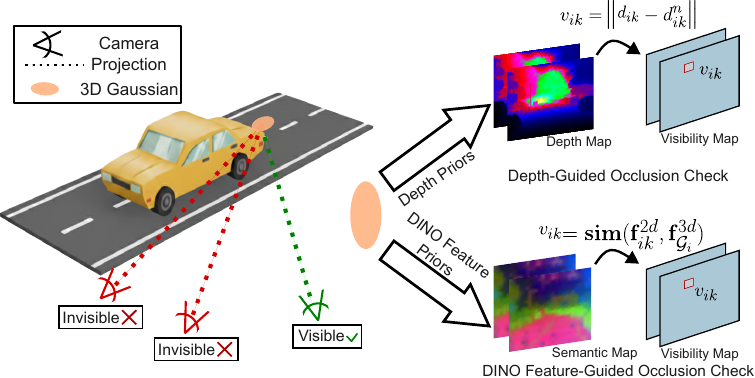}
        \caption{Occlusion Illustration. }
        \label{fig: occlusion Illustration}
    \end{subfigure}%
    \vspace{0.1cm}
   \captionsetup[subfigure]{}
    \begin{subfigure}{0.45\linewidth}
        \centering
        \includegraphics[width=\textwidth]{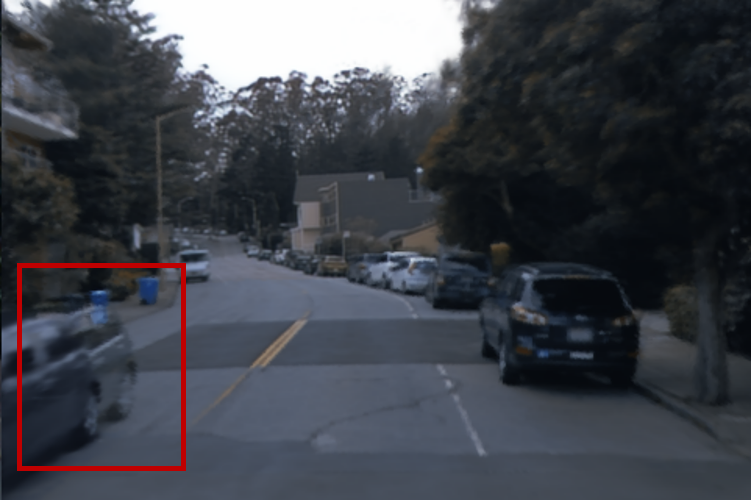}
        \caption{ Depth-Guided Check}
        \label{fig: depth check}
    \end{subfigure}%
    \hspace{0.05\linewidth}%
    \begin{subfigure}{0.45\linewidth}
        \centering
        \includegraphics[width=\textwidth]{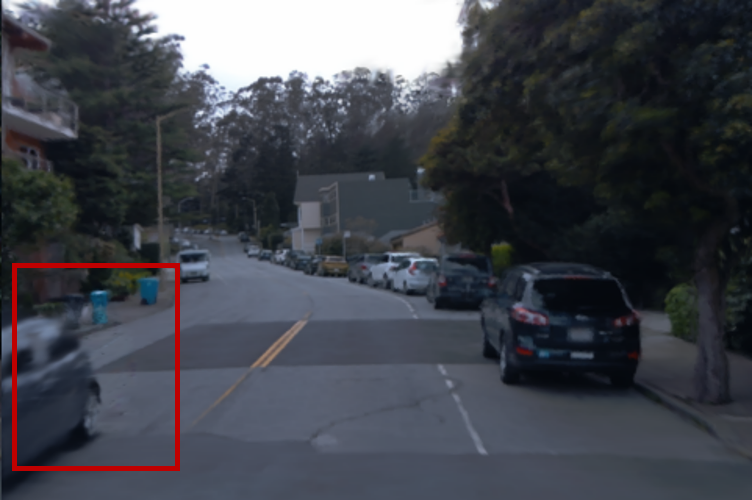}
        \caption{ DINO-Guided Check}
        \label{fig: deino check}
    \end{subfigure}
    
   \caption{\textbf{Occlusion Illustration}. a) One Gaussian in 3D space may retrieve inaccurate color information from 2D reference images due to occlusions. b) The previous method \citep{miao2025evolsplat} uses depth priors to check occlusions, which may suffer from inaccurate monocular depth predictions. c) In contrast, ~\method~comprises robust DINO priors to reduce the impact of invisible colors to enhance rendering quality. }
   \label{fig:occlusion}
   \vspace{-0.4cm}
\end{figure}

\boldparagraph{Occlusion-Aware Appearance Modeling} While our volume-based representation effectively ensures geometric consistency, the inherent downsampling of primitives may result in the loss of high-frequency appearance details. To capture these fine-grained textures, we employ an image-based rendering (IBR) approach to learn Gaussian appearance. However, two primary challenges arise in this setting: 1) imprecise initialization of Gaussian centers leads to inaccurate 2D projections, and 2) even with accurate centers, retrieved colors can be inconsistent due to significant occlusions. To address these issues, we extend the projection location into a local sampling window and propose an occlusion-aware module to robustly blend multi-view colors.

For each Gaussian primitive, we project its center $\mu_i$ to the nearby $K$ reference frames and sample local colors $\{\mathbf{c}_{ik} \in \mathbb{R}^{W \times W \times 3}\}_{k=1}^K$ using a $W \times W$ window centered at the 2D projection. This allows integrating additional 2D texture information and improves robustness against noisy 3D Gaussian locations.  In practice, we find $W=3$ to be effective (see ablation study).  

To handle occlusions, we explore two visibility check options, illustrated in \figref{fig: occlusion Illustration}: (1) depth-guided and (2) DINO feature-guided.  The depth-guided check, utilized in our conference version, leverages the retrieved monocular depth $D_n$ as a pseudo-ground-truth reference to identify occluded regions. Specifically, let $d_{ik}$ denote the projected depth of Gaussian center $\mu_i$ from the $k$-th camera viewpoint. We estimate visibility based on the residual between the projected depth and the retrieved depth $d_{ik}^{n}$ from $D_n$ via $v_{ik} = \|d_{ik} - d_{ik}^{n}\|$. However, this method remains sensitive to inaccuracies in monocular depth estimation within complex environments (see Fig.~\ref{fig: depth check}). 
In contrast, our DINO feature-guided check relies on the intuition that a 3D Gaussian and its unoccluded 2D projection should maintain semantic similarity, thereby removing the dependency on potentially noisy depth priors. We compute the cosine similarity $\text{sim}(\cdot,\cdot)$ between the unprojected 3D feature $\mathbf{f}_{i}^{3d}$ stored in $\mathcal{P}$ and the retrieved 2D features $\mathbf{f}_{ik}^{2d}$ from the $k$-th reference frame. The final visibility $v_{ik}$ is then obtained via a temperature-scaled Softmax normalized over all $K$ reference frames:
\begin{equation}
v_{ik} = \frac{
\exp\!\left(\textbf{sim}(\mathbf{f}_{ik}^{2d}, \mathbf{f}_{i}^{3d}) / \tau \right)
}{
\sum_{k'=1}^{K} \exp\!\left(\textbf{sim}(\mathbf{f}_{ik'}^{2d}, \mathbf{f}_{i}^{3d}) / \tau \right)
}, 
\end{equation}
where $\tau=0.1$ in our experiments. This approach effectively identifies inconsistent observations due to occlusion and focuses on visible views (see \figref{fig: deino check}).  

Instead of discarding invisible projections with a manual threshold, we concatenate the local window colors and their visibility and feed them into a three-layer MLP $\mathcal{M}$, which outputs the color represented using spherical harmonics (SH) coefficients:
\begin{equation}
\mathbf{c}_i = \mathcal{M}(\{\mathbf{c}_{ik}, v_{ik}\}_{k=1}^K).
\end{equation}

\subsection{Generalizable Dynamic Gaussian branch}
\label{sec:dynamic gaussian}

\boldparagraph{Dynamic Gaussian Modeling}
To enable flexible control and model the movements of dynamic instances (\eg cars, trucks) without sacrificing reconstruction quality, we model each instance using an instance-wise canonical space and time-varying rigid transformation. This formulation ensures that each moving instance is represented independently in a static canonical space across all frames, facilitating information accumulation.  
For a scene containing $M$ moving actors, we utilize an off-the-shelf detector and tracker to generate 3D bounding boxes and temporal poses $\overline{\mathbf{T}}_{m,t} \in \operatorname{SE}(3)$ for each actor $m \in\{1, \dots, M\}$. Specifically, we use the official implementation and checkpoint of Casa~\citep{wu2022casa} for LiDAR-based 3D object detection, and PC3T~\citep{wu20213d} for temporal association and tracking. Casa predicts each object with a 3D bounding box parameterized by its center, size, and heading angle. PC3T then associates the detected boxes across frames and assigns consistent track IDs, yielding temporal object trajectories. All main experiments use predicted 3D boxes and tracks from LiDAR observations rather than ground-truth object annotations, unless explicitly stated in an ablation. While the tracking is performed on LiDAR observations following the common practice of dynamic urban reconstruction methods~\citep{streetgaussian,chen2025s,tonderski2024neurad}, the 3D points within these bounding boxes can be derived from either LiDAR observations or monocular depth predictions. By leveraging these tracked poses, we transform the 3D points within the bounding boxes from all frames into a unified local object coordinate system and aggregate them to form a semantic point cloud with DINO features $\overline{\mathcal{P}}_{m}^\text{dyn} \in \mathbb{R}^{N_d \times 16}$ in the canonical space. This canonical representation captures the time-invariant geometry of each rigid actor. At a target time $t$, the points are transformed back into world space by applying the corresponding pose transformation:
\begin{equation}
{\mathcal{P}_\text{m,t}^\text{dyn} = \overline{\mathbf{T}}_{m,t} \overline{\mathcal{P}}^\text{dyn}_{m}} \label{eq:rigid_transform}
\end{equation}

\boldparagraph{Motion-Adjusted Image-Based Rendering}
Unlike our static branch, which employs a 3D CNN to refine geometry through learned position offsets, we utilize the accumulated canonical point locations directly as the Gaussian centers $\mu$ for dynamic actors. We find that while thin structures or complex topologies in the static close-range volume benefit more from learned geometric refinement, dynamic vehicles possess more regularized geometries where multi-frame aggregation in a constrained canonical space provides a sufficiently stable foundation. Consequently, we task our motion-adjusted IBR module to predict all other Gaussian attributes, including opacity $\alpha$, covariance $\mathbf{\Sigma}$, and color $\mathbf{c}$. To effectively compensate for any remaining inaccuracies in $\b\mu$ or the tracked bounding boxes, we also leverage a window-based IBR design.

\begin{figure}
  \centering
   \includegraphics[width=0.98\linewidth]{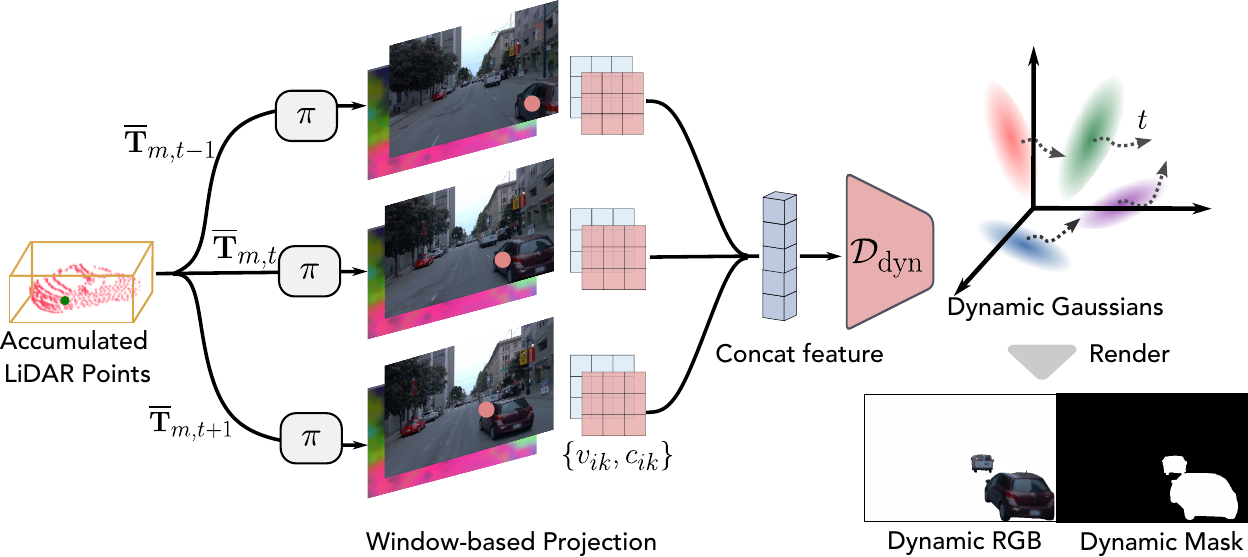}
   \caption{\textbf{Motion Adjusted IBR.} LiDAR points in the canonical space for an instance $m$ are transformed using time-specific poses and projected into reference frames. A Window-based Projection strategy then samples coherent appearance features $c_{ik}$ and visibility map $v_{ik}$, which are input to the dynamic Gaussian decoder $D_\text{dyn}$. The decoder regresses the 3D Gaussian attributes for each point, enabling the final rendering of the dynamic object's image and binary mask.}
   \label{fig:dynibr}
\end{figure}

Specifically, our motion-adjusted IBR accounts for rigid object motion, as illustrated in Fig.~\ref{fig:dynibr}. Taking a triplet $\{t-1, t, t+1\}$ and their corresponding rigid transformations $\overline{\mathbf{T}}_{m, t-1}$, $\overline{\mathbf{T}}_{m, t}$, and $\overline{\mathbf{T}}_{m, t+1}$ as an example, we apply \eqref{eq:rigid_transform} to place the canonical points into the world space corresponding to the precise timestamp of each captured frame. Subsequently, the projection operator $\pi$ maps these time-adjusted 3D points onto the respective 2D reference image planes to query color cues. This motion-adjusted projection ensures that each dynamic point gathers coherent appearance information across multiple timestamps.

Similar to \secref{sec:close-range gs}, we retrieve appearance cues $\{\mathbf{c}_{ik}\}_{k=1}^K$ and visibility terms $\{v_{ik}\}_{k=1}^K$ from DINO features within a predefined window to enhance robustness against noisy predicted 3D boxes. Our ablation study demonstrates that this window-based sampling strategy achieves performance comparable to using ground-truth bounding boxes. The aggregated 2D features are then processed by a shared three-layer MLP for all $M$ moving actors, which serves as a dynamic Gaussian decoder $D_\text{dyn}$ to predict the time-dependent Gaussian attributes $\mathcal{G}_{t}^\text{dyn}=\{(\alpha_i,\boldsymbol{\Sigma}_i, \mathbf{c}_i)\}_{i=1}^{N_d}$.

\subsection{Generalizable Far-Field Branch}
\label{sec: far-field scenery gaussian}
As our volume-based representation covers only limited close-range street scenes, it is insufficient for the unbounded region, such as the sky and distant landscapes (often over 100 meters away). Previous works~\citep{chen2023periodic,ye2024gaustudio,xu2024splatfacto,urbanradiancefield,streetgaussian} construct an environment map conditioned on input direction for a special scene, yet struggle to handle multiple scenes simultaneously. To this end, we introduce our generalizable far-field branch tailored for driving scenes to model infinite sky and background landscapes.

Inspired by LGM~\citep{tang2024lgm}, we employ a 2D U-Net backbone $\mathcal{F}^\text{2D}$ with cross-view self-attention to aggregate information from nearby reference images and predict per-pixel Gaussians. The input to the  $\mathcal{F}^\text{2D}$ is a 10-channel feature map, where each feature vector $\mathbf{e}_i$ is formed by concatenating its RGB color $\mathrm{c}_i$, a binary mask $b_i$, and its Plücker ray embeddings:
\begin{equation}
\mathbf{e}_i=\left\{\mathrm{c}_i, b_i, \mathbf{o}_i \times \mathbf{r}_i, \mathbf{r}_i\right\}
\end{equation}
Here, $\mathbf{o}_i$ and $\mathbf{r}_i$ denote the ray origin and direction, respectively. The binary mask $b_i$, provided by our dynamic Gaussian branch (\secref{sec:dynamic gaussian}), is utilized to suppress interference from dynamic actors, preventing them from being erroneously baked into the static background representation.

Within the U-Net architecture, self-attention is restricted to the two deepest layers to maintain computational efficiency while capturing long-range dependencies. Each pixel of the output feature map is interpreted as a 3D Gaussian primitive with 12 channels representing RGB color $\mathbf{c}_i \in \mathbb{R}^{3}$, scale $\mathbf{s}_i \in \mathbb{R}^{3}$, a rotation quaternion $\mathbf{q}_i \in \mathbb{R}^{4}$, opacity $\alpha_i \in \mathbb{R}$, and a 1-channel ray distance $z_i \in \mathbb{R}$. The Gaussian center is defined as $\mathbf{\mu}_i = \mathbf{o}_i + z_i \mathbf{r}_i$. The final far-field representation $\mathcal{G}^\text{far}$ is obtained by merging the Gaussians predicted from all $K$ views. Although this model effectively captures far-range regions, our ablation study demonstrates that it is insufficient for modeling close-range structures with the same fidelity as our volume-based representation.

\subsection{Compositional Rendering}
\label{sec: compositional rendering}
The integration of these three specialized components, the close-range volume, dynamic actors, and the far-field scenery, constitutes our comprehensive model for unbounded street scenes. To synthesize the scene from a target viewpoint at time $t$, we unify these primitives into a single representation: $\mathcal{G}_{t}^{\text{scene}} = \mathcal{G}^{\text{cr}} \cup \mathcal{G}_{t}^{\text{dyn}} \cup \mathcal{G}^{\text{far}}$. These composite Gaussians are rendered onto the 2D image plane using standard 3D Gaussian Splatting rasterization. Specifically, $\alpha$-blending is first employed to accumulate color and opacity for the combined close-range volume and dynamic actors:
\begin{equation}
\mathbf{C}^{(\text{cr}+\text{dyn})}=\sum_{i \in \mathcal{G}^{\text{cr}},\mathcal{G}_{t}^{\text{dyn}}}  \bc_i \alpha_i \prod_{j=1}^{i-1}\left(1-\alpha_j\right)
\end{equation}
\begin{equation}
\quad \mathbf{O}^{(\text{cr}+\text{dyn})}=\sum_{i \in \mathcal{G}^{\text{cr}},\mathcal{G}_{t}^{\text{dyn}}}\alpha_i \prod_{j=1}^{i-1}\left(1-\alpha_j\right)
\end{equation}

Finally, we integrate the far-field rendering $\mathbf{C}_{\text{far}}$ as a background layer to reconstruct the full scene:
\begin{equation}
\mathbf{C}=\mathbf{C}^{(\text{cr}+\text{dyn})}+\left(1-\mathbf{O}^{(\text{cr}+\text{dyn})}\right) \mathbf{C}_{\text {far}}
\end{equation}

\section{Implementation Details}

\subsection{Loss Function}
Our model is trained in an end-to-end manner, jointly optimizing the scene representation across various driving scenarios. The total loss function is defined as:
\begin{equation}
\mathcal{L}=\mathcal{L}_{\mathrm{rgb}}+\lambda_{m}\mathcal{L}_{\text {mask}}
\end{equation}
where $\lambda_{m}$ is a balancing weight. In the following, we elaborate on the individual loss components.

\boldparagraph{Photometric loss}
We apply L1 and SSIM losses between a rendered image $\mathbf{I}$ and the GT image $\hat{\mathbf{I}}$ for supervision of RGB rendering. The balance term $\lambda_\mathrm{SSIM}$ is set to 0.2 in our setting.
\begin{equation}
\mathcal{L}_{\mathrm{rgb}}=\left(1-\lambda_\mathrm{SSIM}\right)\|\mathbf{I}-\hat{\mathbf{I}}\|_1+\lambda_\mathrm{SSIM} \operatorname{SSIM}(\mathbf{I}, \hat{\mathbf{I}})
\end{equation}

\boldparagraph{Decomposition loss}
A primary challenge in our hybrid representation is ensuring that the close-range volume ($\mathcal{G}^\text{cr}$) and the far-field background ($\mathcal{G}^\text{far}$) exclusively model their respective portions of the scene. To prevent geometric ambiguity, we introduce a mask loss $\mathcal{L}_{\text{mask}}$ to enforce explicit spatial separation between these components:
\begin{equation}
\mathcal{L}_{\text{mask}} = \|\mathbf{O}^{(\text{cr})} - \hat{\mathbf{O}}\|_1
\end{equation}
The binary mask $\hat{\mathbf{O}}$ is generated by projecting the initial global point cloud $\mathcal{P}$ onto the image plane, where pixels corresponding to the close-range volume are assigned a value of 1 and otherwise set to 0. This supervision encourages the close-range branch to maintain high opacity in its designated region while penalizing the far-field branch for erroneously rendering content within the foreground volume.

\baselineCategorize

\subsection{Fine-Tuning}
Once fine-tuning is applied, our method can achieve photorealism on par with or surpassing other per-scene optimization methods, leveraging powerful pretrained weights. Specifically, we first predict a set of 3D Gaussian primitives for initialization via a direct feed-forward process, where both the geometry and color attributes are generated. After this feed-forward initialization, the network is no longer queried; all scene components are represented by explicit Gaussian sets for optimization.
The fine-tuning process follows the vanilla 3DGS, where we optimize all Gaussian attributes and refine tracked bounding boxes directly. We also enable the growing and pruning of 3D primitives during fine-tuning. During fine-tuning, the number of Gaussians is significantly reduced, as the initial feed-forward Gaussians are redundant. As a result, our model maintains high fidelity while reducing memory consumption. We demonstrate that the pretrained priors accelerate network training and enable faster convergence compared to other test-time optimization methods.

\subsection{Training Details}
We train~\method~end-to-end on multiple scenes using the Adam optimizer~\citep{kingma2014adam} with learning rate $1 \times 10^{-3}$. 
We use torchsparse~\citep{tang2022torchsparse} as the implementation of 3D sparse convolution and choose gsplat~\citep{ye2024gsplatopensourcelibrarygaussian} as our Gaussian rasterization library. We set the SH degree to 1 for simplicity, following StreetGaussian~\citep{streetgaussian}. During training, we set $\lambda_m = 0.1$ as loss weights in our work. For each iteration, our model randomly selects a single image from a random scene as supervision.

\section{Experiment}
\label{sec:experiment}
\begin{table*}[ht]
\centering
\renewcommand{\arraystretch}{1.2}

\resizebox{\textwidth}{!}{%
\begin{tabular}{@{}l|ccc|ccc|cc@{}}
\toprule
\multirow{2}{*}{Method} & \multicolumn{3}{c|}{KITTI-$360$ (In-Domain)} & \multicolumn{3}{c|}{Waymo (Out-of-Domain) } & \multirow{1}{*}{FPS$\uparrow$} & \multirow{1}{*}{Mem.(GB)$\downarrow$}\\
             & PSNR$\uparrow$  & SSIM$\uparrow$ & LPIPS$\downarrow$ & PSNR$\uparrow$ & SSIM$\uparrow$    & LPIPS$\downarrow$ & \multicolumn{2}{c}{(Render res.: 376$\times$1408)}\\ \midrule
MVSNeRF~\citep{mvsnerf}  &18.44 &0.638  &0.317 &17.86 &0.595 &0.433 &0.025 &12.03 \\
MuRF~\citep{xu2023murf}   &22.77 &0.780  &0.229 &23.33 &0.770 &0.269 &0.31 &26.45 \\
EDUS~\citep{miao2024efficient}   & 22.13  & 0.745  & 0.178 &23.18  &0.745 &\textbf{0.164} &0.14 & 5.75   \\ 
\midrule
PixelSplat~\citep{charatan2024pixelsplat}   &19.41   &0.584  &0.357  & 16.65 &0.541  &0.579 &81.58 &26.75 \\
MVSplat~\citep{chen2024mvsplat}    &21.22  &0.695 &0.268 & 21.33  & 0.665  &0.308 &70.46 &16.14 \\
DrivingForward~\citep{tian2024drivingforward} & 21.44 & 0.746 & 0.293 & 21.78 &0.721 & 0.367 & 74.59 & 12.85 \\
DepthSplat~\citep{xu2025depthsplat}         &21.44&  0.687&  0.265&  20.95&  0.664& 0.400& 76.24& 17.62\\
AnySplat~\citep{Jiang2025AnySplatF3}         &  20.09& 0.629&0.264  &\textbf{24.58}  &0.784  &0.193  &79.36 &22.05 \\
\midrule
EVolSplat          & 23.26 & 0.797 &0.179  &23.43  &0.786  &0.202 &\textbf{83.81} &\textbf{10.41}\\
EVolSplat4D          &\textbf{23.36}&  \textbf{0.798}&\textbf{0.177}  &24.43  &\textbf{0.829}  &0.184 &82.46 &10.98\\
\bottomrule
\end{tabular}
}
\caption{\textbf{Quantitative results on Static Scenes} with other feed-forward baselines on both KITTI-360 and Waymo open datasets. All models are trained on the KITTI-360 dataset using a drop50\% sparsity level. Metrics are averaged on five validation scenes without any finetuning. It is also worth noting that our method is more memory efficient compared with other 3DGS-based methods.}
\label{tab: static table}
\end{table*}

\figStaticComparison
\subsection{Dataset}
\boldparagraph{Static Scenes}
 We pretrain our static model on the KITTI-360 dataset~\citep{liao2022kitti}. Specifically, we begin with 160 sequences from the KITTI-360 (same as in our conference version \citep{miao2025evolsplat}) to learn robust representations for static scenes reconstruction. To simulate sparse driving scenarios, we apply a \textbf{50\% drop rate} (Drop 50\%) during all static training and validation. Additionally, we evaluate the zero-shot generalization performance on Waymo~\citep{waymo}. 
 
\boldparagraph{Dynamic Scenes}
Since KITTI-360 scenarios are primarily static, we supplement them with 181 motion-rich sequences from the Waymo~\citep{waymo} and KITTI~\citep{kitti} to improve generalization to dynamics. This results in a total of 341 training sequences, each containing 60 images. For validation, we use a total of 10 clips as an in-domain dataset, consisting of 5 sequences from the KITTI validation set and 5 from Waymo NOTR~\citep{emernerf}, ensuring no overlap with the training data. Furthermore, we utilize PandaSet~\citep{xiao2021pandaset} as an out-of-domain dataset to evaluate the zero-shot generalization performance in dynamic environments. We follow the experimental setup from STORM~\citep{yang2024storm} and use a more sparse \textbf{80\% drop rate} (Drop 80\%) for all sequences, which enables us 
to assess the robustness of our model to varying sparsity levels. As we find that our dynamic baseline methods suffer from high memory cost when training with high-resolution images, we downsample input images for all methods: $4\times$ for Waymo and PandaSet and $2\times$ for KITTI and KITTI-360.

\subsection{Experimental Setup}
\boldparagraph{Metrics} We evaluate novel view synthesis on both interpolated and extrapolated views. Interpolated-view evaluation is used as a supervised protocol to measure whether the model can recover accurate geometry and appearance from sparse observations, while extrapolated lane-shift evaluation is used to assess trajectory modification and re-simulation capability. For interpolated evaluation, we adopt existing evaluation protocols, including PSNR, SSIM, and LPIPS, for quantitative assessments. For the extrapolated views, quantitative metrics like PSNR that require ground-truth images cannot be computed. Instead, we use the Kernel Inception Distance (KID)~\citep{binkowski2018demystifying} to measure the similarity in distribution between extrapolated views and real captured images. Regarding the computation efficiency, we report the inference memory usage (GB) and frame per second (FPS) on the same device (NVIDIA RTX 5880) to ensure a fair comparison.

\boldparagraph{Volume Setup}
For the Close-Range Volume, we set its Z-axis to cover the vehicle’s forward trajectory with a length of $80\mathrm{m}$, whereas the height (Y-axis) and width (X-axis) is $12.8\mathrm{m}$ and $40\mathrm{m}$, respectively. 
The input point cloud is voxelized with a voxel size of $(0.1\mathrm{m}, 0.1\mathrm{m}, 0.1\mathrm{m})$, resulting in the volume dimension of $128 \times 400 \times 800$. Notably, our 3D CNN effectively handles arbitrary volume size well during inference.

\boldparagraph{Baselines}
We compare~\method~with several representative feed-forward methods across both static and dynamic scenes. We explicitly group the baselines according to task scope. \tabref{tab:baseline_group} summarizes representative methods by sensor inputs, 3D box track usage, representation type, feed-forward capability, dynamic-scene support, real-time rendering, and scene-editing capability. This categorization clarifies that static feed-forward methods, dynamic feed-forward methods, and per-scene optimization methods target different settings. For static scenes, we evaluate against NeRF-based approaches including MVSNeRF~\citep{mvsnerf}, MuRF~\citep{xu2023murf}, EDUS~\citep{miao2024efficient}, and 3DGS-based approaches, including
 MVSplat~\citep{chen2024mvsplat}, PixelSplat~\citep{charatan2024pixelsplat}, DepthSplat~\citep{xu2025depthsplat}, AnySplat~\citep{Jiang2025AnySplatF3}, and DrivingForward~\citep{tian2024drivingforward}. In dynamic scenes, we compare with recent feed-forward 4DGS methods like STORM~\citep{yang2024storm} and DrivingRecon~\citep{lu2024drivingrecon}. Unless stated otherwise, we train and evaluate all feed-forward models using the same data as \method. The only exception is AnySplat~\citep{Jiang2025AnySplatF3}, which builds upon VGGT~\citep{wang2025vggt} and is pre-trained on large indoor and outdoor datasets. Therefore, we directly use its official checkpoint without retraining. Notably, since AnySplat cannot accept pose inputs, we follow its protocol to estimate poses for the test views before evaluation. Additionally, we compare the NVS quality and reconstruction time with other per-scene optimization methods designed for urban scenes, including SUDS~\citep{turki2023suds}, StreetGaussian~\citep{streetgaussian}, OmniRe~\citep{chen2024omnire}, EmerNeRF~\citep {emernerf} and DeSiRe-GS~\citep{peng2025desire}.

\begin{table*}[!t]
    \centering
    \renewcommand{\arraystretch}{1.2} %
\resizebox{\textwidth}{!}{%
    \begin{tabular}{@{} l | ccc | ccc | ccc | cc @{}}
    \toprule
    \multirow{2}{*}{Method} & \multicolumn{3}{c|}{KITTI (In-Domain)} & \multicolumn{3}{c|}{Waymo (In-Domain)} & \multicolumn{3}{c|}{PandaSet (Out-of-Domain)} & \multirow{1}{*}{FPS$\uparrow$} & \multirow{1}{*}{Mem.(GB)$\downarrow$} \\
                          & PSNR$\uparrow$ & SSIM$\uparrow$ & LPIPS$\downarrow$ & PSNR$\uparrow$ & SSIM$\uparrow$ & LPIPS$\downarrow$ & PSNR$\uparrow$ & SSIM$\uparrow$ & LPIPS$\downarrow$ & \multicolumn{2}{c}{(Render res.: 320$\times$480}) \\ \midrule
    
    DrivingRecon~\citep{lu2024drivingrecon} & 19.50       & 0.597       & 0.283       & 20.21       & 0.572       & 0.277       & 14.92       & 0.506       & 0.389 & 190.78 & 9.48      \\
    STORM~\citep{yang2024storm}  &18.96 &0.658 &0.256   &23.76  &0.764 &0.182 &23.13 &0.766   &0.207     & 178.57 & 22.31       \\
    \midrule
    EVolSplat      & 16.52   &0.583 &0.275  &22.86 &0.753 &0.201  &23.32   & 0.746    &0.253   &\textbf{212.42} & \textbf{6.41}     \\
    EVolSplat4D     &\textbf{20.76 }    &\textbf{0.738}     &\textbf{0.162}      &\textbf{26.32} &\textbf{0.822}     &\textbf{0.121}             &\textbf{26.27}         &\textbf{0.823}    &\textbf{0.132}  &201.06 &6.71           \\
    \bottomrule
    \end{tabular}
 } %
 \caption{\textbf{Quantitative results on Dynamic Scenes} with feed-forward baselines.  Metrics are averaged on five validation scenes without any finetuning in drop80\% sparsity level. Our \method~generalizes better than all the baseline in a large margin on both \textbf{in-domain} (KITTI, Waymo) and \textbf{out-of-domain} (PandaSet) datasets.}
\label{tab:dynamic table}
\end{table*}

\begin{figure*}[t]
    \centering
    \renewcommand{\arraystretch}{0}
    \setlength{\tabcolsep}{1pt}
    \def\mywidth{6.2cm}
    \resizebox{\textwidth}{!}{%
        \begin{tabular}{@{} P{0.5cm} P{\mywidth}P{\mywidth}P{\mywidth} @{}}
        
        \rotatebox{90}{\textbf{DrivingRecon}} & 
        \includegraphics[width=\linewidth]{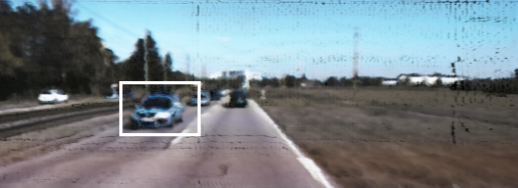} & 
        \includegraphics[width=\linewidth]{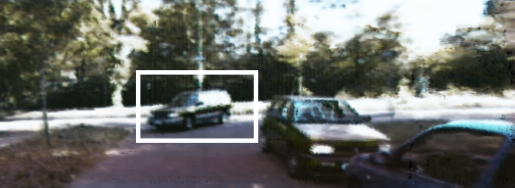} & 
        \includegraphics[width=\linewidth]{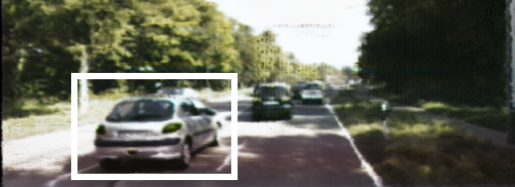} \\

        \noalign{\vspace{1pt}}

        \rotatebox{90}{\textbf{STORM}} & 
        \includegraphics[width=\linewidth]{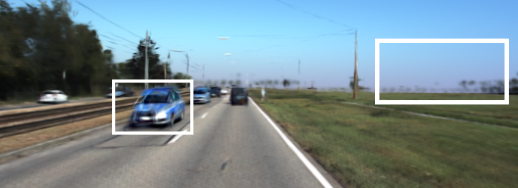} & 
        \includegraphics[width=\linewidth]{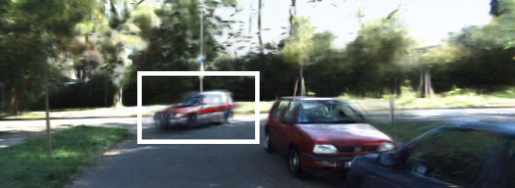} & 
        \includegraphics[width=\linewidth]{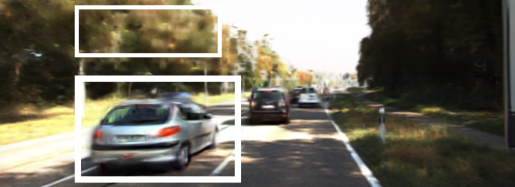} \\
        \noalign{\vspace{1pt}}
        \rotatebox{90}{\textbf{Ours}} & 
        \includegraphics[width=\linewidth]{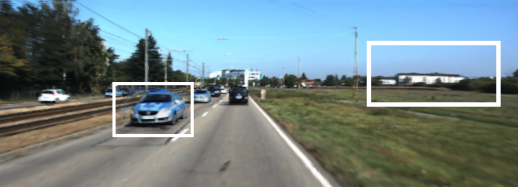} & 
        \includegraphics[width=\linewidth]{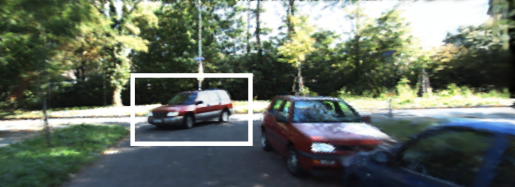} & 
        \includegraphics[width=\linewidth]{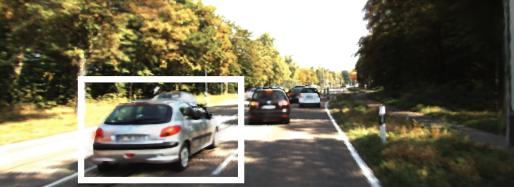} \\
        \noalign{\vspace{1pt}}
        
        \rotatebox{90}{\textbf{GT}} & 
        \includegraphics[width=\linewidth]{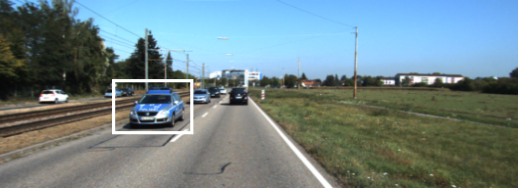} & 
        \includegraphics[width=\linewidth]{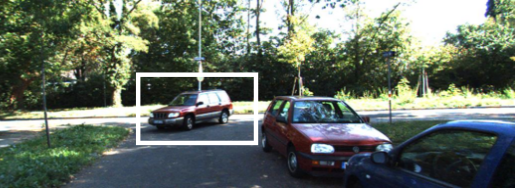} & 
        \includegraphics[width=\linewidth]{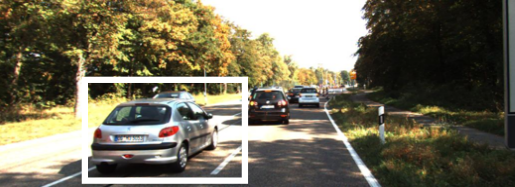} \\
        
        \end{tabular}%
    }
    
    \caption{\textbf{Qualitative Comparisons} with baselines for feed-forward inference on the KITTI dataset. The rendering images are downscaled 2x to 188$\times$621.}
    \label{fig:dynamic kitti} 
\end{figure*}

\figDynamicComparison
  
\subsection{Evaluation on Static Scenes}

\boldparagraph{In-Domain Rendering Quality}
\tabref{tab: static table} and \figref{fig:static fig} present quantitative and qualitative comparison results for feed-forward inference in static scenes of KITTI-360, evaluated under a 50\% drop rate. Our proposed EVolSplat4D achieves superior photorealism on this in-domain setting compared to state-of-the-art feed-forward NeRF- and 3DGS-based approaches. While EDUS and MuRF show promising metrics, EDUS suffers from artifacts in thin structures, and MuRF struggles with geometry in texture-less areas, such as roads, due to its reliance on feature mapping. MVSplat and DepthSplat utilize Transformers for per-frame depth estimation, which can result in inconsistent 3D Gaussian primitives and ghost artifacts under large camera movements. For AnySplat, which operates in a zero-shot setting here as it was pre-trained on separate large-scale datasets, the sparse views on KITTI-360 lead to inaccurate geometric reconstruction and artifacts in far-field background regions. DrivingForward suffers from artifacts in thin structures under large camera movements, whereas EVolSplat4D achieves better reconstruction quality and improves PSNR from 21.44 to 23.36 on KITTI-360 and from 21.78 to 24.43 on Waymo.

Compared with our conference version, EVolSplat, we replace the hemisphere background model with our Generalizable Far-Field Branch, which improves background fidelity and final results. It is worth noting that while our unified model utilizes LiDAR priors to initialize the close-range volume branch, our depth ablation in Table~\ref{tab:depth_sensitivity} demonstrates that replacing these with monocular metric depth (e.g., Metric3D) yields comparable performance. This confirms that the superiority of EVolSplat4D stems from our architectural design rather than the sensor modality.

\boldparagraph{Zero-Shot Inference on Waymo} 
To further verify the generalization capability of the evaluated models, we conduct experiments on an unseen dataset, the Waymo Open dataset~\citep{waymo}. This evaluation represents an out-of-domain setting for all methods. Specifically, all models except for AnySplat are trained exclusively on KITTI-360. AnySplat is uniquely advantaged in this comparison as it is pre-trained on a broader collection of large-scale indoor and outdoor scenes. During evaluation, we set the image resolution to $640 \times 960$ and assess the resulting rendering quality. As demonstrated in ~\tabref{tab: static table}, our method achieves superior performance across most benchmarks. While AnySplat reports a slightly higher PSNR on Waymo than on KITTI-360, this is largely due to Waymo's higher capture frequency: the dataset maintains larger view overlap than KITTI-360, facilitating easier reconstruction. Nevertheless, EVolSplat4D outperforms AnySplat in terms of SSIM and LPIPS despite using much less training data, demonstrating strong generalization capability and superior perceptual quality. Finally, we observe that EDUS yields the lowest LPIPS on Waymo. This can be attributed to its NeRF-based architecture, which often excels at modeling fine-grained details compared to splatting-based approaches.
More qualitative results are provided in the supplementary material.

\boldparagraph{Model Efficiency}
We report the memory consumption during inference on a full-resolution $376\times1408$ image of KITTI-360 in \tabref{tab: static table}. MuRF and PixelSplat fail to infer with full-resolution on a consumer GPU (24GB) since their transformer architecture requires heavy computational resources. In contrast, our method leverages an efficient sparse 3DCNN for feed-forward reconstruction with only 10.98GB usage, demonstrating superior memory efficiency and practical utility.  Additionally, we note that EDUS consumes 5.75GB of memory for inference as its NeRF-based representation only casts 4096 rays each iteration,  resulting in lower memory usage but significantly slow rendering speed (0.14fps). We further provide a scaling analysis of memory usage and inference time with respect to the number of input frames and image resolution in the supplementary material.

\subsection{Evaluation on Dynamic Scenes}
\boldparagraph{In-Domain Rendering Quality}
\tabref{tab:dynamic table}, \figref{fig:dynamic kitti}, and \figref{fig:dynamic fig} present quantitative and qualitative comparison results for in-domain feed-forward inference on dynamic scenes. To ensure a fair comparison, all dynamic feed-forward baselines are trained on a unified dataset comprised of sequences from KITTI~\citep{kitti}, KITTI-360~\citep{liao2022kitti}, and Waymo~\citep{waymo}. Note that our model achieves the best performance among all the baselines. Specifically, DrivingRecon~\citep{lu2024drivingrecon} employs temporal cross-attention to generate per-pixel Gaussians but struggles to reliably regress Gaussian attributes and motion across timestamps, leading to suboptimal rendering quality. STORM~\citep{yang2024storm} leverages self-supervised learning to decompose scene dynamics and predict scene flow, yet still produces noticeable artifacts in dynamic regions. Although EVolSplat performs well on static scenes, it lacks dedicated mechanisms for modeling dynamic content, resulting in significant motion blur.

\boldparagraph{Zero-Shot Inference on PandaSet}
We further evaluate the zero-shot generalization of all feed-forward baselines on PandaSet, with results reported in \tabref{tab:dynamic table}. On this out-of-domain dataset, our method consistently outperforms prior baselines across all visual metrics by a large margin, demonstrating strong robustness and generalization. The qualitative results in \figref{fig:dynamic fig} further support this conclusion: DrivingRecon still produces blurry results, and STORM sometimes fails to capture distant buildings in PandaSet.

\begin{figure}[t]
    \centering
    \includegraphics[width=0.49\columnwidth]{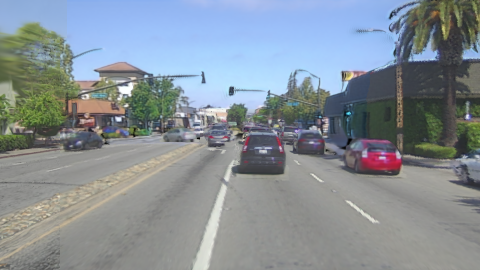}
    \hfill %
    \includegraphics[width=0.49\columnwidth]{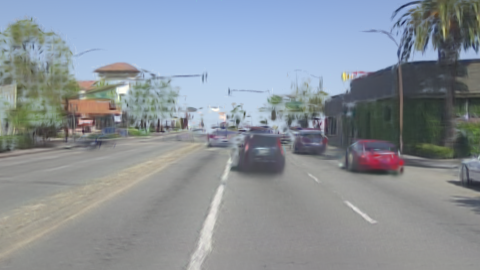}
    \\[2pt] %

    \includegraphics[width=0.49\columnwidth]{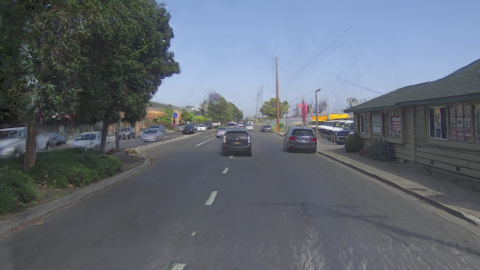}
    \hfill
    \includegraphics[width=0.49\columnwidth]{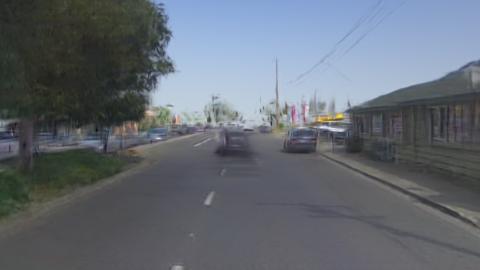}
    \\[1pt] %

    \makebox[0.49\columnwidth][c]{\small Ours}
    \hfill
    \makebox[0.49\columnwidth][c]{\small STORM}

    \caption{\textbf{Extrapolated Views Qualitative Comparison} with STORM on PandaSet.}
    \label{fig:extrapolate_res}
    \vspace{-0.4cm}
\end{figure}

We further evaluate view extrapolation quality by reporting the KID score under a 1 m lane-shift setting, where no ground-truth images are available (see \tabref{tab:kid_comparison}). Our method achieves superior performance in this challenging extrapolation scenario. As shown in \figref{fig:extrapolate_res}, EVolSplat4D produces high-fidelity images with clear lane markings, avoiding the distortion and blur observed in STORM.
\begin{table}[t] %
    \centering
    \renewcommand{\arraystretch}{1.2} %

    \resizebox{\linewidth}{!}{%
        \begin{tabular}{@{} l ccc @{}}
        \toprule
        Method & KITTI & Waymo & PandaSet \\
        \midrule 
        DriveRecon~\citep{lu2024drivingrecon}  & 0.350  & 0.315  & 0.383 \\
        STORM~\citep{yang2024storm}    & 0.104    & 0.080  & 0.102 \\
        EVolSplat4D (Ours)           & \textbf{0.062}&  \textbf{0.063}&  \textbf{0.080} \\
        \bottomrule
        \end{tabular}%
    }
    \caption{\textbf{Quantitative Comparison of View Extrapolation Quality}. We report averaged KID $\downarrow$ (lower values indicate better performance) on diverse datasets without finetuning in Drop 80\% sparsity level.}
    \label{tab:kid_comparison}
    \vspace{-0.2cm}
\end{table}
  
\figfinetuning
\tablefinetuning

\subsection{Comparison with Optimization-Based Methods}
\boldparagraph{Interpolated View Comparison}
We further compare our model's fine-tuning performance on dynamic scenes against other test-time optimization methods under the challenging sparse 80\% drop rate, reporting the PSNR and LPIPS curves in \figref{fig:fineuning}. In the supplementary material, we further evaluate a more practical setting with three 120-frame Waymo NOTR validation sequences, a denser Drop 25\% sparsity level, and higher-resolution $640\times960$ images.

 As shown in \tabref{tab:table_finetuning}, the baselines exhibit clear limitations under this sparse setting. NeRF-based methods, including SUDS~\citep{turki2023suds} and EmerNeRF, struggle to decompose dynamic elements in a fully self-supervised manner, resulting in significant artifacts with sparse inputs (see supplementary video). While 3DGS-based baselines such as StreetGaussian and OmniRe achieve better results by leveraging 3D bounding box priors, they still require reconstructing the entire scene from scratch, leading to long training time. Note that EmerNeRF, StreetGaussian, and OmniRe all require LiDAR observations as input, ensuring a fair comparison. DeSiRe-GS~\citep{peng2025desire} adopts a complex two-stage pipeline that distills 2D motion into Gaussian space based on the PVG~\citep{chen2023periodic} representation, resulting in excessive training time (14 minutes). In contrast, our model enables rapid initialization in just 1.3s (denoted as Ours$_\text{0-iter}$ in \tabref{tab:table_finetuning}) and supports subsequent fine-tuning to further enhance reconstruction quality, converging within $\approx$1000 steps to achieve superior fidelity.

\boldparagraph{Extrapolated View Comparison}
We further evaluate extrapolated view rendering under both 1 m and 3 m lane shifts, and compare with optimization-based baselines, including StreetGaussian and OmniRe. As shown in \tabref{tab:lane_shift_kid_optimization}, Ours-FT achieves the best KID at 1 m on both Waymo and PandaSet. Under the more challenging 3 m lane shift, Ours-FT remains comparable to the strongest optimization-based baseline. This indicates that the generalizable priors provide a strong initialization and help preserve geometry under shifted trajectories. Ours-FF shows a performance drop under the larger 3 m shift, but still preserves clear close-range structures in qualitative results. We provide qualitative extrapolation results and editing comparisons with OmniRe in the supplementary material.

\begin{table}[htbp]
\centering
\small
\setlength{\tabcolsep}{2.5pt}
\renewcommand{\arraystretch}{1.2}
\begin{tabular*}{\linewidth}{@{\extracolsep{\fill}}>{\scriptsize}lcccc@{}}
\toprule
\multirow{2}{*}{\normalsize Method}
& \multicolumn{2}{c}{\normalsize Waymo}
& \multicolumn{2}{c}{\normalsize PandaSet} \\
\cmidrule(lr){2-3} \cmidrule(lr){4-5}
& \makecell{\normalsize KID$\downarrow$\\\normalsize @ 1 m}
& \makecell{\normalsize KID$\downarrow$\\\normalsize @ 3 m}
& \makecell{\normalsize KID$\downarrow$\\\normalsize @ 1 m}
& \makecell{\normalsize KID$\downarrow$\\\normalsize @ 3 m} \\
\midrule
StreetGaussian~\citep{streetgaussian}
& 0.055 & 0.203 & 0.062 & 0.198 \\ 
OmniRe~\citep{chen2024omnire}
& 0.041 & 0.158 & 0.033 & \textbf{0.116} \\
EVolSplat4D (Ours-FF)
& 0.072 & 0.230 & 0.095 & 0.244 \\
EVolSplat4D (Ours-FT)
& \textbf{0.038} & \textbf{0.114} & \textbf{0.029} & 0.118 \\
\bottomrule
\end{tabular*}
\caption{\textbf{Quantitative comparison of view extrapolation quality} with optimization-based baselines. We report KID at 1 m and 3 m lane shifts on Waymo and PandaSet.}
\label{tab:lane_shift_kid_optimization}
\end{table}

\subsection{Ablation Study and Analysis}
For fast ablation experiments, we train on a smaller subset of sequences from the Waymo dataset. The resulting pretrained models are then evaluated in a feed-forward manner under an 80\% drop rate sparsity setting to ablate the design choices of \method. \tabref{tab:ablation} and \figref{fig:Ablation study.} present the quantitative and qualitative results, respectively.
\figAblation
\begin{table}[t]
\centering
\small
\renewcommand{\arraystretch}{1.2}

\begin{tabular*}{\linewidth}{@{\extracolsep{\fill}} l c c c @{}}
\toprule
& PSNR$\uparrow$  & SSIM$\uparrow$  & LPIPS$\downarrow$ \\
\cmidrule(lr){2-4} %
w/o volume branch       & 27.02  & 0.839 & 0.132   \\
w/o pixel branch        & 26.29  & 0.817 & 0.194   \\
w/o motion-adjusted IBR & 26.77  & 0.837 & 0.121   \\
w/o Occ. check          & 27.33  & 0.844 & 0.113   \\
w/o $\mathcal{L}_\text{mask}$ & 27.45 & 0.847 & 0.108  \\
\midrule
Full                    & 27.78  & 0.856 & 0.102   \\
\bottomrule
\end{tabular*}
\caption{\textbf{Ablation Analysis} of method components.}
\label{tab:ablation}
\end{table}

\boldparagraph{Effectiveness of Hybrid Representation}
We employ a hybrid representation that integrates 3D Gaussians from both volumetric and pixel-aligned branches to model static environments. To validate this architectural choice, we evaluate two variants, one omitting the volumetric branch and another omitting the far-field branch. In the first variant that excludes the 3D volumetric CNN, the pixel branch is responsible for modeling the entire static scene, including both the far-field background and the close-range volume, yielding a configuration conceptually similar to MVSplat. This setup leads to a performance degradation of 0.76 dB in PSNR, 0.017 in SSIM, and 0.03 in LPIPS. Under this per-pixel scheme, novel views of the close-range region exhibit ghosting artifacts (see \figref{fig:wo volume branch}) caused by redundant and inconsistent Gaussian predictions in overlapping image areas.

Conversely, we evaluate a variant without the pixel branch, replacing it with the simple hemisphere background model from our conference version, EVolSplat \citep{miao2025evolsplat}. This model only roughly approximates the geometry of the far-field scenery, failing to capture complex structures and sometimes resulting in artifacts in the background region (see Figure \ref{fig: wo pixel branch}).

\begin{table}[t]
\centering
\small
\renewcommand{\arraystretch}{1.2}

\begin{tabular*}{\linewidth}{@{\extracolsep{\fill}} l c c c @{}}
\toprule
Setting & PSNR$\uparrow$  & SSIM$\uparrow$  & LPIPS$\downarrow$ \\
\midrule
$W=1$, Pred bbx & 27.45 & 0.850 & 0.115 \\
\textbf{$W=3$, Pred bbx} & \textbf{27.78} & 0.856 & 0.102 \\ 
$W=5$, Pred bbx  & 27.56  & 0.853 & 0.106   \\
\midrule 
$W=3$, Gt bbx    & 27.73  & \textbf{0.860} & \textbf{0.095}   \\
\bottomrule
\end{tabular*}
\caption{\textbf{Ablation study on window size ($W$) and bounding box source.} Using $W=3$ with predicted bounding boxes (Pred bbx) yields the best results among window sizes. Notably, this performance is comparable to using ground-truth bounding boxes (Gt bbx).}
\label{tab:window}
\end{table}

\begin{figure}[t]
    \centering
    \includegraphics[width=0.49\columnwidth]{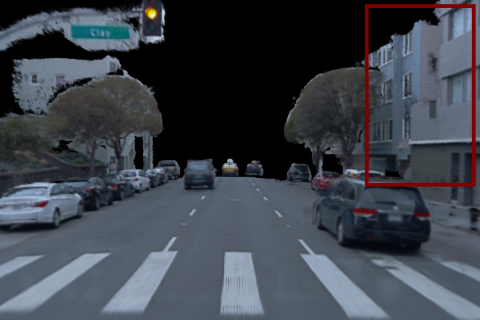}
    \includegraphics[width=0.49\columnwidth]{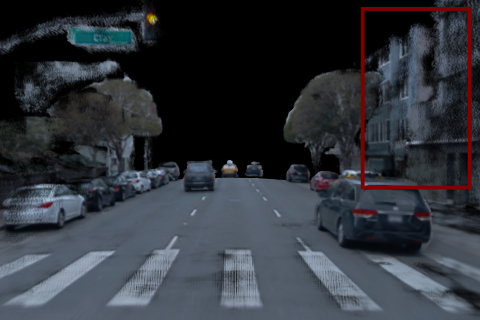}
    \noindent %
    \makebox[0.5\columnwidth]{w $\mathcal{L}_\text{mask}$}%
    \makebox[0.5\columnwidth]{w/o $\mathcal{L}_\text{mask}$}%
    \caption{\textbf{Qualitative Results of Scene Decompose Loss.} The images are rendered only from the Gaussians within the close-range region. Without $\mathcal{L}_\text{mask}$, the far-field pixel branch partially models the close-range building, causing Gaussians from the $\mathcal{G}^{\mathrm{cr}}$ to be semi-transparent. }
    \label{fig:scene_decompose_ablation} %
\end{figure}

\boldparagraph{Effectiveness of Motion-Adjusted IBR}
We now validate the importance of our motion-adjusted IBR strategy, a core component of the Generalizable Dynamic Gaussian branch (\secref{sec:dynamic gaussian}). Our full method transforms an actor's canonical points into the world space corresponding to each reference frame's specific timestamp before querying appearance features. We compare this to a baseline variant that omits this temporal alignment, which directly projects primitives at the target time to nearby reference images. The results in \tabref{tab:ablation} show a clear decrease in rendering quality (PSNR 27.78 → 26.77). As seen in \figref{fig: wo motion ibr}, this performance drop is mostly due to the severely degraded reconstruction quality of the dynamic vehicles. This demonstrates that our motion-adjusted IBR strategy is essential for querying consistent appearance features across time, facilitating high-fidelity dynamic rendering.

\boldparagraph{Effectiveness of Occlusion Check}
To investigate the occlusion check importance, we perform a study that removes the visibility maps $v_{ik}$ and instead naively aggregates all 2D colors retrieved from $K$ reference views to learn Gaussian appearance. As illustrated in \figref{fig:occ check}, the rendering quality in occluded regions suffers from noticeable blurriness. By incorporating DINO feature as priors, our model effectively removes the impact of the occlusion-caused feature, resulting in an overall improvement of about 0.45dB PSNR as reported in \tabref{tab:ablation}.

\boldparagraph{Effectiveness of Scene Decomposition}
We now evaluate the effectiveness of our Scene Decomposition loss $\mathcal{L}_{\text{mask}}$, which is designed to enforce spatial separation between the close-range ($\mathcal{G}^{\mathrm{cr}}$) and far-field ($\mathcal{G}^{\mathrm{far}}$) branches. As reported in \tabref{tab:ablation} (w/o $\mathcal{L}_\text{mask}$), while removing this loss has only a minor impact (0.33dB drop in PSNR) on the rendering fidelity, clean decomposition is crucial for maintaining geometric accuracy of the close-range volume. 
Without it, the network may still synthesize plausible images but fail to disentangle the scene geometry. As shown in the rendering of the close-range volume in \figref{fig:scene_decompose_ablation}, a part of Gaussians $\mathcal{G}^{\mathrm{cr}}$ in the close-range volume branch becomes semi-transparent without $\mathcal{L}_{\text{mask}}$.

\boldparagraph{Analysis of Color Window}
We further investigate the impact of the projection window size $W$ on reconstruction quality. As shown in Table~\ref{tab:window}, we compare three window configurations ($W=\{1,3,5\}$) through feed-forward inference on novel scenes. At $W=1$, each Gaussian retrieves appearance information from only a single pixel per reference frame, resulting in the lowest PSNR. This performance drop stems from inaccurate Gaussian positions or noisy bounding box priors, which lead to misaligned 2D projections. Increasing the window size to $W=3$ improves image quality by integrating local texture information, which effectively compensates for these projection inaccuracies. However, at $W=5$, the inclusion of redundant color information introduces blurring, leading to an increase in LPIPS and higher memory overhead.

Importantly, this window-based sampling strategy also mitigates the effects of noisy 3D bounding box predictions for dynamic actors. A comparison between Row 2 and Row 4 in Table~\ref{tab:window} validates this robustness. Our full model, utilizing predicted boxes (27.78 dB PSNR), achieves performance on par with the variant using ground-truth bounding boxes (27.73 dB PSNR), with only a negligible 0.004 difference in SSIM. This experiment confirms that our aggregation strategy successfully compensates for the noise inherent in automated object tracking, enabling robust and high-fidelity 4D reconstruction.

\boldparagraph{Analysis of Bounding Box Noise}
To further evaluate the sensitivity to box quality, we perturb the predicted 3D bounding boxes during inference by adding center translation Gaussian noise with $\sigma=0.25\,\mathrm{m}$ and $\sigma=0.5\,\mathrm{m}$, and yaw Gaussian noise with $\sigma=10^\circ$ and $\sigma=15^\circ$. As shown in \tabref{tab:box_noise}, our method remains reasonably robust to box perturbations. Compared with the clean predicted-box setting, center noise with $\sigma=0.25\,\mathrm{m}$ decreases PSNR from 27.78 to 27.28, while yaw noise with $\sigma=10^\circ$ maintains a similar PSNR of 27.52. Even under stronger perturbations, the results remain stable, showing that the dynamic branch is not sensitive to moderate box noise.

\begin{table}[t]
\centering
\setlength{\tabcolsep}{4pt}
\renewcommand{\arraystretch}{1.2}
\resizebox{\linewidth}{!}{%
\begin{tabular}{lccc}
\toprule
\textbf{Noisy Setting} & \textbf{PSNR$\uparrow$} & \textbf{SSIM$\uparrow$} & \textbf{LPIPS$\downarrow$} \\
\midrule
Pred Box+Center noise $\sigma=0.25\,\mathrm{m}$ & 27.28 & 0.836 & 0.120 \\
Pred Box+Center noise $\sigma=0.5\,\mathrm{m}$ & 26.76 & 0.821 & 0.125 \\
Pred Box+Yaw noise $\sigma=10^\circ$ & 27.52 & 0.844 & 0.114 \\
Pred Box+Yaw noise $\sigma=15^\circ$ & 27.31 & 0.841 & 0.117 \\
Pred Box & \textbf{27.78} & \textbf{0.856} & \textbf{0.102} \\
\bottomrule
\end{tabular}
}
\caption{\textbf{Ablation study of bounding box perturbation} on the Waymo dataset.}
\label{tab:box_noise}
\end{table}

\boldparagraph{Analysis of $\Delta \mu_i$ Update}
\label{sec:delta_mu}
As mentioned in Sec.~3.2, $\Delta \mu_i$ depends on the previous estimation $\Delta \mu_i^{prev}$. To empirically evaluate the stability of this recursive update, we further conduct experiments by recursively updating the offsets ($i=0,1,2,3$) during inference. As reported in \tabref{tab:recursion offset}, our pretrained model successfully refines the noisy primitive's position after the first update (increasing PSNR by approximately 0.52 dB) and maintains a stable value of 27.78dB even with additional updates.

\begin{table}[t]
    \centering
    \small 
    \renewcommand{\arraystretch}{1.2}

    \begin{tabular*}{\linewidth}{@{\extracolsep{\fill}} l c c c @{}}
    \toprule
     & PSNR$\uparrow$  & SSIM$\uparrow$  & LPIPS$\downarrow$ \\
    \cmidrule(lr){2-4} 
    $i=0$ & 27.260 & 0.833 & 0.142  \\
    $i=1$ & 27.787 & 0.856  & 0.102  \\
    $i=2$  & 27.781  & 0.856 & 0.102   \\
    $i=3$  & 27.785  & 0.856 & 0.102   \\
    \bottomrule
    \end{tabular*}
        
    \caption{\textbf{Ablation Study} on recursion of $\Delta \mu_i$. The results are tested on the Waymo dataset.}
    \label{tab:recursion offset}
\end{table}

\newcolumntype{L}[1]{>{\raggedright\arraybackslash\hsize=#1\hsize}X}
\newcolumntype{C}[1]{>{\centering\arraybackslash\hsize=#1\hsize}X}
\renewcommand{\tabularxcolumn}[1]{>{\arraybackslash}m{#1}}

\begin{table}[t]
\centering
\renewcommand{\arraystretch}{1.2} %

\resizebox{\linewidth}{!}{%
    \begin{tabular}{@{} l l c c c @{}}
    \toprule
    Modality & Method & PSNR$\uparrow$ & SSIM$\uparrow$ & LPIPS$\downarrow$ \\
    \midrule
    
    \multirow{2}{*}{\makecell[l]{Monocular\\Pred.}} 
      & UniDepth~\citep{piccinelli2024unidepth} & 27.59 & 0.854  & 0.105  \\
      & Metric3D~\citep{yin2023metric3d}        & 27.86 & 0.859  & 0.104  \\
      
    \midrule
    
    \makecell[l]{LiDAR\\Compl.} 
      & \makecell[l]{Prior Depth Anything\\\citep{wang2025depth}} 
      & 27.78 & 0.856 & 0.102 \\
      
    \bottomrule
    \end{tabular}%
}
\caption{\textbf{Depth Sensitivity Experiment.} The results are averaged on five testsets from the Waymo dataset.}
\label{tab:depth_sensitivity}
\end{table}

\begin{figure*}[htbp]
    \centering
    \setlength{\tabcolsep}{1pt} %
    
    \begin{tabular}{@{} >{\centering\arraybackslash}m{0.245\textwidth}
                         >{\centering\arraybackslash}m{0.245\textwidth}
                         >{\centering\arraybackslash}m{0.245\textwidth}
                         >{\centering\arraybackslash}m{0.245\textwidth} @{}}
    
    {\includegraphics[width=\linewidth]{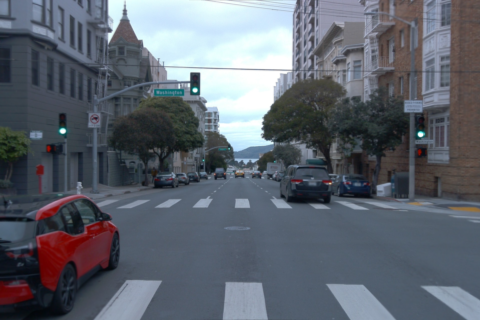}}
    & {\includegraphics[width=\linewidth]{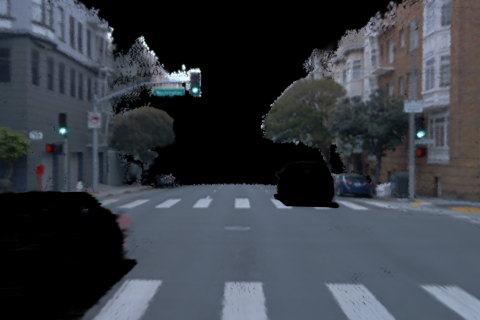}}
    & {\includegraphics[width=\linewidth]{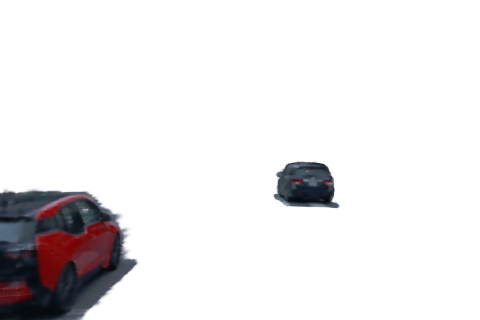}}
    & {\includegraphics[width=\linewidth]{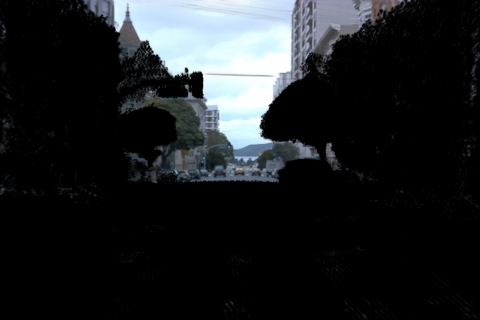}} \\
    
    \addlinespace[2pt] %
    
    {\includegraphics[width=\linewidth]{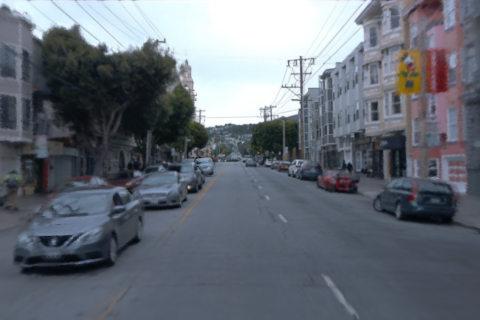}}
    & {\includegraphics[width=\linewidth]{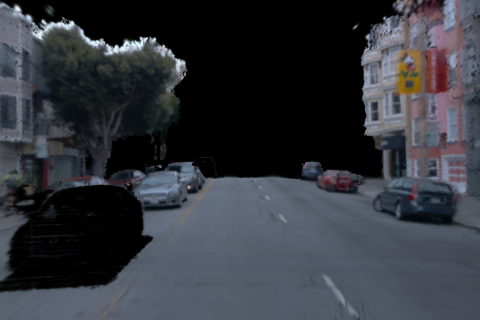}}
    & {\includegraphics[width=\linewidth]{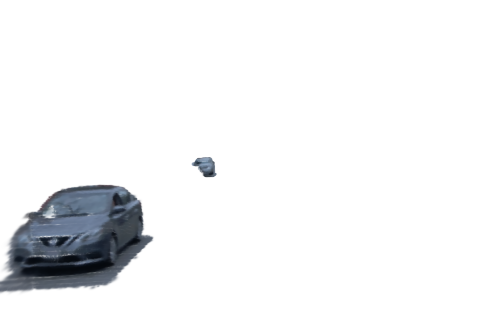}}
    & {\includegraphics[width=\linewidth]{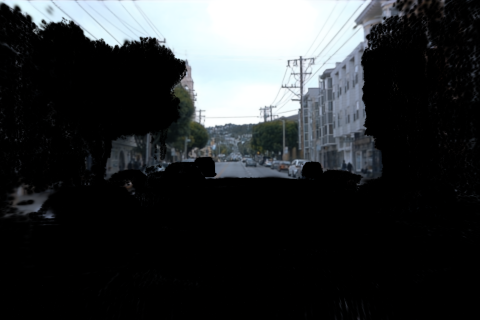}} \\

    \addlinespace[3pt] %
    
    \bfseries (a) Reference Image
    & \bfseries (b) Predicted Close-Range
    & \bfseries (c) Dynamic Actors
    & \bfseries (d) Predicted Far-Field \\
    
    \end{tabular}
    \vspace{-0.2cm}
    \caption{\textbf{Scene Decomposition on Waymo.} Our approach enables clear decomposition of the close-range volume, dynamic actors, and the far-field scenery.}
    \label{fig:scene_decompose}
\end{figure*}%

\boldparagraph{Analysis of Depth Modularity}
We further validate the robustness of our framework with respect to the choice of depth initialization. Specifically, we replace our default LiDAR priors with monocular predictions from pre-trained models (UniDepth \citep{piccinelli2024unidepth} and Metric3D \citep{yin2023metric3d}). In these monocular variants, we utilize the estimated depth maps to initialize both the global point cloud and dynamic vehicles in canonical space. As shown in  \tabref{tab:depth_sensitivity}, the results are highly consistent across all input modularities. Our model achieves nearly identical performance regardless of the depth source, with all PSNR approximately 27dB and SSIM at 0.86. These results confirm that our system is flexible and can be initialized with depth maps from various distributions.

\figEdit
\subsection{Application}

\boldparagraph{Scene Editing}
Our instance-aware scene representation facilitates a variety of scenario-level editing operations. Leveraging the disentangled 3D Gaussians, we can seamlessly replace a specific vehicle with another instance (\figref{fig:replaced}), translate actors to novel positions (\figref{fig:shift}), or entirely remove all dynamic objects from the environment (\figref{fig:delete}). These capabilities demonstrate the flexibility of our object-centric canonical modeling for simulation purposes. We also provide qualitative comparisons with the optimization-based OmniRe baseline and failure cases for actor shifts in the supplementary material.

\boldparagraph{Scene Decomposition}
The proposed hybrid reconstruction framework captures comprehensive 4D geometry by factorizing the environment into three constituent components: the \textbf{close-range volume}, \textbf{dynamic actors}, and the \textbf{far-field scenery
}. As illustrated in \figref{fig:scene_decompose}, our method successfully disentangles these elements into independent representations, yielding cleanly decomposed layers suitable for downstream autonomous driving tasks.

\boldparagraph{Auxiliary Outputs}
Beyond held-out RGB rendering, the reconstructed dynamic Gaussians can be rendered into auxiliary outputs, including depth maps, dynamic masks, and optical flow. They can also be re-rendered under modified camera intrinsics, extrinsics, or camera aspect ratios. Representative visualizations are provided in the supplementary material.

\section{Limitations}
Our method has several limitations. First, while our core method is not inherently dependent on active sensors, we currently utilize LiDAR-derived 3D bounding boxes to define the object-centric canonical spaces. This design follows common practice in per-scene optimization~\citep{streetgaussian,tonderski2024neurad} and is motivated by the higher accuracy of LiDAR-based detection, which significantly outperforms existing RGB-only methods~\citep{sun2025sparsedrive,hu2022monocular} by about 40\% mAP on the KITTI 3D detection benchmark~\citep{kitti}. 
However, our method remains architecturally decoupled from the LiDAR sensor: it processes RGB images and depth priors for feature learning and rendering, where monocular-based depth priors yield rendering quality comparable to LiDAR-based alternatives. This modularity ensures that the approach remains functional using only camera data, provided that 3D bounding boxes are supplied by an external module.
As monocular 3D detection performance continues to bridge the gap with LiDAR-based methods, our approach can be seamlessly transitioned to a fully camera-only pipeline. Future work will explore the end-to-end integration of vision-based detection and tracking.

Secondly, despite that our method produces stable extrapolated rendering results, we observe a performance decline in large-baseline extrapolation experiments (e.g., translation $>3$m). While the close-range volume maintains robust synthesis quality, the performance in the far-field tends to degrade. This stems from the difference in geometric quality: the close-range branch leverages explicit metric depth for initialization, thus achieving high geometric fidelity and better generalization to extrapolated views. In contrast, the far-field branch is supervised solely by rendering loss without explicit depth priors. While reasonable rendering quality can be observed during interpolation, the synthesis quality drops under large camera translations. A potential solution is to incorporate stronger depth priors~\citep{yu2026infinidepth} or external supervision to recover more plausible geometry for the background model.

Finally, we assume that all moving objects follow simple rigid motion, which may cause blurring for non-rigid moving actors (e.g., pedestrians). This could be addressed by incorporating non-rigid dynamic reconstruction methods~\citep{yang2024deformable,mao2025unire} into our pipeline.

\section{Conclusion}
We propose~\method, an efficient feed-forward approach for reconstructing challenging dynamic urban scenes. By effectively incorporating valuable geometry cues from LiDAR, we decompose the scene into three independent components: a static close-range volume, dynamic actors, and a far-field region, each handled by a dedicated generalizable module. Our compositional reconstruction model unifies both dynamic and static feed-forward reconstruction from sparse camera inputs. The practical applications of our method include rapid, multiple urban scene reconstruction, object manipulation, and high-fidelity scene decomposition. We believe that our method pushes the frontier of feed-forward models for unbounded autonomous driving scenes with complex geometry and motion.

\vspace{3em}
\noindent \textbf{Acknowledgements}
This work was supported by the Natural Science Foundation of Zhejiang Province, China under Grant No. LR26F010002; the National Natural Science Foundation of China under Grant No. 62522317 and 62373322; the Natural Science Foundation of Zhejiang Province, China under Grant No. LD25F030001; the Key R\&D Program of Hangzhou under Grant No. 2024SZD1A30; as well as the Scientific Research Fund of Zhejiang University under Grant XY2025028.
\vspace{2em}

\noindent \textbf{Author Contributions}
All authors contributed to the conception and design of the study. Yiyi Liao led the entire project, including conceptualization, methodology design, experiment design, and manuscript writing. Sheng Miao and Sijin Li performed the core method development, conducted all experiments, and analyzed the results. Sheng Miao drafted the initial manuscript. All authors provided critical feedback on the previous version of the manuscript and approved the final submission.
\vspace{2em}

\noindent \textbf{Funding}
This work was supported by the Natural Science Foundation of Zhejiang Province, China under Grant No. LR26F010002; the National Natural Science Foundation of China under Grant No. 62522317 and 62373322; the Natural Science Foundation of Zhejiang Province, China under Grant No. LD25F030001; the Key R\&D Program of Hangzhou under Grant No. 2024SZD1A30; as well as the Scientific Research Fund of Zhejiang University under Grant XY2025028.
\vspace{2em}

\noindent \textbf{Data Availability}
All datasets analyzed during this study are publicly available in the following public repositories:
\begin{itemize}
    \item \textbf{KITTI-360:} \url{https://www.cvlibs.net/datasets/kitti-360/}
    \item \textbf{KITTI:} \url{https://www.cvlibs.net/datasets/kitti/}
    \item \textbf{Waymo Open Dataset:} \url{https://waymo.com/open/}
    \item \textbf{PandaSet:} \url{https://pandaset.org/}
\end{itemize}

\bibliographystyle{spbasic}
\bibliography{11_references}

\clearpage
\setcounter{section}{0}
\setcounter{figure}{0}
\setcounter{table}{0}
\setcounter{equation}{0}
\setcounter{page}{1} %

\renewcommand{\thesection}{S\arabic{section}}
\renewcommand{\thefigure}{S\arabic{figure}}
\renewcommand{\thetable}{S\arabic{table}}
\renewcommand{\theequation}{S\arabic{equation}}

\twocolumn[
  \begin{@twocolumnfalse}
    \begin{flushleft}
      {\Large \bfseries Supplementary Material for: \\ EVolSplat4D: Efficient Volume-based Gaussian Splatting for 4D Urban Scene Synthesis \par}
      
      \vspace{2em} %
    \end{flushleft}
  \end{@twocolumnfalse}
]

\section{Implementation Details}
\subsection{Remove Depth Outliers}
We begin by applying a depth consistency check to filter out noisy depth data. Specifically, we unproject the depth map $D_i$ of $i$th frame to 3D and reproject it to a nearby view $j$, obtaining a projected depth $D_{i\rightarrow j}$. Next, we compare $D_{i\rightarrow j}$ and $D_j$ and filter out depths where the absolute relative error exceeds an empirical threshold $\sigma = 0.2m$. We formulate this process as:

\begin{equation}
M_d=\left|D_{i\rightarrow j}-D_j\right|<\sigma, D_{i\rightarrow j}=D_{j}\left(\pi_{j} \pi_i^{-1}\left(\mathbf{u}_i\right)\right)
\end{equation}
where $\mathbf{u}_i$ denotes pixel coordinates of $i$th frame and $M_d$ represents the geometric consistency mask. We further utilize the 3D statistical filter in Open3D~\citep{zhou2018open3d} to remove the clear floaters in the unprojected point clouds for each frame. In our implementation, we set the number of neighbors to 20 and the standard deviation ratio to 2.0.

\subsection{ Close-Range Gaussian Prediction Details}
Following~\citep{3dgs}, $\boldsymbol{\Sigma}$ is decomposed into two learnable components: rotation matrix $\mathbf{R}$ and a scaling matrix $\mathbf{S}$ to hold practical physical significance, see the following formula:
\begin{equation}
\boldsymbol{\Sigma}=\mathbf{R} \mathbf{S} \mathbf{S}^T \mathbf{R}^T
\end{equation}
To allow independent optimization of both factors, we use a 3D vector $s$ representing scaling and a quaternion $q$ for rotation separately. Instead of directly learning scales $s$, we initialize the $s_\text{init}$ with the average distance of $K$ nearest neighbors using the KNN algorithm and learn the scales residual $\Delta s$ from the volume latent feature $\mathbf{h_i}$ via a decoder $D_{\text{cr}}^{cov}$. We experimentally observe that the residual learning strategy helps our network converge faster and enhances the model capacity. 
\begin{equation}
\mathbf{s} = \mathbf{s}_\text{init} + \Delta \mathbf{s}, ~~~ \Delta \mathbf{s}=D_{\text{cr}}^{cov}(\mathbf{h_i})
\end{equation}

\subsection{SparseCNN Network Architecture}
We build a generalizable efficient 3DCNN $\mathcal{F}^\text{3D}$ to provide the geometric priors for the foreground contents.  Given the global neural point cloud $\mathcal{P} \in \mathbb{R}^{N_p \times 16}$, we quantize the point cloud with the voxel size $v_\text{size} = 0.1m$ and feed these sparse tensors into the $\mathcal{F}^\text{3D}$ to predict the latent feature volume $\boldsymbol{\mathcal{V}}$. The sparse 3DCNN uses a U-Net like architecture with skip connections, comprising some convolution and transposed convolution layers.  The details of 3DCNN are listed in the \cref{tab:sparse cnn}. We use torchsparse as the implementation of $\mathcal{F}^\text{3D}$.

\begin{table}[ht]
  \centering
  \renewcommand{\arraystretch}{1.2}
  \begin{tabular*}{\columnwidth}{@{} l @{\extracolsep{\fill}} l r @{}} 
  \toprule
  \multicolumn{3}{@{}l}{\textbf{SparseCNN Network Architecture}} \\
  \midrule
  \textbf{Layer} & \textbf{Description} & \textbf{In/Out Ch.} \\
  \midrule
  Conv$_0$ & kernel = $3\times3\times3$, stride = $1$ & 16/16 \\
  Conv$_1$ & kernel = $3\times3\times3$, stride = $2$ & 16/16 \\
  Conv$_2$ & kernel = $3\times3\times3$, stride = $2$ & 16/32 \\
  Conv$_3$ & kernel = $3\times3\times3$, stride = $1$ & 32/32 \\
  Conv$_4$ & kernel = $3\times3\times3$, stride = $2$ & 32/64 \\
  Conv$_5$ & kernel = $3\times3\times3$, stride = $1$ & 64/64 \\
  DeConv$_6$ & kernel = $3\times3\times3$, stride = $2$ & 64/32 \\
  DeConv$_7$ & kernel = $3\times3\times3$, stride = $2$ & 32/32 \\
  DeConv$_8$ & kernel = $3\times3\times3$, stride = $2$ & 32/16 \\
  DeConv$_9$ & kernel = $3\times3\times3$, stride = $2$ & 16/16 \\
  \bottomrule
  \end{tabular*}%
  \vspace{0.1cm}
  \caption{\textbf{Architecture of SparseConvNet}. Each layer consists of sparse convolution, batch normalization, and ReLU.}
  \label{tab:sparse cnn}%
\end{table}%

\begin{figure*}[htbp]
    \centering
    \setlength{\tabcolsep}{0pt}
    \begin{tabular}{cc}
        \begin{minipage}{0.5\textwidth}
            \includegraphics[width=0.48\textwidth]{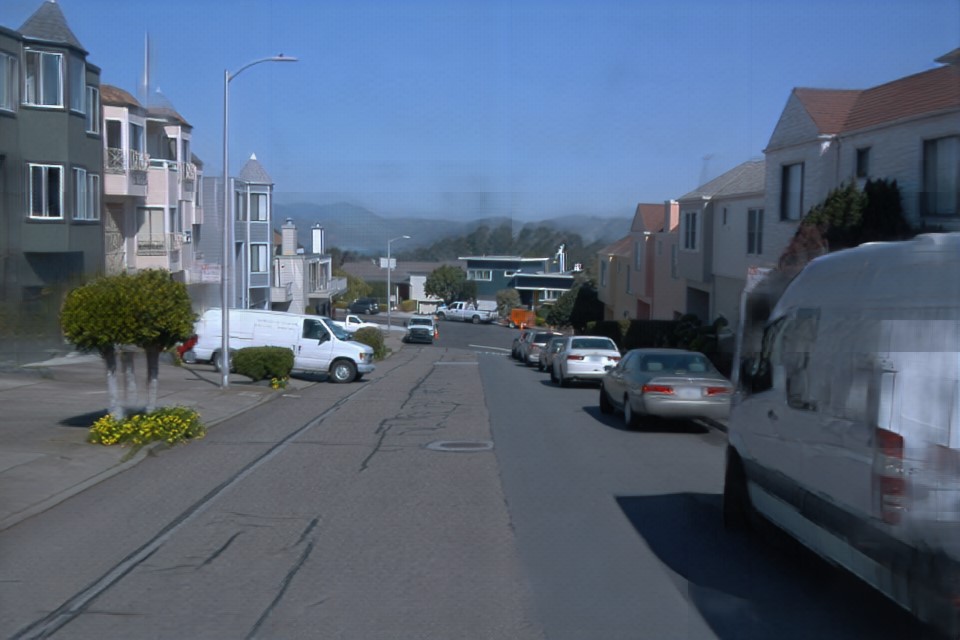} %
            \includegraphics[width=0.48\textwidth]{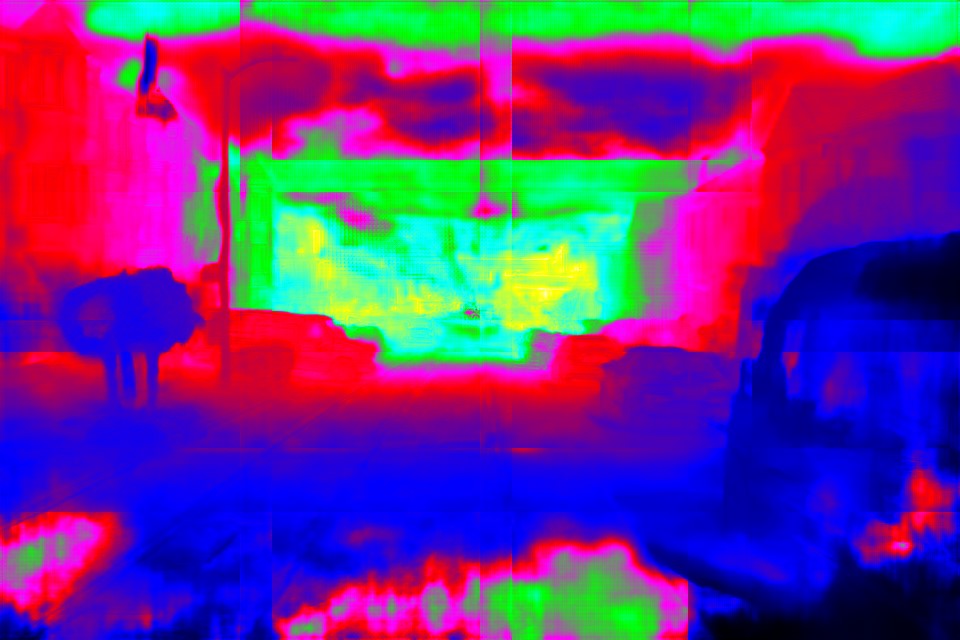} %
            \vspace{-0.25cm}
            \caption*{a) MuRF}
        \end{minipage} & 
        \begin{minipage}{0.5\textwidth}
            \includegraphics[width=0.48\textwidth]{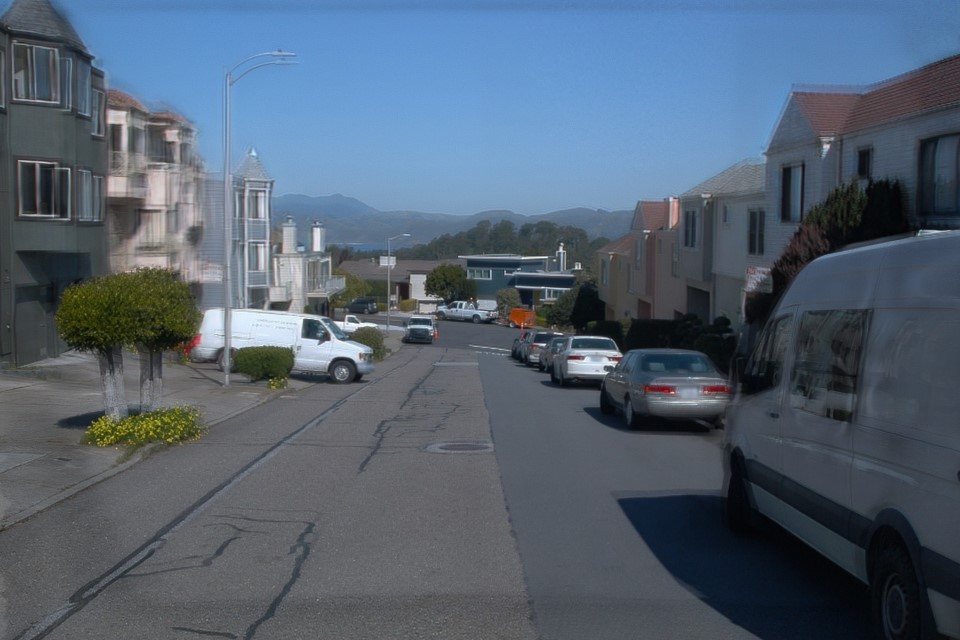} %
            \includegraphics[width=0.48\textwidth]{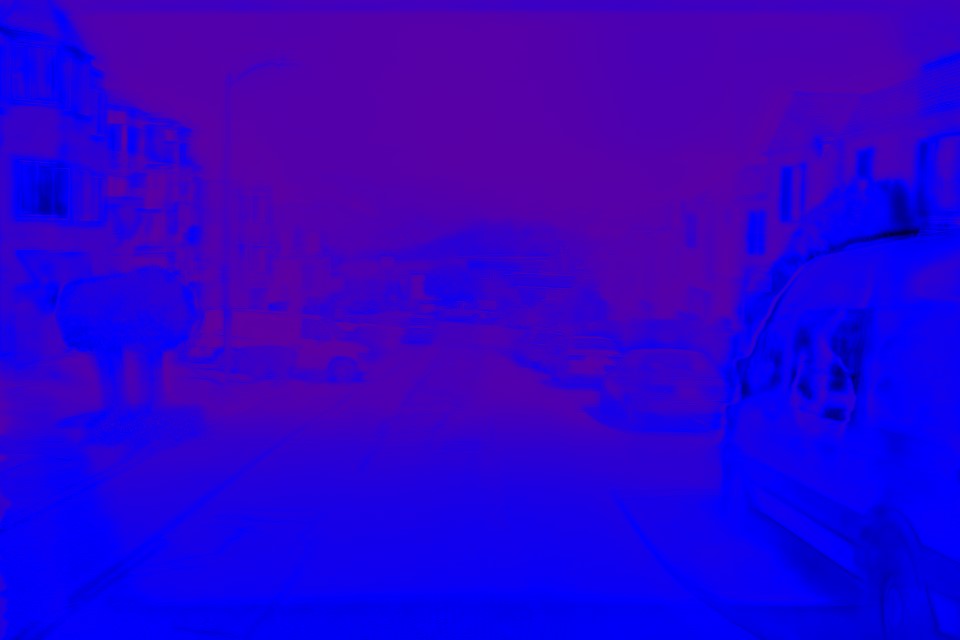} %
            \vspace{-0.25cm}
            \caption*{b) MVSplat}
        \end{minipage} 
        \vspace{0.1cm}  \\
        
        \begin{minipage}{0.5\textwidth}
            \includegraphics[width=0.48\textwidth]{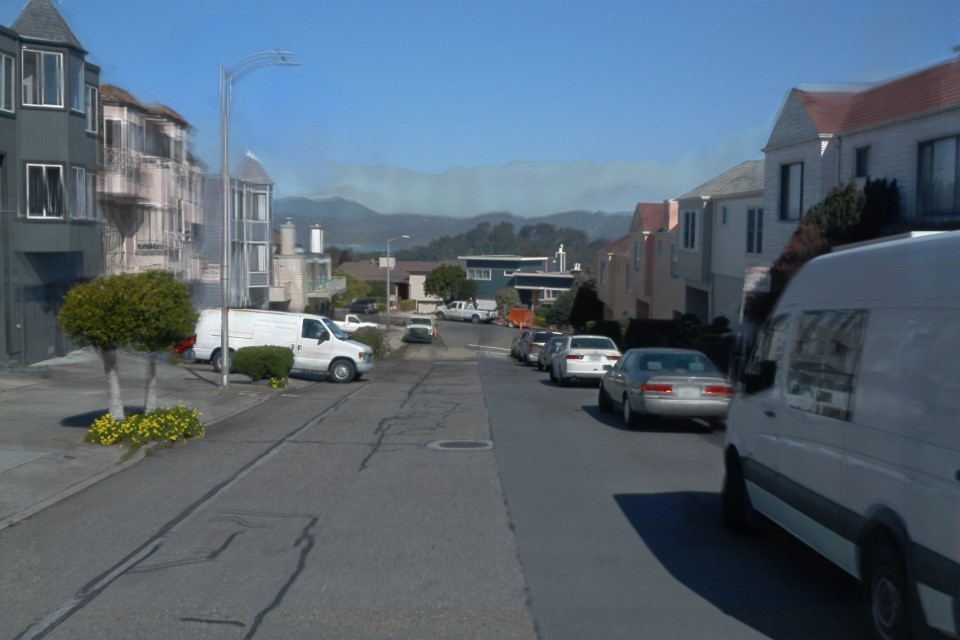} %
            \includegraphics[width=0.48\textwidth]{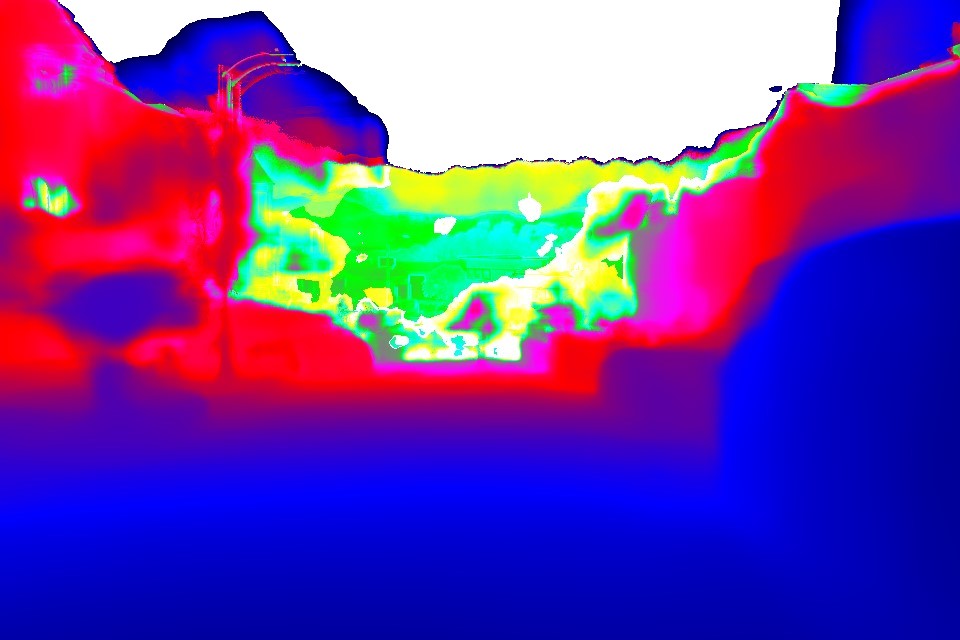} %
            \vspace{-0.25cm}
            \caption*{c) EDUS}
        \end{minipage} & 
        \begin{minipage}{0.5\textwidth}
            \includegraphics[width=0.48\textwidth]{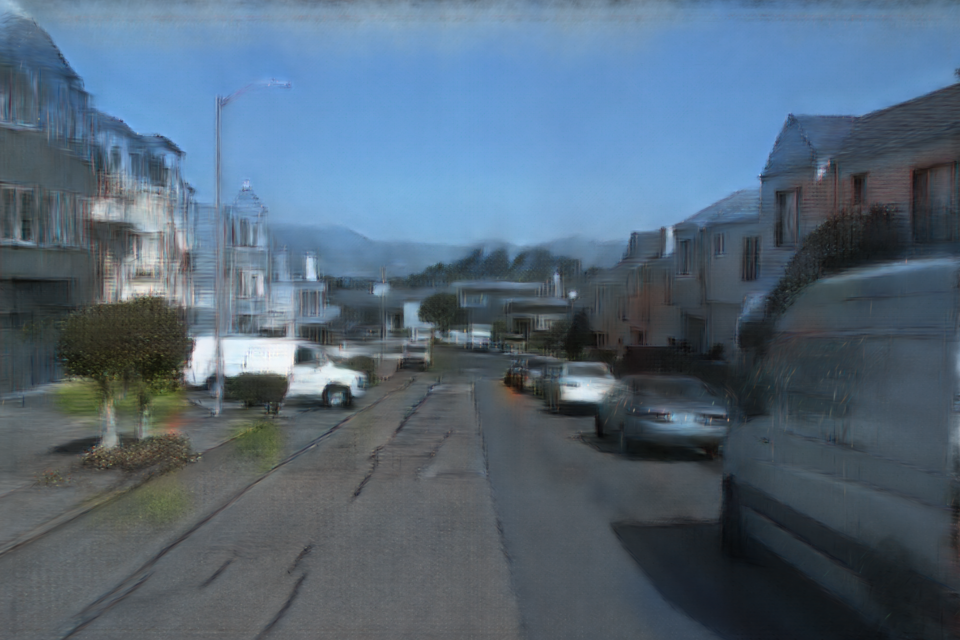} %
            \includegraphics[width=0.48\textwidth]{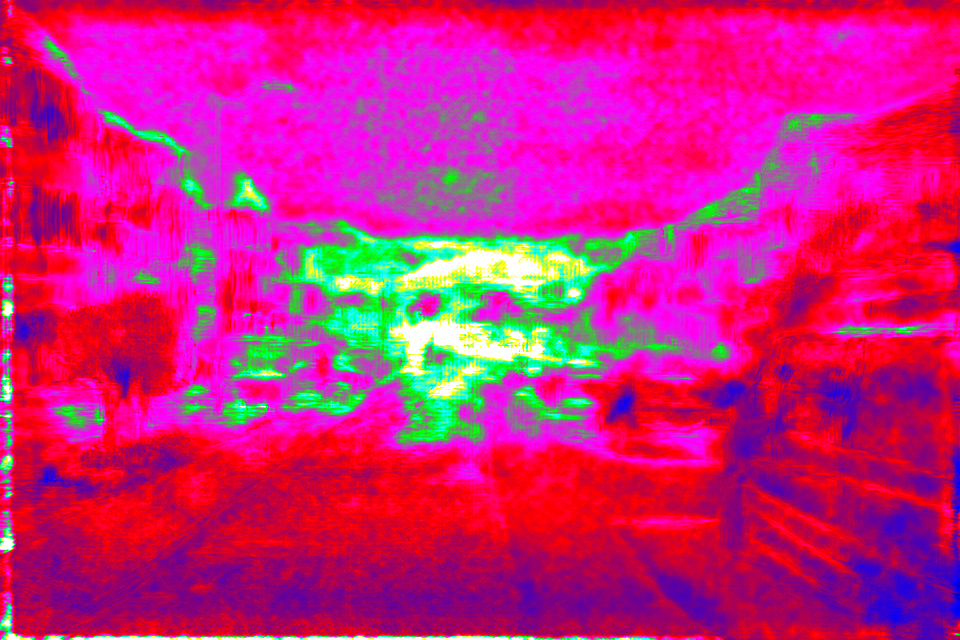} %
            \vspace{-0.25cm}
            \caption*{d) PixelSplat}
        \end{minipage} 
        \vspace{0.1cm}  \\
        
         \begin{minipage}{0.5\textwidth}
                \includegraphics[width=0.48\textwidth]{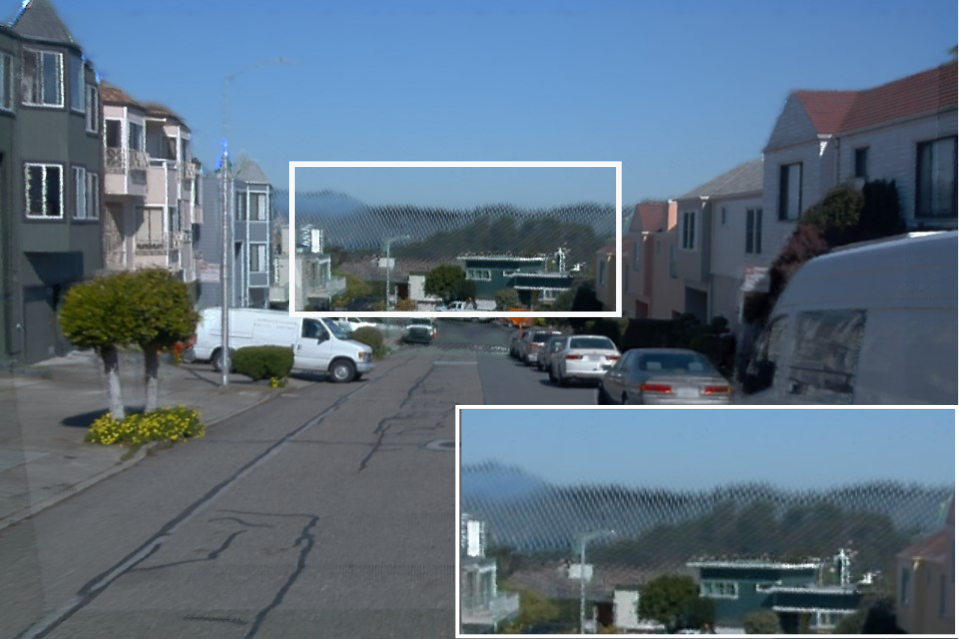} %
                \includegraphics[width=0.48\textwidth]{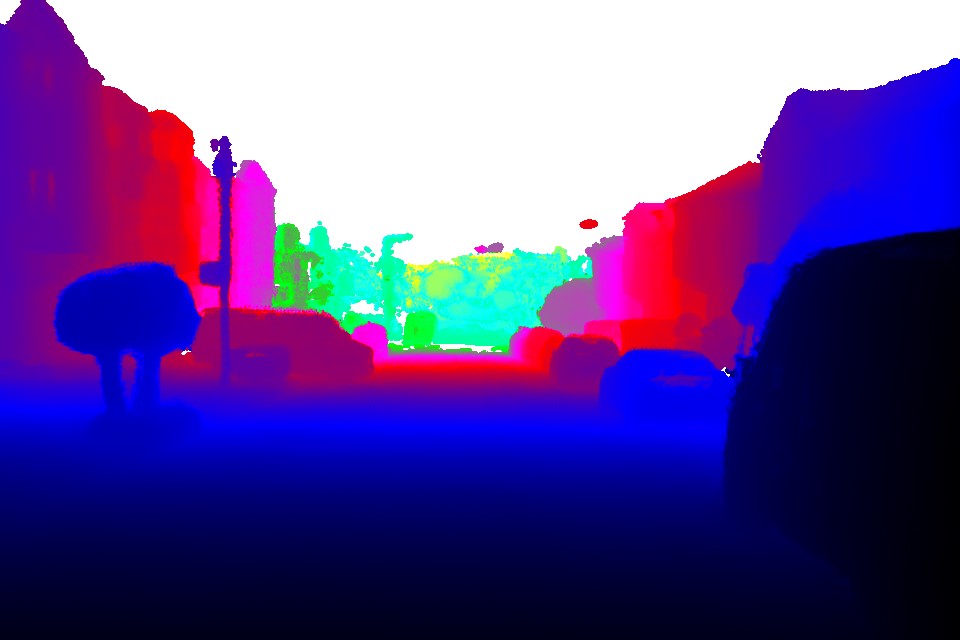} %
                \vspace{-0.25cm}
                \caption*{e) EVolSplat}
            \end{minipage} & 
            \begin{minipage}{0.5\textwidth}
                \includegraphics[width=0.48\textwidth]{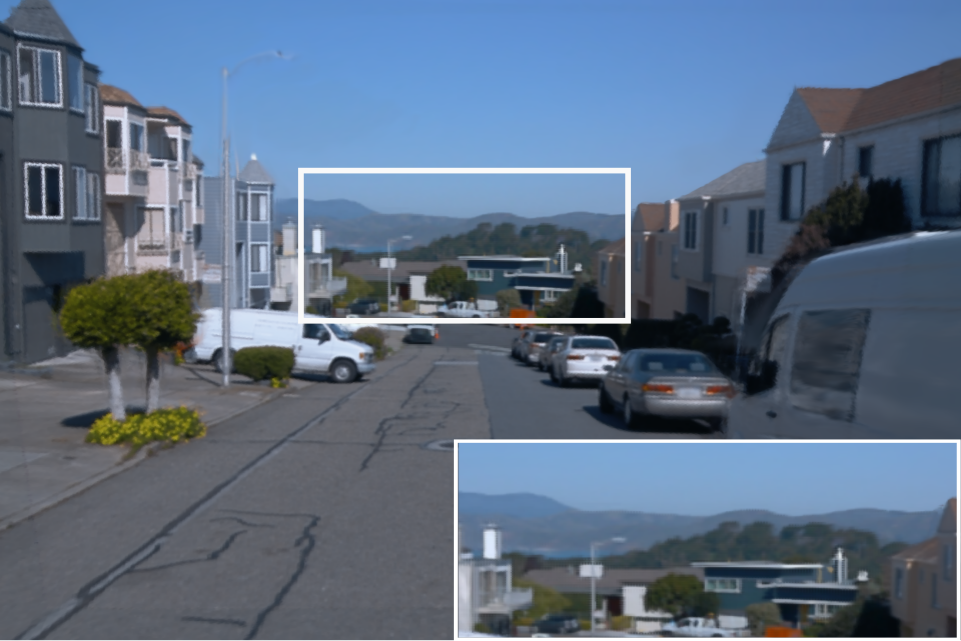} %
                \includegraphics[width=0.48\textwidth]{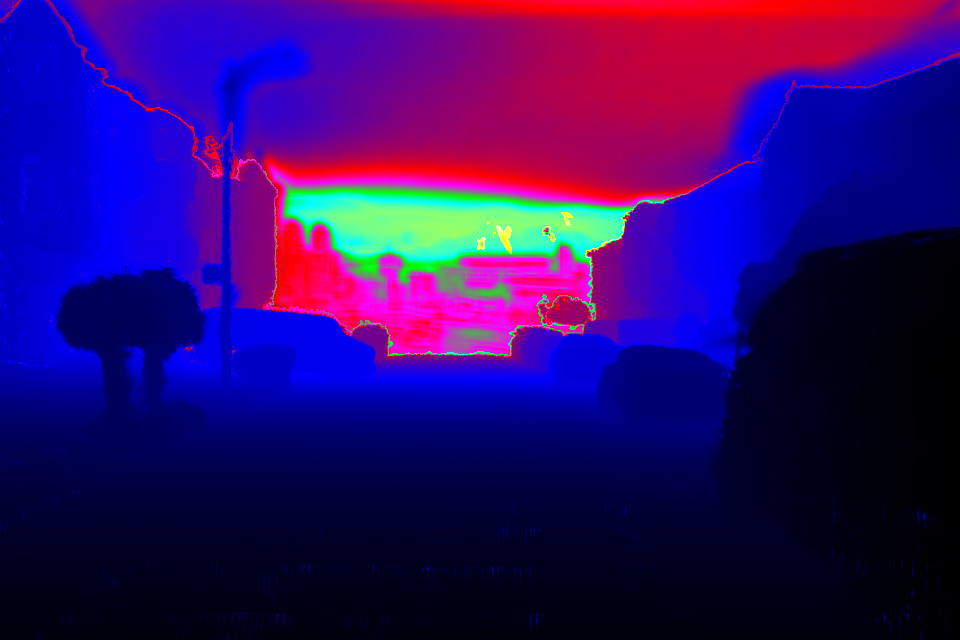} %
                \caption*{f) Ours}
            \end{minipage} 
            \vspace{0.1cm}  \\
        
    \end{tabular}
    \vspace{-0.2cm}
    \caption{\textbf {Qualitative Comparisons} with baselines for zero-shot inference on the Waymo static scenes. Better to zoom in for the far-field scenery improvements.}
    \label{fig:waymoQualitive}
\end{figure*}

\section{Datasets}
 We conduct experiments on four autonomous driving datasets: KITTI-360~\citep{liao2022kitti}, KITTI~\citep{kitti}, Waymo Open Dataset~\citep{waymo} and PandaSet~\citep{xiao2021pandaset}. 

\boldparagraph{KITTI-360} 
We use the KITTI-360 dataset \citep{liao2022kitti} for pre-training our static scenes. The training set consists of 160 static clips from sequences 00, 02, and 04, using both the front two stereo cameras.  For static experiments, we follow the Drop 50\% protocol  of KITTI-360. To be specific, we define a stereo image pair (left and right camera) as the base unit and select only the odd-numbered pairs (id=0 2 4 6…) for training. The test frames are selected as the left-eye image of even-numbered pairs(id=1 3 5 7...) to avoid overlapping with the training frames.

\boldparagraph{Waymo}
The Waymo Open Dataset is collected from various locations across the United States, containing a rich mix of static and dynamic scenes. In our feed-forward experiment for static scenes, the Waymo dataset is not used for training but serves as an out-of-domain dataset to evaluate our model’s zero-shot generalization. In the dynamic feed-forward experiment, we follow the 80\% drop rate as in STORM \citep{yang2024storm} experimental setup for a fair comparison.  To be specific, every 5th frame (id=0 5 10 15…) is selected as the training frame, while the remaining frames are used for testing. For comparison with per-scene optimization baselines, we maintain the 80\% drop rate to simulate sparse inputs.

\boldparagraph{KITTI}
To comprise more dynamic scenes, we augment our model with clips from the KITTI dataset.  We select five clips to evaluate the model generalization capacity.  Same as other dataset, the testsets from KITTI have no overlap with the training sets. Moreover, we follow the MARS~\citep{wu2023mars} using sequence 01 and 06 to demonstrate our model's capability for scene editing.

\boldparagraph{PandaSet}
 PandaSet \citep{xiao2021pandaset} serves as our primary out-of-domain dataset for dynamic feed-forward reconstruction. Our model is never trained on PandaSet, but we select five representative sequences to evaluate its zero-shot generalization performance.

\section{Baselines}
In this section, we discuss the state-of-the-art baselines used for comparison with our approach. All experiments are conducted on NVIDIA RTX 5880, except PixelSplat, MVSplat, and STORM are trained on NVIDIA V100 since their transformer-based architecture requires a large amount of GPU memory.

\boldparagraph{Feed-Forward Static Baselines}
Our static baselines include two categories: \textbf{NeRF-based methods} (MVSNeRF~\citep{mvsnerf}, MURF~\citep{xu2023murf}, and EDUS~\citep{miao2024efficient}) and \textbf{3DGS-based methods} (PixelSplat~\citep{charatan2024pixelsplat}, MVSplat~\citep{chen2024mvsplat}, DepthSplat~\citep{xu2025depthsplat}, EVolSplat~\citep{miao2025evolsplat}) and AnySplat~\citep{Jiang2025AnySplatF3}. For a fair comparison, we adopt the official implementations and default settings for all baselines but retrain them from scratch on our static dataset under the Drop50\% setting.

For NeRF-based baselines, MVSNeRF and MuRF utilize multi-view stereo (MVS) algorithms to construct the cost volume and apply 3DCNN to reconstruct a neural field while EDUS leverages the depth priors to learn a generalizable scene representation. For 3DGS-based baselines, PixelSplat predicts 3D Gaussians with a two-view epipolar transformer and then spawns per-pixel Gaussians. MVSplat exploits multi-view correspondence information for geometry learning and predicts 3D Gaussians from image features. Built on MVSPlat, DepthSplat further leverages pre-trained monocular depth features to enhance model capacity. AnySplat distills geometry priors from powerful VGGT~\citep{wang2025vggt} and employs a geometry transformer to directly decode Gaussian parameters and camera poses from images.
While EVolSplat achieves on-par results on KITTI-360, its simple hemisphere background struggles to model the complex, fine-grained far-range landscapes in the Waymo dataset. This limitation directly motivates the design of our new \textbf{Generalizable Far-Field Branch}, which is introduced in the main paper.

\begin{figure*}[htbp]
    \centering
    \small %
    
    \def\mywidth{0.32\linewidth}
    
    \begin{subfigure}[h]{\mywidth}
        \centering
        \includegraphics[width=\textwidth]{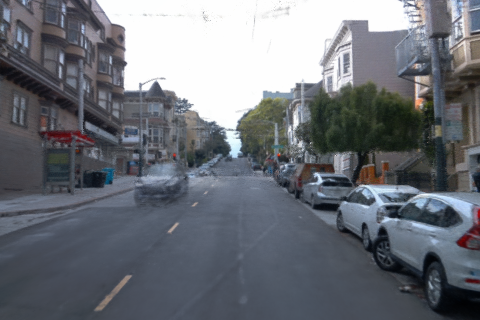}
        \caption{EmerNeRF~\citep{emernerf}}
        \label{fig:EmerNeRF}
    \end{subfigure}\hfill %
    \begin{subfigure}[h]{\mywidth}
        \centering
        \includegraphics[width=\textwidth]{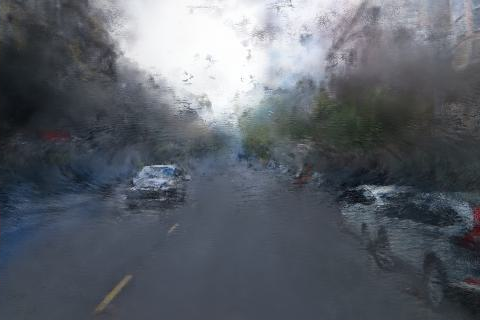}
        \caption{SUDS~\citep{turki2023suds}}
        \label{fig:SUDS} %
    \end{subfigure}\hfill %
    \begin{subfigure}[h]{\mywidth}
        \centering
        \includegraphics[width=\textwidth]{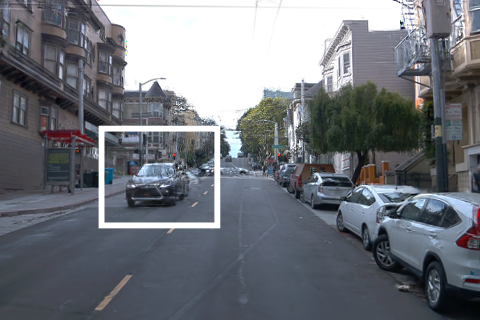}
        \caption{Street Gaussian~\citep{streetgaussian}}
        \label{fig:streetgaussian}
    \end{subfigure}
    
    \vspace{3mm} %
    
    \begin{subfigure}[h]{\mywidth}
        \centering
        \includegraphics[width=\textwidth]{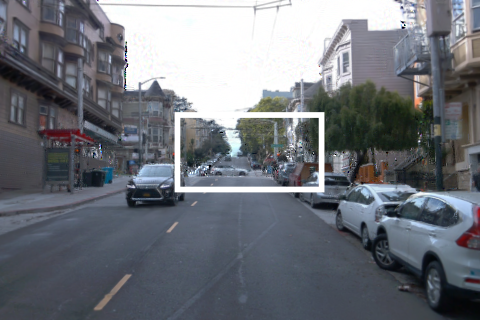}
        \caption{Omnire~\citep{chen2024omnire}}
        \label{fig:Omnire}
    \end{subfigure}\hfill %
    \begin{subfigure}[h]{\mywidth}
        \centering
        \includegraphics[width=\textwidth]{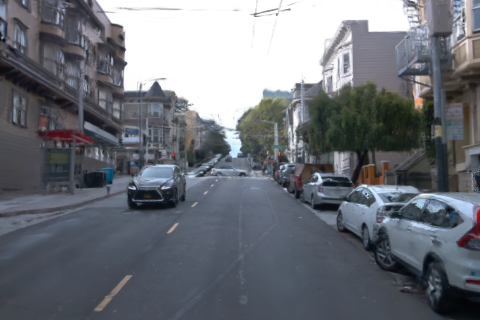}
        \caption{\method}
        \label{fig:ours_finetune} %
    \end{subfigure}\hfill %
    \begin{subfigure}[h]{\mywidth}
        \centering
        \includegraphics[width=\textwidth]{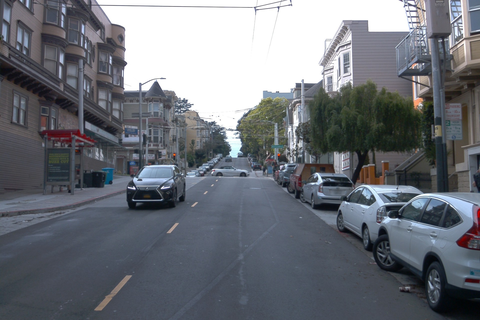}
        \caption{Ground Truth}
        \label{fig:gt_finetune} %
    \end{subfigure} 

    \caption{\textbf{Qualitative scene reconstruction comparisons} with per-scene optimization baselines under the Drop80\% sparsity level}
    \label{fig:finetune_comparison}
\end{figure*}

\boldparagraph{Feed-Forward Dynamic Baselines}
We compare against two state-of-the-art dynamic baselines, DrivingRecon \citep{lu2024drivingrecon} and STORM \citep{yang2024storm}, as well as our static-only conference version, EVolSplat \citep{miao2025evolsplat}. This comparison demonstrates that our new version can effectively handle dynamic environments. We adopt the official implementations and
default setting for all baselines, training and evaluating them under the Drop80\% setting. 

DrivingRecon is a pioneering method for generalizable dynamic reconstruction from surround-view inputs. It utilizes an encoder-decoder model with temporal cross-attention to predict 4D Gaussians directly. The latest work STORM (ICLR 2025), uses a data-driven Transformer architecture to predict per-pixel Gaussian attributes and additional velocity in a self-supervised manner. These independent Gaussians can then be warped to any particular timestamp using the velocity estimates to perform novel view rendering.
    
\boldparagraph{Per-Scene Optimization-based Baseline}
To evaluate how our generalizable priors help improve reconstruction quality from sparse inputs, we compare them against the recent per-scene optimization methods for autonomous driving scenarios under 80\% drop rate, including SUDS~\citep{turki2023suds}, EmerNeRF~\citep{emernerf}, StreetGaussian~\citep{streetgaussian}, DeSiRe-GS~\citep{peng2025desire} and OmniRe~\citep{wei2025omni}.

SUDS and EmerNeRF capture scene appearance, motion, and semantic features by decomposing scenes into static and dynamic fields, learning a flow field to represent a 4D neural radiance field. StreetGaussian and OmniRe leverage ground-truth 3D bounding boxes to decompose static background and dynamic elements. They model each dynamic actor with a dedicated local canonical space based on the Gaussian Splatting representation, showing impressive reconstruction ability. DeSiRe-GS adopts a two-stage pipeline to first extract 2D motion masks from the feature difference between the rendered and supervision images, and then distill this 2D motion information into Gaussian space.

\section{Additional Analysis and Visualizations}
\subsection{Denser and Longer-Sequence Reconstruction}
To evaluate whether our method remains effective under a more practical reconstruction setting, we conduct an additional comparison on longer sequences of the Waymo NOTR validation set. Specifically, we use three 120-frame sequences, a denser Drop 25\% sparsity level, and higher-resolution $640\times960$ images. We compare the fine-tuned version of EVolSplat4D with representative per-scene optimization baselines, including EmerNeRF, StreetGaussian, and OmniRe.

As shown in \tabref{tab:denser_comparision}, EVolSplat4D (Ours-FT) achieves the best SSIM and LPIPS after fine-tuning, while obtaining PSNR comparable to the strongest baseline, OmniRe. Meanwhile, Ours-FT requires shorter reconstruction time than the optimization-based baselines.

\begin{table}[t]
\centering
\resizebox{0.95\linewidth}{!}{%
\begin{tabular}{@{}lcccc@{}}
\toprule
Method & PSNR$\uparrow$ & SSIM$\uparrow$ & LPIPS$\downarrow$ & Rec. Time$\downarrow$ \\
\midrule
EmerNeRF~\citep{emernerf} & 26.70 & 0.863 & 0.125 & 43 min \\
StreetGaussian~\citep{streetgaussian} & 26.91 & 0.870 & 0.092 & 15 min \\
OmniRe~\citep{chen2024omnire} & \textbf{27.35} & 0.884 & 0.087 & 21 min \\
EVolSplat4D (Ours-FT) & 27.28 & \textbf{0.901} & \textbf{0.071} & \textbf{6 min 04 s} \\
\bottomrule
\end{tabular}
}
\caption{\textbf{Comparison with optimization-based baselines} on three 120-frame Waymo NOTR validation sequences under the denser Drop 25\% sparsity setting and higher resolution ($640\times960$). \textbf{Rec. Time} denotes the reconstruction time required to achieve the reported metrics.}
\label{tab:denser_comparision}
\end{table}

\subsection{Efficiency Scaling Analysis}
\begin{figure*}[t]
\centering
\includegraphics[width=0.48\linewidth]{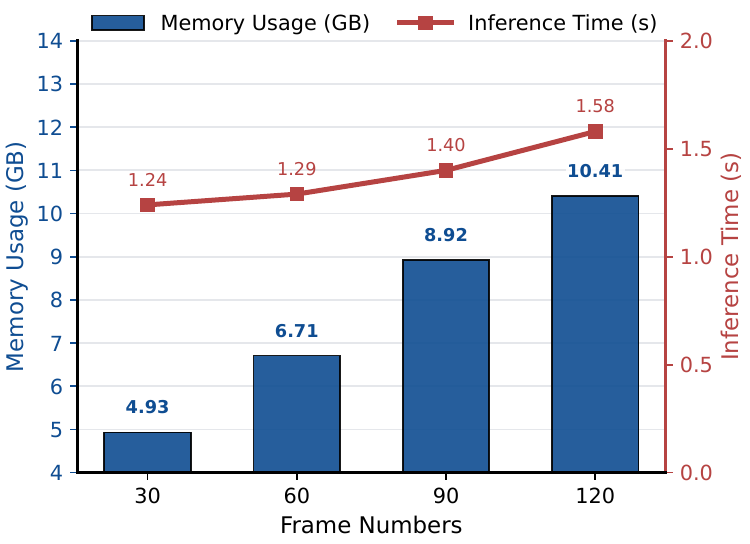}
\includegraphics[width=0.48\linewidth]{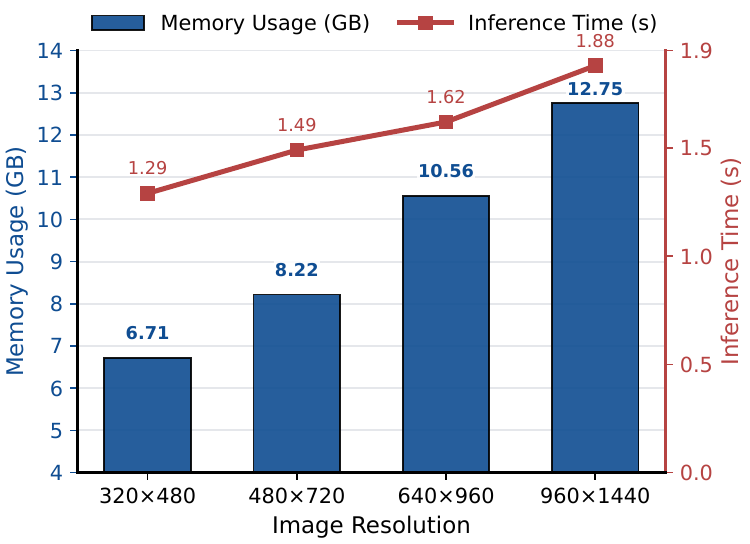}
\caption{\textbf{Efficiency scaling analysis.} We report memory usage and inference time with respect to the number of input frames and image resolution.}
\label{fig:supp_efficiency_scaling}
\end{figure*}

We further analyze the memory usage and inference time of EVolSplat4D with respect to the number of input frames and image resolution. These results in \figref{fig:supp_efficiency_scaling} demonstrate that \method~maintains efficient memory and inference scaling with both the number of input frames and image resolution. When increasing the sequence length from 30 to 120 frames at $320\times480$ resolution, memory usage increases moderately, while inference time remains nearly stable. Similarly, when increasing the image resolution from $320\times480$ to $960\times1440$, both memory usage and inference time increase gradually but remain within a practical range.  Thanks to our volume-based representation, the initial number of Gaussians is mainly determined by the voxelized 3D volume rather than the number of input pixels or frames, and therefore does not grow linearly with image resolution or sequence length. This makes \method~manageable for longer sequences and higher-resolution inputs.

\subsection{Downstream Utility Visualization}
\begin{figure*}[t]
\centering
\setlength{\tabcolsep}{3pt}
\renewcommand{\arraystretch}{0.9}
\begin{tabular}{ccc}
\includegraphics[width=0.29\linewidth]{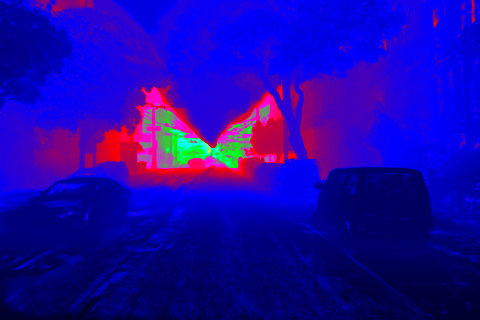} &
\includegraphics[width=0.29\linewidth]{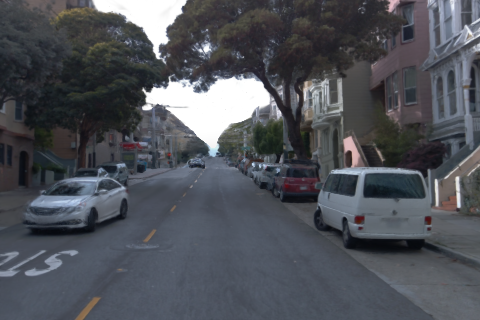} &
\includegraphics[width=0.29\linewidth]{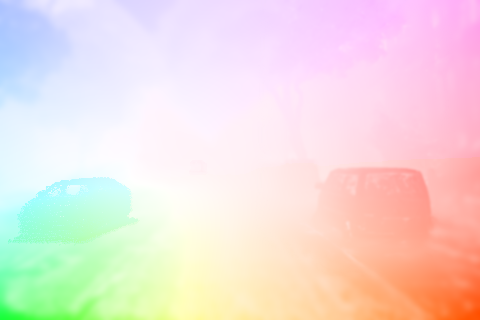} \\
\small (a) Depth & \small (b) Novel view rendering & \small (c) Optical flow \\
\multicolumn{3}{c}{
\begin{tabular}{@{}c@{\hspace{0.12\linewidth}}c@{}}
\includegraphics[width=0.29\linewidth]{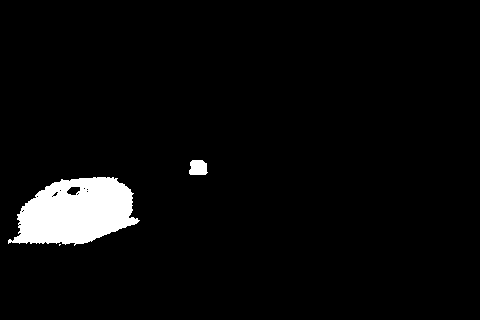} &
\includegraphics[width=0.29\linewidth]{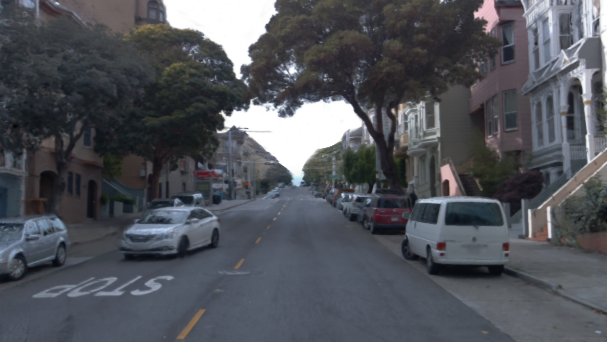} \\
\small (d) Dynamic mask & \small (e) Re-rendering
\end{tabular}
}
\end{tabular}
\caption{\textbf{ Downstream utility visualization,} including novel RGB\&Depth rendering, optical flow, dynamic mask, and re-rendering with modified camera parameters.}
\label{fig:supp_downstream_utility}
\end{figure*}

Beyond held-out RGB rendering, the reconstructed dynamic Gaussians can be rendered into auxiliary outputs, including depth maps, dynamic masks, and optical flow. They can also be re-rendered under modified camera intrinsics, extrinsics, or camera aspect ratios. We provide representative qualitative results in \figref{fig:supp_downstream_utility}.

\subsection{Qualitative Extrapolated View Comparison}
\begin{figure*}[t]
\centering
\renewcommand{\arraystretch}{0}
\setlength{\tabcolsep}{2pt}

\def\labelw{0.035\linewidth}
\def\imgw{0.208\linewidth}

\begin{tabular}{@{} P{\labelw} P{\imgw} P{\imgw} P{\imgw} P{\imgw} @{}}
\rotatebox{90}{\scriptsize 1 m shift} &
\includegraphics[width=\linewidth]{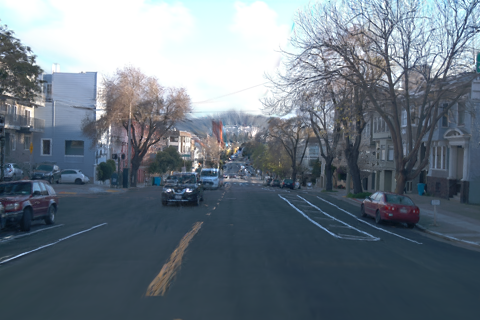} &
\includegraphics[width=\linewidth]{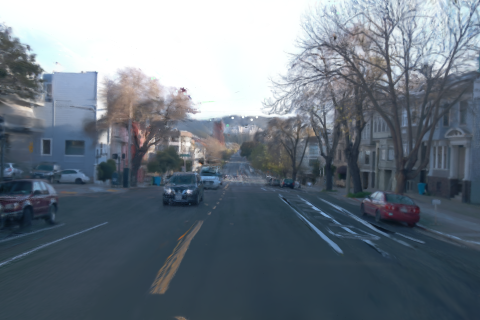} &
\includegraphics[width=\linewidth]{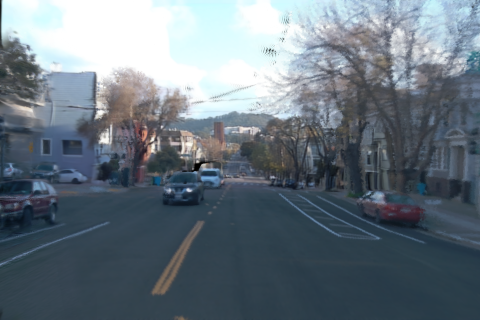} &
\includegraphics[width=\linewidth]{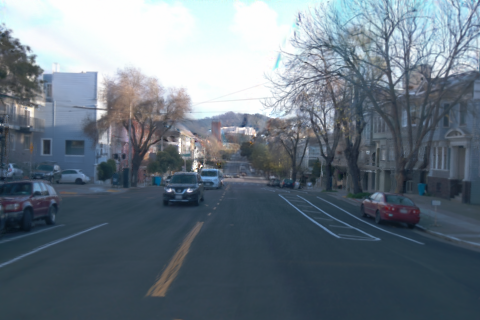} \\

\noalign{\vspace{2pt}}
\rotatebox{90}{\scriptsize 3 m shift} &
\includegraphics[width=\linewidth]{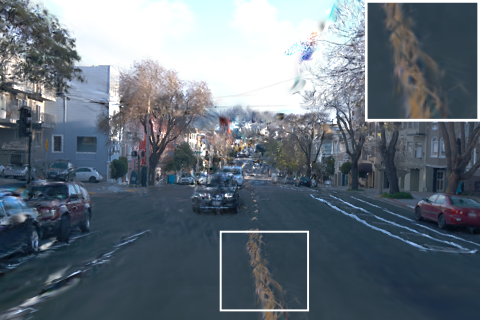} &
\includegraphics[width=\linewidth]{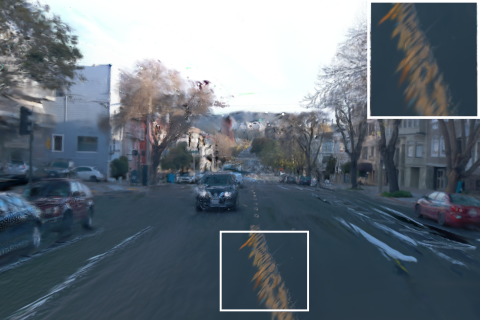} &
\includegraphics[width=\linewidth]{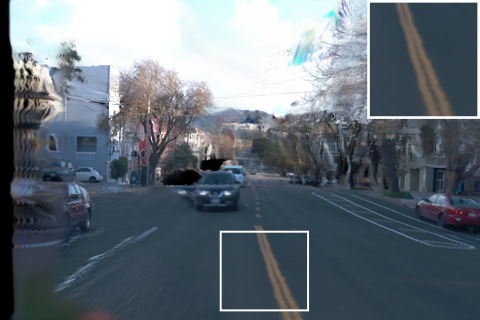} &
\includegraphics[width=\linewidth]{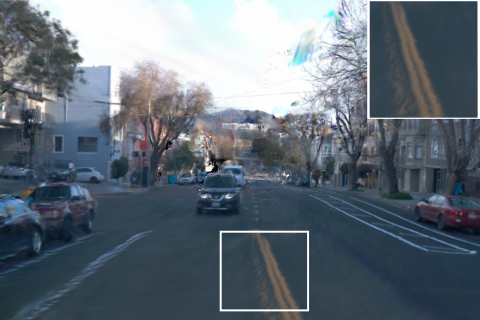} \\

\noalign{\vspace{4pt}}
& \small StreetGaussian & \small OmniRe & \small Ours-FF & \small Ours-FT
\end{tabular}

\caption{\textbf{Qualitative comparison of view extrapolation quality.} We compare 1 m and 3 m lane-shifted rendering results with optimization-based baselines.}
\label{fig:supp_extrap_shift_comparison}
\end{figure*}

We provide qualitative comparisons of 1 m and 3 m lane-shifted rendering results with optimization-based baselines, including StreetGaussian and OmniRe. As shown in  \figref{fig:supp_extrap_shift_comparison}, our fine-tuned model achieves the best KID at 1 m on both Waymo and PandaSet (0.038 and 0.029, respectively). Under the more challenging 3 m lane shift, Ours-FT remains comparable to the strongest optimization-based baseline. As discussed in the limitation section, our purely feed-forward model shows a performance drop in background regions under the larger 3 m shift, since the far-field branch becomes less reliable under large camera translations. Nevertheless, the qualitative results show that it still preserves clear close-range structures, such as lane markings, and avoids distortion and blur observed in StreetGaussian and OmniRe.

\subsection{Scene Editing Comparisons and Failure Cases}
\begin{figure}[t]
\centering
\renewcommand{\arraystretch}{0}
\setlength{\tabcolsep}{1pt}

\def\imgwidth{0.46\linewidth}
\def\labelwidth{0.04\linewidth}

\begin{tabular}{@{} P{\labelwidth} P{\imgwidth} P{\imgwidth} @{}}

\rotatebox{90}{\scriptsize Reference} &
\includegraphics[width=\linewidth]{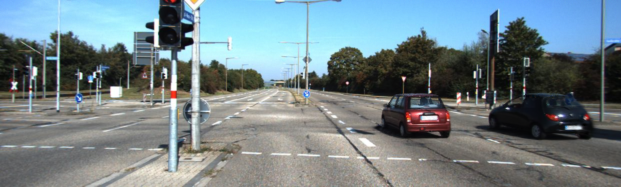} &
\includegraphics[width=\linewidth]{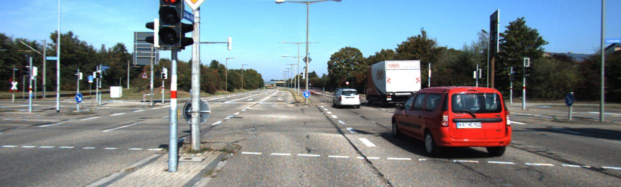}
\\

\noalign{\vspace{1pt}}
\rotatebox{90}{\scriptsize OmniRe} &
\includegraphics[width=\linewidth]{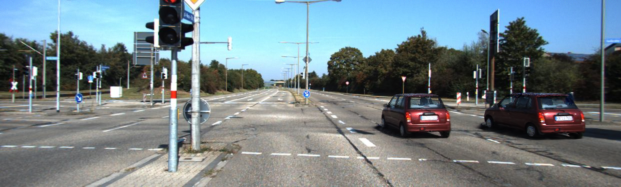} &
\includegraphics[width=\linewidth]{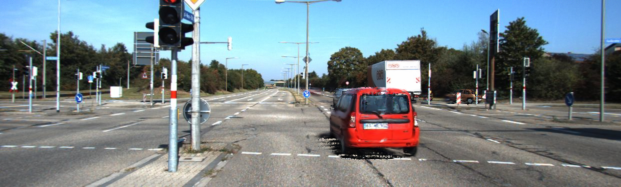}
\\

\noalign{\vspace{1pt}}
\rotatebox{90}{\scriptsize Ours-FF} &
\includegraphics[width=\linewidth]{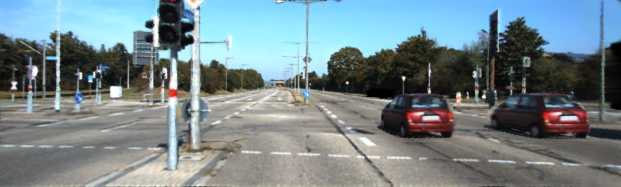} &
\includegraphics[width=\linewidth]{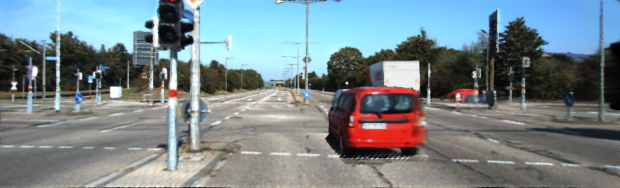}
\\

\noalign{\vspace{1pt}}
\rotatebox{90}{\scriptsize Ours-FT} &
\includegraphics[width=\linewidth]{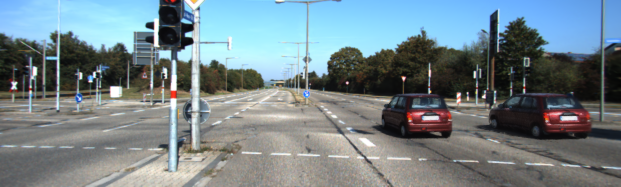} &
\includegraphics[width=\linewidth]{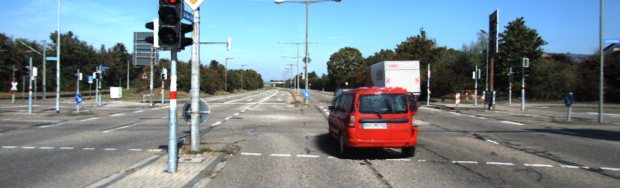}
\\

\noalign{\vspace{2pt}}
&
\small (a) Object replacement &
\small (b) Object shift

\end{tabular}

\caption{\textbf{Qualitative editing comparison} with OmniRe on actor replacement and 2 m lane shift.}
\label{fig:supp_edit_comparison}
\end{figure}

We provide qualitative editing comparisons with OmniRe in \figref{fig:supp_edit_comparison}. Ours-FF denotes the purely feed-forward result, while Ours-FT denotes the result after 1000 fine-tuning iterations. Ours-FT produces more realistic actor appearance with sharper shadow details, while OmniRe shows visible editing artifacts and blurry rendering in these regions.

\begin{figure}[t]
\centering
\includegraphics[width=0.85\linewidth]{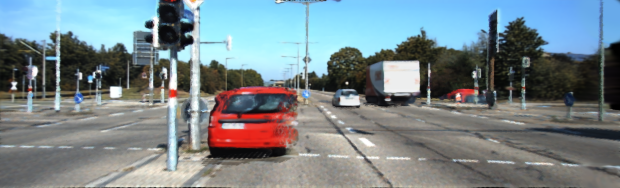}
\caption{\textbf{Failure case} for actor lane shift under a large translation.}
\label{fig:supp_edit_failure}
\end{figure}

We also provide a failure case for the lane shift in \figref{fig:supp_edit_failure}. When translating a vehicle by a large margin (6 m), artifacts may appear in previously unobserved regions, such as the occluded side of the vehicle.

\subsection{Qualitative Results under Bounding Box Perturbations}
\begin{figure}[t]
\centering
\setlength{\tabcolsep}{2pt}
\renewcommand{\arraystretch}{0.95}
\begin{tabular}{cc}
\includegraphics[width=0.48\linewidth]{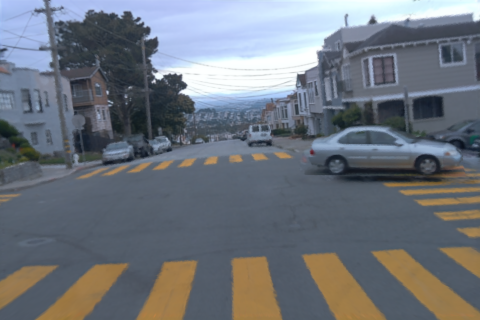} &
\includegraphics[width=0.48\linewidth]{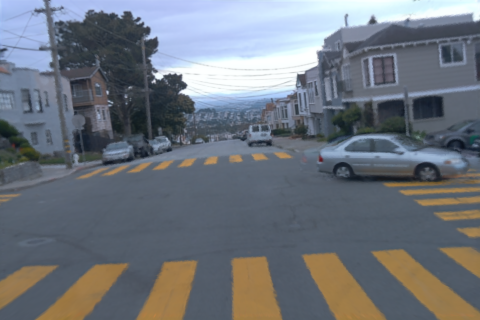} \\
\small Center perturbation $\sigma=0.5\,\mathrm{m}$ & \small Yaw perturbation $\sigma=15^\circ$ \\[3pt]
\includegraphics[width=0.48\linewidth]{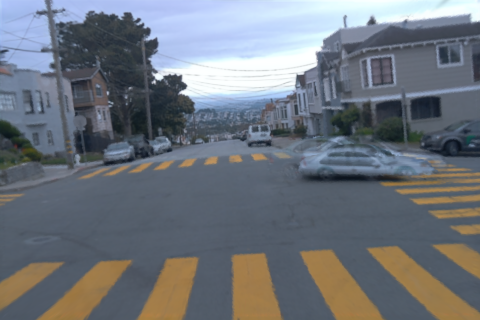} &
\includegraphics[width=0.48\linewidth]{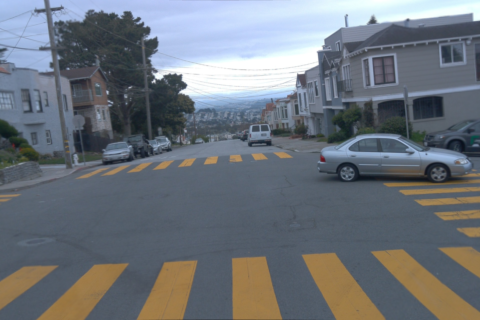} \\
\small Failure case & \small Ground truth
\end{tabular}
\caption{\textbf{Qualitative results of bounding box perturbation} on Waymo.}
\label{fig:supp_box_noise_qualitative}
\end{figure}

We provide qualitative results under perturbed predicted 3D bounding boxes in \figref{fig:supp_box_noise_qualitative}. Center perturbation may cause the vehicle to float slightly from its original location, but the rendered actor remains relatively sharp. In contrast, yaw perturbation mainly introduces blur around object boundaries. We also show one failure case under degraded boxes with translation noise $\sigma=0.8\,\mathrm{m}$ and yaw noise $\sigma=30^\circ$.

\section{More Experimental Results}
\subsection{More Static Feed-Forward Qualitative results}
 We provide more qualitative results on the Waymo dataset via a feed-forward inference on the static scenes. MuRF~\citep{xu2023murf} achieves promising rendering results on Waymo, but it struggles with poor geometric reconstruction in texture-less areas, such as roads, due to its strong reliance on feature mapping. To better present the improvements in the far-range regions, we provide zoomed-in comparisons in \figref{fig:waymoQualitive}. EVolSplat~\citep{miao2025evolsplat}, despite achieving high-quality rendering in the close range, presents clear artifacts in the background region due to its coarse hemisphere background model. In contrast, our Generalizable Far-Field Branch effectively models the complex, fine-grained distant landscapes, overcoming the limitations of prior work.

\subsection{Qualitative Comparison with Optimization-based Methods}

We present qualitative comparisons in \figref{fig:finetune_comparison} under the Drop80\% setting. The quantitative results are reported in the main paper. Our method achieves superior performance against all baselines, which we attribute to the robust, generalizable priors.

\subsection{More Dynamic Feed-Forward Qualitative results}
\begin{figure*}[htbp]
     \centering
     \setlength{\tabcolsep}{1pt}
     \def\mywidth{.33}
     \def\reduceheight{-3.pt}
     \begin{tabular}{ccc}
      \includegraphics[width=\mywidth\linewidth]{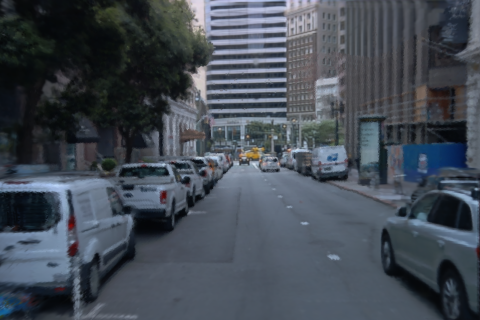} &
      \includegraphics[width=\mywidth\linewidth]{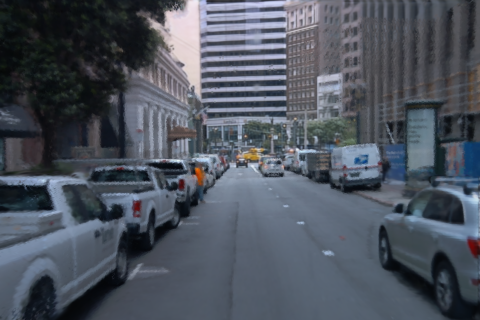} &
      \includegraphics[width=\mywidth\linewidth]{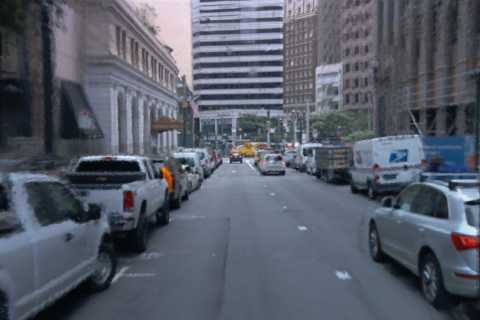} 
      \vspace{\reduceheight}\\
      \includegraphics[width=\mywidth\linewidth]{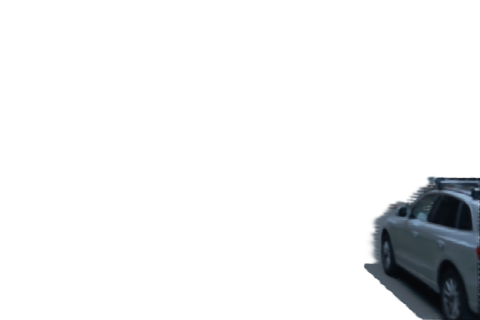} &
      \includegraphics[width=\mywidth\linewidth]{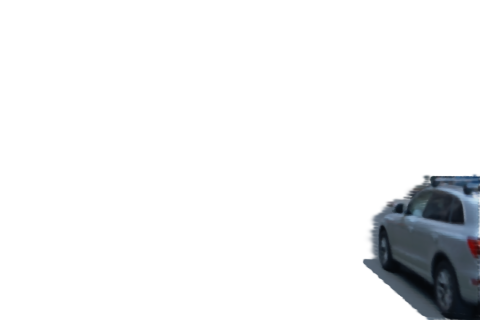} &
      \includegraphics[width=\mywidth\linewidth]{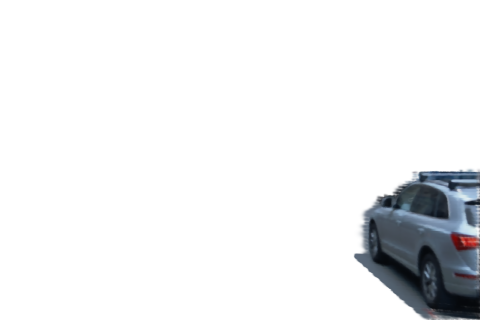} 
      \\

      \includegraphics[width=\mywidth\linewidth]{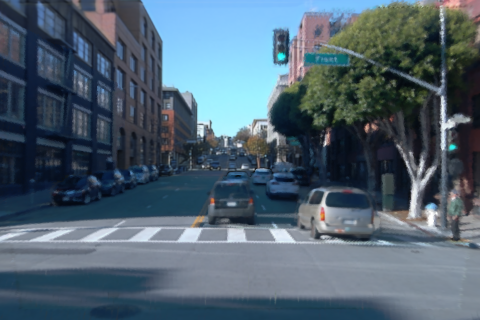} &
      \includegraphics[width=\mywidth\linewidth]{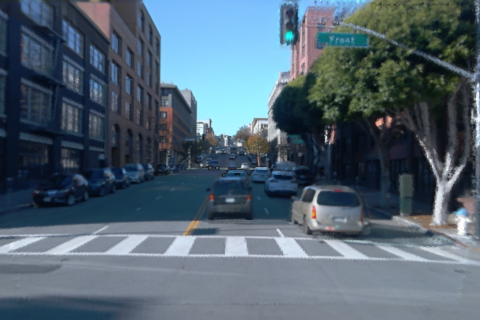} &
      \includegraphics[width=\mywidth\linewidth]{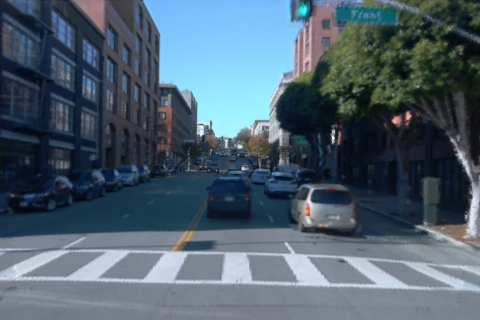} 
      \vspace{\reduceheight}\\
      \includegraphics[width=\mywidth\linewidth]{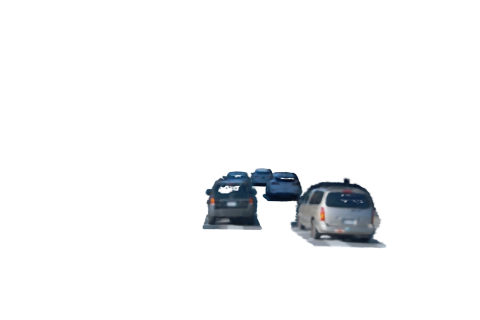} &
      \includegraphics[width=\mywidth\linewidth]{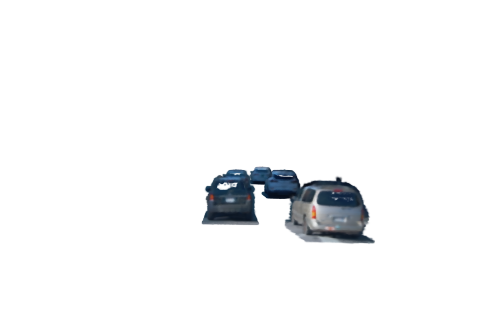} &
      \includegraphics[width=\mywidth\linewidth]{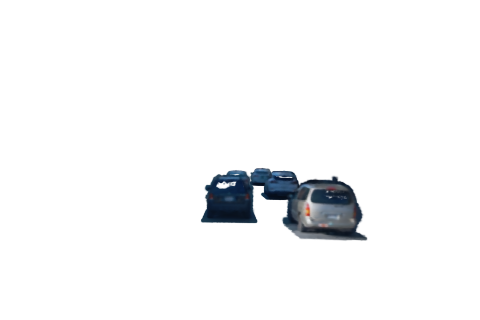} 
      \\
      
     \end{tabular}\vspace{-0.2cm}
     \caption{{\bf More Qualitative Results on Waymo}. We visualize synthesized images (odd rows) and the corresponding predicted dynamic actors (even rows) on novel scenes generated by our pretrained model through a feed-forward inference.}
     \label{fig:more waymo res}
     \vspace{-0.021cm}
    \end{figure*}

\begin{figure*}[htbp]
     \centering
     \setlength{\tabcolsep}{1pt}
     \def\mywidth{.33}
     \def\reduceheight{-3.pt}
     \begin{tabular}{ccc}
      \includegraphics[width=\mywidth\linewidth]{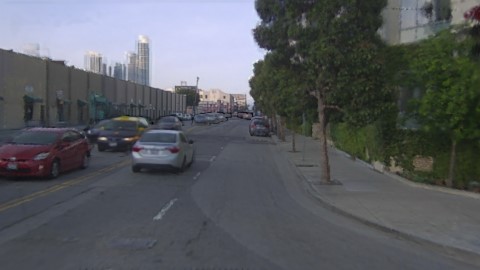} &
      \includegraphics[width=\mywidth\linewidth]{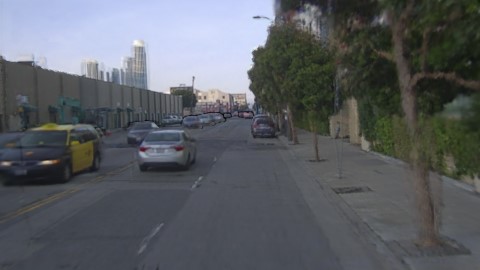} &
      \includegraphics[width=\mywidth\linewidth]{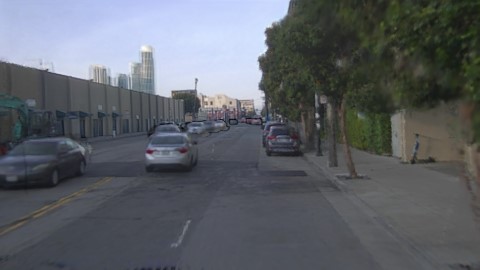} 
      \vspace{\reduceheight}\\
      \includegraphics[width=\mywidth\linewidth]{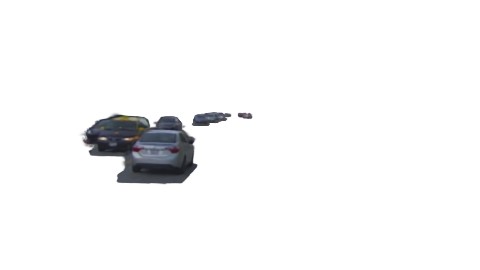} &
      \includegraphics[width=\mywidth\linewidth]{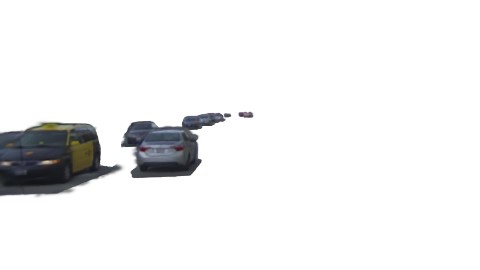} &
      \includegraphics[width=\mywidth\linewidth]{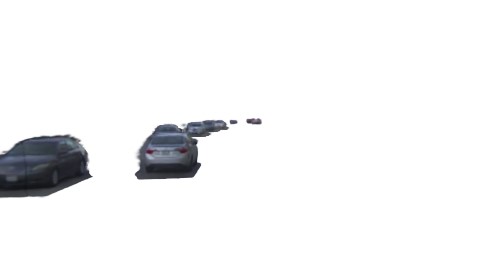} 
      \\

      \includegraphics[width=\mywidth\linewidth]{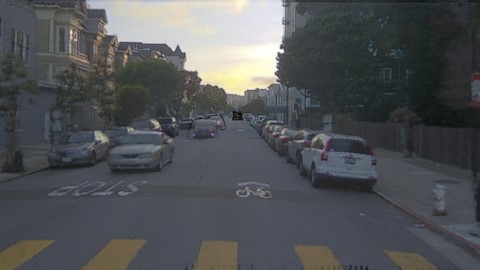} &
      \includegraphics[width=\mywidth\linewidth]{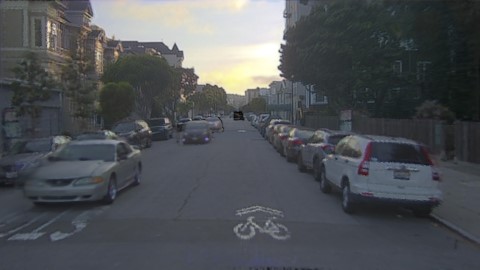} &
      \includegraphics[width=\mywidth\linewidth]{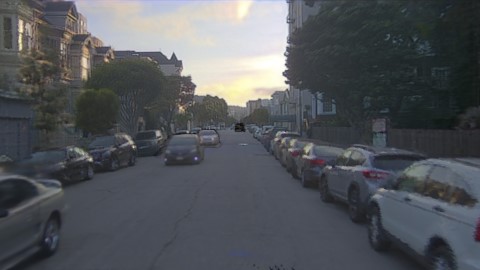} 
      \vspace{\reduceheight}\\
      \includegraphics[width=\mywidth\linewidth]{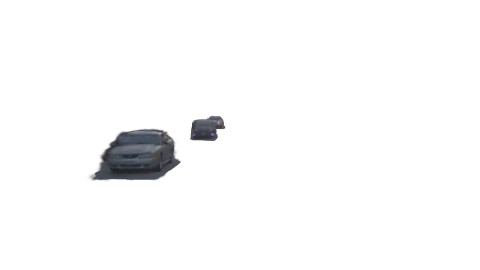} &
      \includegraphics[width=\mywidth\linewidth]{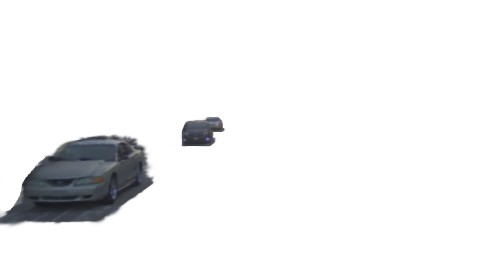} &
      \includegraphics[width=\mywidth\linewidth]{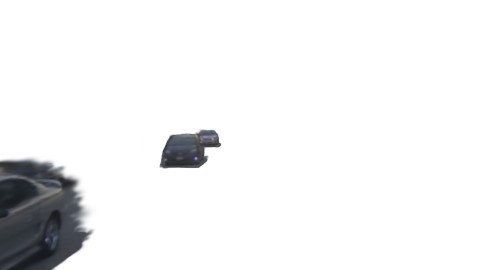} 
      \\
      
     \end{tabular}\vspace{-0.2cm}
     \caption{{\bf More Qualitative Results on out-of-domain dataset PandaSet}. We visualize synthesized images (odd rows) and the corresponding predicted dynamic actors (even rows) on novel scenes generated by our pretrained model through a feed-forward inference.}
     \label{fig:more pandaset res}
     \vspace{-0.021cm}
    \end{figure*}

\begin{figure*}[tbp]
     \centering
     \setlength{\tabcolsep}{1pt}
     \def\mywidth{.33}
     \def\reduceheight{-3.pt}
     \begin{tabular}{ccc}
      \\
      
      \includegraphics[width=\mywidth\linewidth]{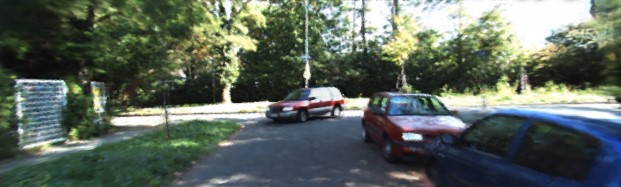} &
      \includegraphics[width=\mywidth\linewidth]{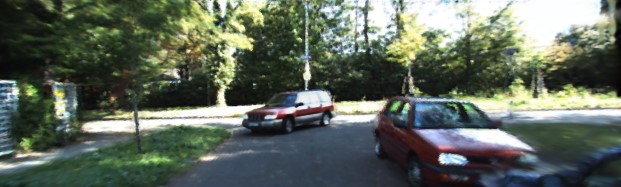} &
      \includegraphics[width=\mywidth\linewidth]{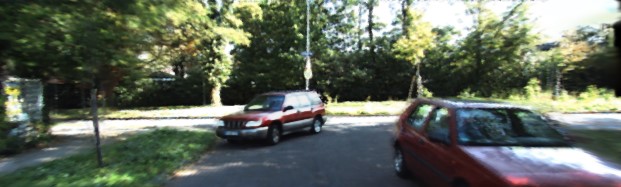} 
      \vspace{\reduceheight}\\
      \includegraphics[width=\mywidth\linewidth]{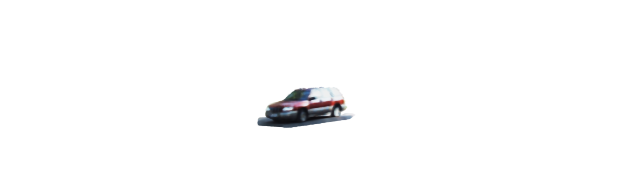} &
      \includegraphics[width=\mywidth\linewidth]{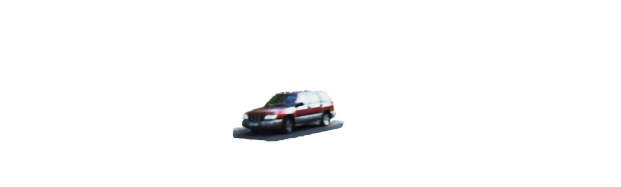} &
      \includegraphics[width=\mywidth\linewidth]{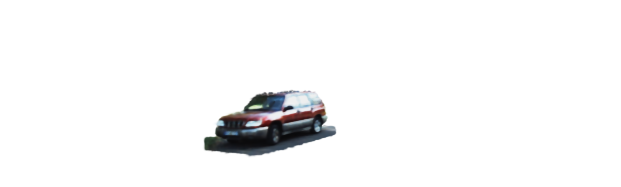} 
      \\
      
      \includegraphics[width=\mywidth\linewidth]{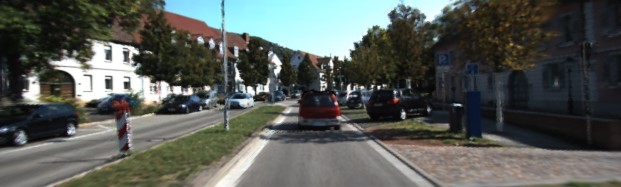} &
      \includegraphics[width=\mywidth\linewidth]{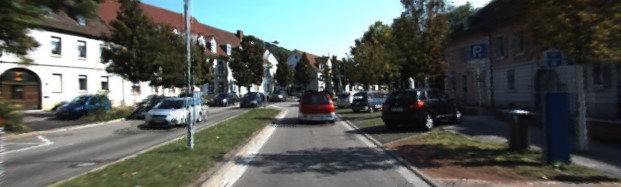} &
      \includegraphics[width=\mywidth\linewidth]{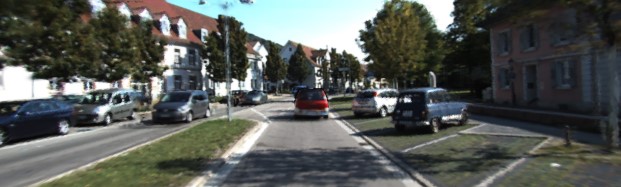} 
      \vspace{\reduceheight}\\
      \includegraphics[width=\mywidth\linewidth]{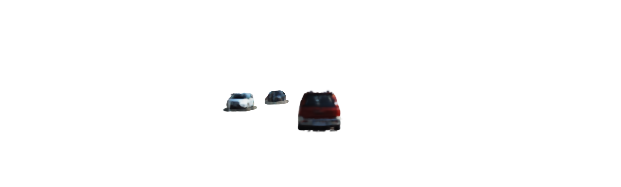} &
      \includegraphics[width=\mywidth\linewidth]{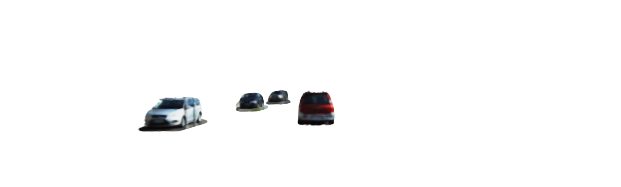} &
      \includegraphics[width=\mywidth\linewidth]{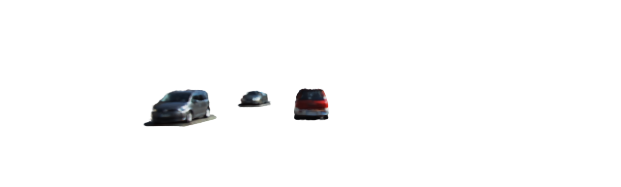} 
      
     \end{tabular}
     \caption{{\bf More Qualitative Results on KITTI}. We visualize synthesized images (odd rows) and the corresponding predicted dynamic actors (even rows) on novel scenes generated by our pretrained model through a feed-forward inference.}
     \label{fig:more kitti res}
     \vspace{-0.021cm}
    \end{figure*}

Our method enables efficient dynamic reconstruction and real-time photorealistic NVS from flexible sparse street view images. We provide more qualitative dynamic reconstruction results on three different datasets (KITTI, Waymo and PandaSet) as shown in \figref{fig:more waymo res} and \figref{fig:more pandaset res}, and \figref{fig:more kitti res}.

\end{document}